\documentclass{article}
\RequirePackage{xcolor}
\usepackage[final]{corl_2018} %

\usepackage{url}
\usepackage{graphicx}
\usepackage{amsmath}
\usepackage{tikz}
\usepackage{todonotes}
\usepackage{amsfonts}
\usepackage{float}
\usepackage{rotating}
\usetikzlibrary{automata,positioning}
  \usepackage{multirow}
  \newcommand*\samethanks[1][\value{footnote}]{\footnotemark[#1]}

\title{Reward Estimation for Variance Reduction in Deep Reinforcement Learning}

\author{
Joshua Romoff$^{1,2}$\thanks{Authors Contributed Equally}, Peter Henderson$^1$\samethanks,\\ \textbf{ Alexandre Pich\'{e}$^{3}$, Vincent Fran\c{c}ois-Lavet$^1$, Joelle Pineau$^{1,2}$}\\
$^1$ MILA, McGill University, Montr\'{e}al, Qu\'{e}bec, Canada \\
$^2$ Facebook AI Research, Montr\'{e}al, Qu\'{e}bec, Canada \\
$^3$ MILA, Universit\'{e} de Montr\'{e}al, Qu\'{e}bec, Canada \\
}

\begin{document}
\maketitle

\begin{abstract}
Reinforcement Learning (RL) agents require the specification of a reward signal for learning behaviours. However, introduction of corrupt or stochastic rewards can yield high variance in learning. Such corruption may be a direct result of goal misspecification, randomness in the reward signal, or correlation of the reward with external factors that are not known to the agent. Corruption or stochasticity of the reward signal can be especially problematic in robotics, where goal specification can be particularly difficult for complex tasks. While many variance reduction techniques have been studied to improve the robustness of the RL process, handling such stochastic or corrupted reward structures remains difficult. As an alternative for handling this scenario in model-free RL methods, we suggest using an estimator for both rewards and value functions. We demonstrate that this improves performance under corrupted stochastic rewards in both the tabular and non-linear function approximation settings for a variety of noise types and environments. The use of reward estimation is a robust and easy-to-implement improvement for handling corrupted reward signals in model-free RL.
\end{abstract}

\keywords{Reinforcement Learning, Uncertainty, Goal Specification} 

\section{Introduction}

Reinforcement Learning (RL) agents learn from a generated reward provided by the environment. 
However, it is possible that the generated reward is corrupted~\cite{moreno2006noisy,everitt2017reinforcement}, stochastic~\cite{campbell2015handling}, or misspecified~\cite{everitt2017reinforcement}. 
The specification of rewards which do not exhibit these problems can be especially difficult in robotics and has resulted in data-driven approaches for reward specification~\cite{fu2018variational,hadfield2017inverse,sermanet2016unsupervised}. 
However, these data-driven approaches may also yield corrupted reward signals where the sensory-based features used for reward generation are themselves corrupted. 
As such, handling corrupted or stochastic rewards in the learning process is a necessity for successfully learning complex behaviours in robotics using RL \citet{everitt2017reinforcement}.

In particular, such scenarios can result in high variance in the gradients during learning and impede successful convergence to an optimal policy. Several methods have already been used to reduce variance, sometimes at the cost of bias. These include generalized advantage estimation~\citep{Schulmanetal_ICLR2016}, 
constrained updates~\citep{schulman2017proximal},
updating the target policy via the expectation of its actions~\citep{ciosek:aaai18,asadi2017mean}, and updating the value function via the posterior mean of an estimated uncertain value distribution~\citep{henderson2017bayesian}. However, these don't explicitly account for corrupted rewards and aim to address variance induced with a deterministic true reward.

Here, we propose a simple method for updating model-free RL algorithms to compensate for stochastic corrupted reward signals. We suggest learning an estimator for both the local expected reward and the value function -- that is, using a direct estimate of rewards $\hat{R}(s_t)$ to update the discounted value function $V^\pi_\gamma(s_t)$ and policy $\pi_{\theta}(s_t)$, rather than the sampled rewards. 

We show that this method results in theoretical variance reductions in the tabular case and corresponds to empirical performance gains in the tabular and function approximation settings in situations where rewards are highly stochastic and corrupted. We validate this on the MuJoCo environments~\cite{todorov2012mujoco} from OpenAI Gym~\cite{brockman2016openai} for continuous control robotic locomotion benchmark tasks and provide complementary results on Atari games for discrete settings.

\section{Related Work}

A variety of model-based work in robotic and non-robotic domains has used reward estimation~\cite{racaniere2017imagination,silver2016predictron,henaff2017model,feinberg2018model,van2013efficient, franccois2018combined}. However, those works use the predicted (or ``imagined'') reward for planning rather than training a value function. In several cases, estimated rewards are used in imagination augmented rollouts with a stochastic dynamics model accompanying the reward estimator. For example, \cite{feinberg2018model,sutton1990integrated} use a method to apply model-based methods to model-free RL. However, in our case we do not require multi-step imaginary rollouts, avoiding learning system dynamics as our reward estimation method is explicitly aimed at handling corrupted stochastic rewards rather than for planning. Nevertheless, it may be possible to view reward estimation as a single step case of model-based value expansion.

Similarly, the myriad of inverse reinforcement learning (IRL) literature for robotics involves learning reward functions from demonstrations rather than previous rewards. For example in \cite{hadfield2017inverse,sermanet2016unsupervised}, rewards are learned in a data-driven way explicitly to better model the desired robotic behaviours. In some cases, such as in~\cite{sermanet2016unsupervised,metelli2017compatible}, the reward function is modified to be compatible and beneficial to the learning process. However, this doesn't specifically account for corruption in the learned rewards. 

Other works in the RL setting augment rewards via shaping mechanisms~\cite{grzes2009learning, sorg2012variance}; for example, to make robot learning easier with sparse rewards. Our method can be viewed as a shaping mechanism as well, where the transformation is captured within a single function approximator.
While much of the reward shaping literature aims to aid exploration, \citet{talvitie2018learning} comes close to our work by learning to correct the reward function for misspecifications of the model.

Generally, while all of these works model rewards in some way, most do not explicitly seek to address a corrupted reward. \citet{everitt2017reinforcement}, on the other hand set up the problem of corrupted rewards formally and suggest a method to address corrupted reward channels in small GridWorld scenarios. However, the method they introduce is specific to the tabular setting and they do not propose any methods for continuous control as is often needed for handling complex robotic tasks.

\section{Background}
\label{sec:background}

\subsection{Reinforcement Learning}
We formulate our method in the context of a fully observable Markov Decision Process (MDP)~\cite{bellman1957markovian}. In an MDP, an agent can take an action $a_t$ based on its current state $s_t$ and receive a reward $r_t$, before transitioning to the next state of the MDP $s_{t+1}$. We focus on the discounted MDP case, where an agent tries to maximize the cumulative discounted reward $V_\gamma^\pi(s) = \left[\sum_{t=0}^\infty \gamma^{t} r_t | s_0 = s, \pi \right]$, also known as the discounted value of a policy $\pi$. It is common to learn a value estimate of the current policy via temporal difference (TD) learning~\cite{Sutton1988:td}, where the current estimate of the value function is used to bootstrap the next estimate according to the Bellman target $Y_t = r_t + \gamma V_\gamma^\pi(s_{t+1})$, via the loss: $\mathcal{L}(\theta_V) =  \mathbb{E} \left[ \left(Y_t - V_\gamma^\pi (s_t; \theta_V) \right)^2 \right]$. In the case of Advantage Actor Critic (A2C), the synchronous version of Asynchronous Advantage Actor Critic (A3C)~\citep{mnih2016asynchronous}, a stochastic parameterized policy (actor, $\pi_\theta(a | s)$) is learned from this value estimator via the TD error. That is, the actor loss becomes: $\mathcal{L}(\theta_\pi) = \mathbb{E} \left[- \log \pi(a ,s; \theta_\pi) \left( r_t + \gamma V_\gamma^\pi(s_{t+1}; \theta_V) - V_\gamma^\pi (s_t; \theta_V) \right) \right]$. 

Proximal Policy Optimization (PPO)~\cite{schulman2017proximal} can be considered to be a similar method to A2C. In the case of PPO, however, long Monte Carlo rollouts are used while the value function acts primarily as a variance-reducing baseline in the policy update -- typically via generalized advantage estimation~\cite{schulman2015high}. Furthermore, the policy update is constrained via a trust region in the form of a clipping objective (as we use here) or a divergence penalty. The clipping objective for training the policy is:
\begin{equation}
    L^{CLIP} (\theta) = \hat{\mathbb{E}} \left[ \min ( r_t(\theta) \hat{A_t}, \text{clip} (r_t (\theta), 1- \epsilon, 1+\epsilon)\hat{A_t} \right] 
\end{equation}
where the likelihood ratio is $r_t(\theta) = \frac{\pi_\theta (a_t | s_t)}{\pi_{\theta_{old}}(a_t|s_t)}$, $\hat{A_t}$ is the generalized advantage function, and $\epsilon < 1$ is some small factor applied to constrain the update.

\subsection{Stochastic or Corrupted Rewards}

Since we tackle the case of using stochastic or corrupted rewards, it is important to have a clear understanding of this scenario. In stochastic rewards, for a state transition tuple $(s,a,s')$, a reward can be treated as a random variable. That is, the reward is provided from some distribution with a certain probability density. This corresponds to settings in robotics where a reward function may be learned through variational means~\cite{fu2018variational}.

We further define a corrupted reward to be such that the provided reward does not match the true reward due to some noise process, similarly to the Corrupted Reward MDP (CRMDP) setting of~\cite{everitt2017reinforcement}. In the cases we consider, the reward is both corrupted and stochastic. That is, for a given state transition tuple $(s,a,s')$, the true reward is $r(s,a,s')$, whereas the corrupted stochastic reward becomes a random variable $\Tilde{R}$ such that the likelihood of a reward being sampled from the corrupted stochastic random variable is $P(\Tilde{R}=r| s,a,s')$.

\section{Reward Estimation}
\label{method}

Under a corrupted stochastic reward, an additional source of variance is injected into the value function update. To reduce this variance, we introduce an estimator for the reward at a given state $\hat{R}(s_t)$. In the function approximation case, learning this reward estimator becomes a simple regression problem: $\mathcal{L} (\theta_{\hat{R}}) = \mathbb{E} \left[ \left(r_t - \hat{R}(s_t;\theta_{\hat{R}})\right)^2 \right]$. We then use this reward estimator in the TD update of the value function, rather than the sampled reward:  $\mathcal{L}(\theta_V) = \mathbb{E} \left[ \left(\hat{R}(s_t;\theta_{\hat{R}}) + \gamma V_\gamma^\pi(s_{t+1};\theta_V) - V_\gamma^\pi (s_t;\theta_V) \right)^2\right]$. As we will see in Section~\ref{sec:proof}, under corrupted stochastic rewards, this estimation will reduce the variance propagated to the value function.

\begin{figure}[!htbp]
    \centering
    \includegraphics[width=0.5\textwidth]{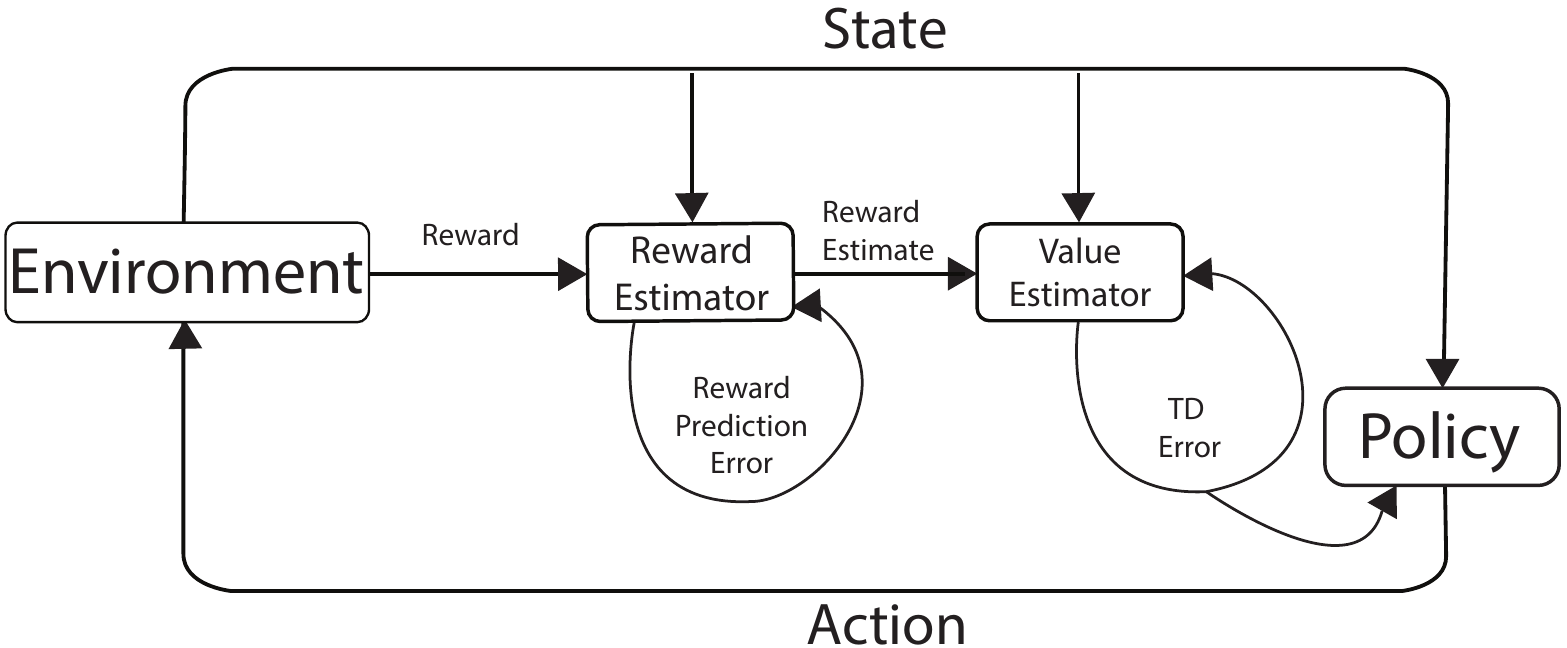}
    \caption{The actor-critic update process with the reward estimator. }
    \label{fig:ac_model}
\end{figure}

We note that this is an easy update to model-free methods which does not significantly change the problem formulation and can be used in \emph{any} model-free method. An example of using the reward estimator in an actor-critic process can be seen in Figure~\ref{fig:ac_model}.

\subsection{Theoretical Variance Reduction in Tabular Domains}
\label{sec:proof}
To determine whether our method for using a reward approximator reduces variance theoretically, we examine the tabular case. In this setting, we use the sample mean for the reward estimator: $\hat{R}(s, a, s') = \left[\frac{1}{N}\sum_i r^i \right]$ where $r^i \sim R(s,a,s')$ for all $i \in [1, \ldots, N]$. That is, given $N$ i.i.d reward samples at a given state $s$ where action $a$ was taken and where we transitioned to $s'$, we determine the mean of those rewards\footnote{In domains where the reward signal is fully determined by $s$ or by $(s,a)$, $\hat{R}(s)$ or $\hat{R}(s, a)$ can be used instead respectively.}. In this scenario, the sample mean is an unbiased estimator. 

First, following a similar methodology to the variance analysis by~\citet{van2009theoretical}, we determine the variance of the standard discounted Bellman equation: \citep{bellman}: $G_t^\gamma = r_t + \gamma V_\pi^\gamma ( s_{t+1})$. The variance of this Bellman estimate is: 
\begin{equation}
    \operatorname{var}\left[ G_{t}^\gamma \right] = \operatorname{var}\left[ r_t \right] + \operatorname{var}\left[ \gamma V_\pi^\gamma (s_{t+1} )\right] + 2 \operatorname{cov}\left[ r_t,\gamma V_\pi^\gamma (s_{t+1}) \right].
\end{equation}

If we instead use an approximator for the reward, the Bellman equation becomes: $\hat{G}_t^\gamma = \hat{R}(s_t, a_t, s_{t+1}) + \gamma V_\pi^\gamma ( s_{t+1})$. Similarly, the variance becomes: 
\begin{equation}
    \operatorname{var}\left[ \hat{G}_{t}^\gamma \right] = \operatorname{var}\left[ \hat{R}(s_t, a_t, s_{t+1}) \right] + \operatorname{var}\left[ \gamma  V_\pi^\gamma (s_{t+1} )\right] + 2 \operatorname{cov}\left[ \hat{R}(s_t, a_t, s_{t+1}), \gamma V_\pi^\gamma (s_{t+1} ) \right].
\end{equation} 

Moreover, since approximation in the tabular case is simply the sample mean, we have that:
\begin{equation}
    \operatorname{var}\left[\hat{R}(s_t, a_t, s_{t+1}) \right] =  \frac{1}{N} \operatorname{var}\left[ r_t \right]
\end{equation} 
and,
\begin{equation}
    \operatorname{cov}\left[ \hat{R}(s_t, a_t, s_{t+1}),  \gamma V_\pi^\gamma (s_{t+1} ) \right] = \frac{1}{N} \operatorname{cov}\left[ r_t,  \gamma V_\pi^\gamma (s_{t+1} ) \right]
\end{equation}
Thus, we arrive at the following equality: 
\begin{equation}
\label{eq:equality}
    \operatorname{var}\left[ \hat{G}_{t}^\gamma \right] - \operatorname{var}\left[ G_{t}^\gamma \right] = \frac{1}{N} \operatorname{var}\left[ r_t \right] +  \frac{2}{N} \operatorname{cov}\left[ r_t,  \gamma V_\pi^\gamma (s_{t+1} ) \right]  - \operatorname{var}\left[ r_t \right] - 2\operatorname{cov}\left[ r_t,  \gamma V_\pi^\gamma (s_{t+1} ) \right] .
\end{equation} 
Analyzing Equation~\ref{eq:equality}, one can see that if the covariance between the reward and the value function at the next state is $\geq 0$ that $
 \operatorname{var}\left[ \hat{G}_{t}^\gamma \right]  \leq \operatorname{var}\left[ G_{t}^\gamma \right], \forall N \geq 1.$ We note that this is always true when the reward function depends only on $(s,a)$ and not $s'$, since the covariance in this case is $0$. Moreover, even when the reward function depends on $s'$ it is likely to have a positive covariance. We refer to the Appendix for a more lengthy discussion and to our results in Section~\ref{sec:varianceexp} and Appendix~\ref{tab:var1} that highlight the variance reduction empirically.  %

Therefore, by using the empirical mean of the rewards in a tabular setting, it is possible to reduce the variance of the update.
The M-Step return case follows similarly, holding under the same covariance assumptions.
The intuitive benefit of this becomes clear in settings with stochastic corrupted rewards. 
In such a case, the error will propagate through longer MDP chains, whereas using the empirical mean will provide a more stable estimate, as will be demonstrated in subsequent experimental sections. 

\subsection{Choosing the Best Estimator}
The features provided to the reward estimator can be updated to refine the estimate or provide an expectation. Specifically, we can model the reward in three different ways: 
\begin{equation}
\label{eq:rhats}
    R(s) = \mathbb{E}_{s',a \sim \pi} [r|s], \quad R(s, a) = \mathbb{E}_{s'} [r|s,a], \quad \text{or} \quad R(s,a,s') = \mathbb{E}[r|s,a,s'].
\end{equation}

Overall, the inputs provided to the reward function capture expectations at different levels, sometimes encompassing the dynamics of the system. In our case, we focus on deterministic dynamics but where the reward can be treated as a random variable so any of these estimators are effective to varying degrees. However, choosing the best features to use as in Equation~\ref{eq:rhats}
can contain various benefits and tradeoffs. 
The most obvious trade-off depends on the true reward function of the underlying MDP that is being estimated. 
If the true reward function depends on the action and next state then not including one or the other as inputs will make estimation of the rewards more difficult (or impossible). 
For example, in some of the OpenAI Gym MuJoCo benchmark robotics environments, a reward is provided based on the action resulting in a penalty for large amounts of generated torque on the simulated motors. Without providing the action as a feature to the learned reward function it may be difficult to successfully learn a reward estimator.
On the other hand, as we describe in the Appendix, in domains like Atari where the reward signal is typically delayed several steps and mostly dependent on only the current state, $R(s)$ is an adequate choice. 
Using $R(s,a)$ or $R(s,a,s')$ would not provide any benefit and simply make estimating the reward more difficult due to the extra inputs. Empirical results which demonstrate these effects can be found in the Appendix.

\section{Experiments}

To validate that using a function approximator $\hat{R}$ for the reward improves performance, we investigate several settings with induced stochastic noise\footnote{Code provided at \url{https://github.com/facebookresearch/reward-estimator-corl}.}. We investigate a small toy MDP problem in the tabular case to show the variance reduction properties of the system in the tabular setting which validate our theoretical reduction shown in Section~\ref{sec:proof}.

We then use several simulated MuJoCo tasks from the OpenAI Gym benchmark environments for continuous control settings~\cite{brockman2016openai}. These tasks are particularly relevant for robotics domains as the action space directly applies torque to simulated motors in various robotic configurations to learn locomotion behaviours.

We generate three types of noise to corrupt the reward system with, or to introduce stochasticity into the reward which may be relevant for robotic domains: Gaussian noise, $\epsilon$-likelihood replacement of the reward with uniform noise, and randomly induced sparsity. First, we add a Gaussian noise to the system - which may occur with sensory noise in robotic systems or with a reward provided from a distribution (as in with a data-driven distributional reward). Next, we investigate uniform replacement of the reward with a random reward which could correspond to misspecification or sensory noise. Finally, we investigate artificially and randomly induced sparsity in the reward signal which could be the case if there is a human in the loop providing a reward signal to the robot such that the human may or may not consider providing a reward at some timestep depending on the teacher as in~\cite{thomaz2006reinforcement}. Alternatively, the stochastically sparse reward case could correspond to delayed rewards provided to the robot as in~\cite{campbell2015handling}.

Finally, we extend these experiments to Atari games to show the benefits in discrete settings as well. In both the Atari and MuJoCo domains we use a neural network function approximator for $\hat{R}$ to match the value function and policy networks in the baselines. We also run several experiment seeds for all settings as indicated by~\citet{henderson2017rlmatters}: 10 for MuJoCo tasks and 3 for Atari domains.

\subsection{Tabular Experiments}

\begin{figure}[!htbp]
    \centering
    \includegraphics[width=.32\textwidth]{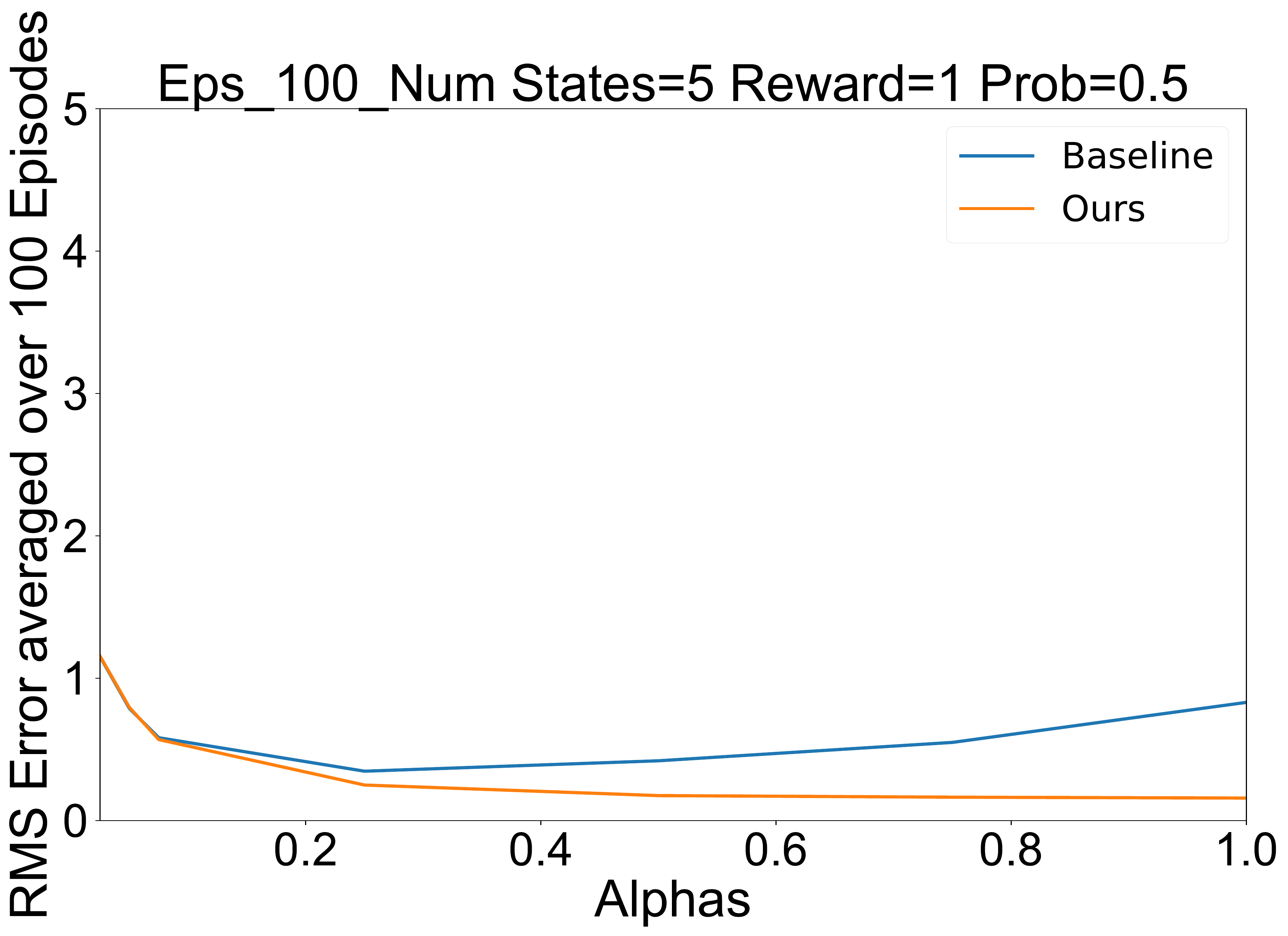}
    \includegraphics[width=.32\textwidth]{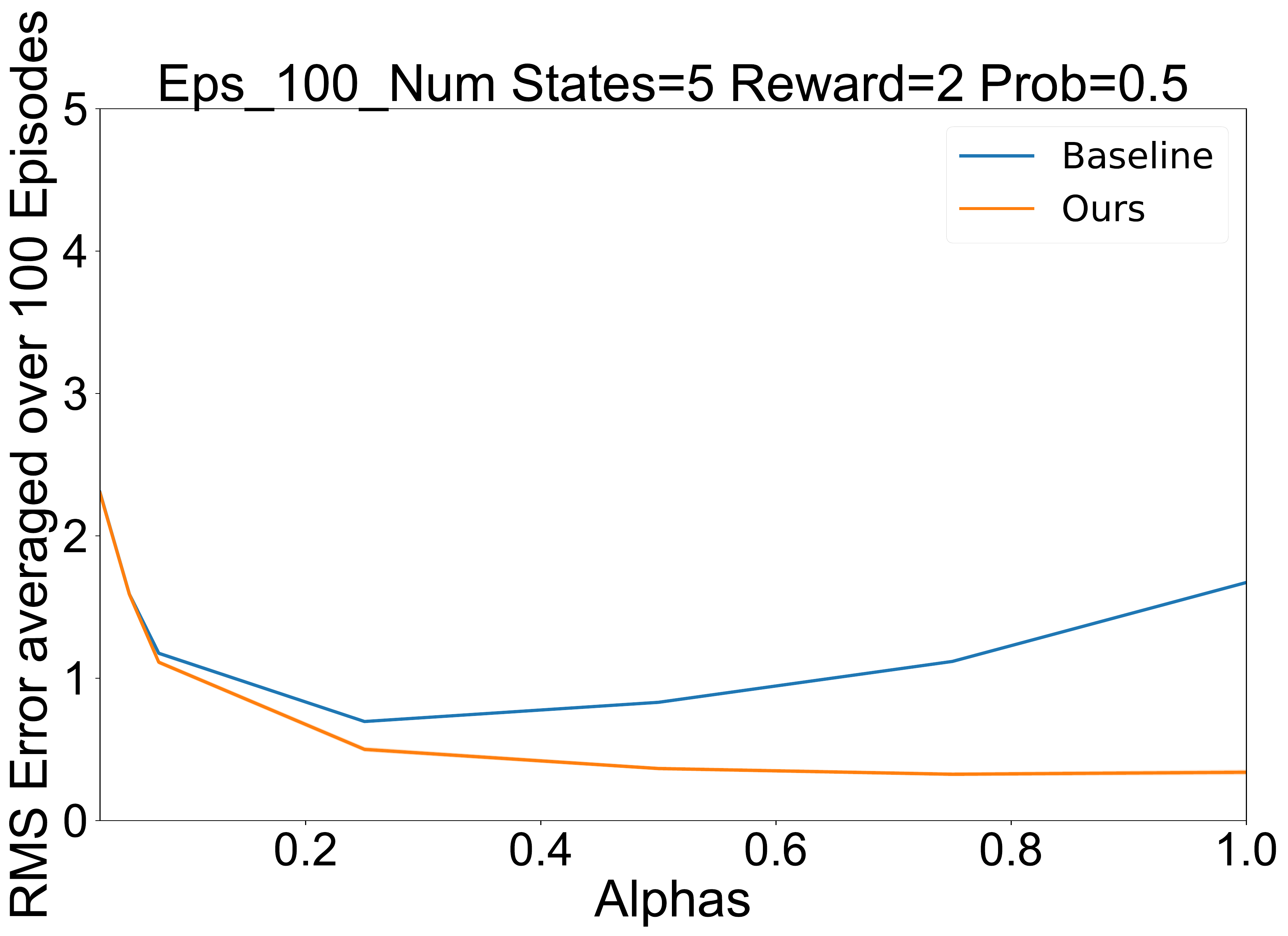}
    \includegraphics[width=.32\textwidth]{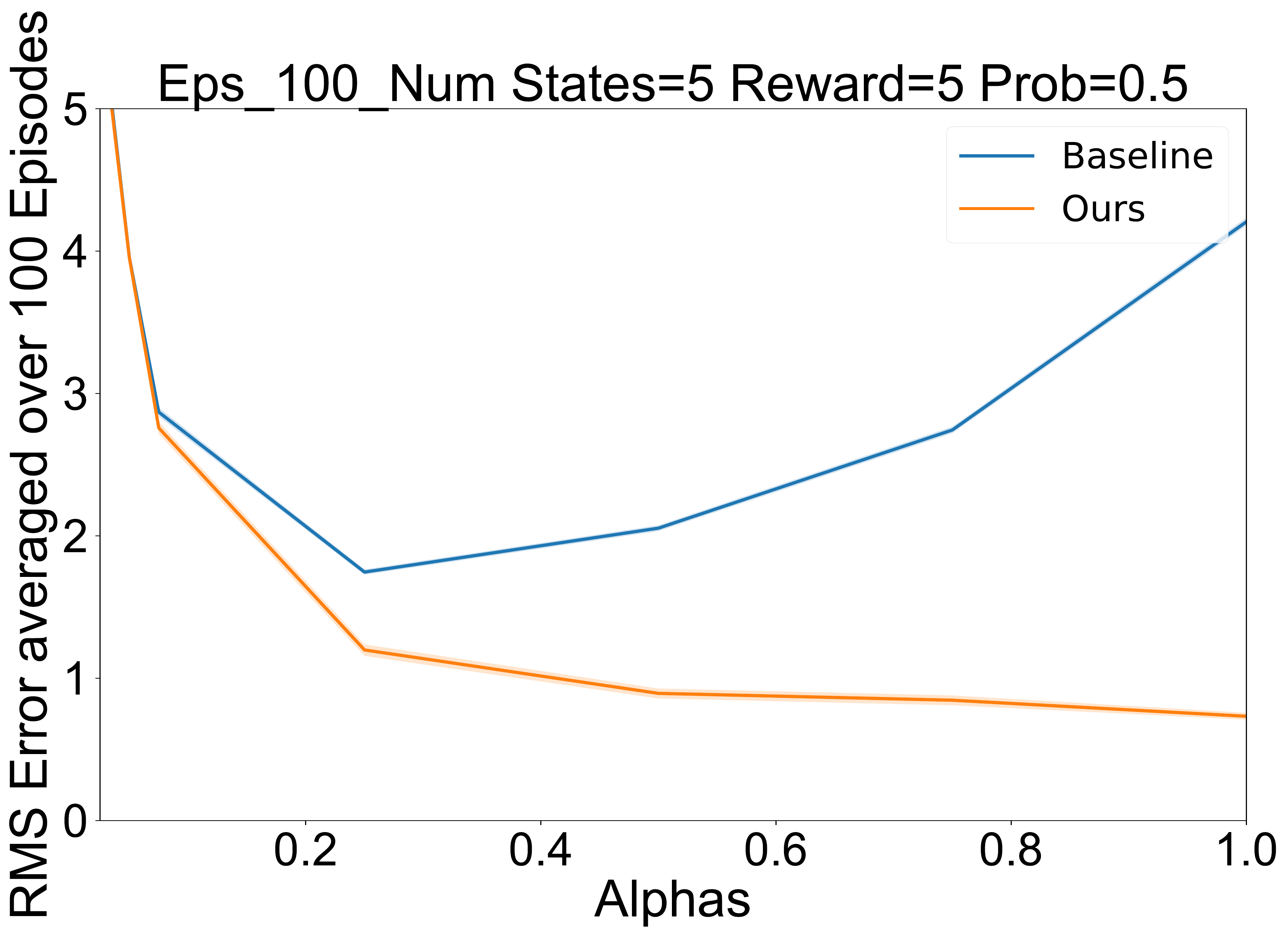}
    \caption{Tabular experiments with a 5-state MDP. In all cases, rewards are assigned with probability $0.5$ and, set to $0$ otherwise (rewards of $+1, +2, +5$, from left to right). The x-axis demonstrates various learning rates for the TD-update. We report the average RMSE over the first 100 episodes of learning - lower is better.
    }
    \label{fig:tabular_results}
\end{figure}

We first empirically investigate the tabular case. We construct a 5 state MDP, as seen in Figure~\ref{fig:fivestate}, for value learning (an extended 10 state MDP can be seen in Appendix~\ref{app:tab}). The MDPs contain deterministic transitions from left to right in the states, and the agent follows a fixed policy moving to the right and terminates on reaching the farthest state to the right. At each state it receives a stochastic reward of $1$, $2$, or $5$ with a fixed probability of $0.5$. The value function is updated via the TD error for $100$ episodes. We measure the robustness to variance by evaluating the root mean squared error (RMSE) of the value function across the $100$ episodes. As is seen in Figure~\ref{fig:tabular_results}, when using the reward estimator, the agent is able to learn more accurate representations of the value function even at high learning rates. This aligns with the aforementioned theoretical variance reduction in Section~\ref{sec:proof}.

\begin{figure}[!htbp]
\label{fig:fivestate}
\centering
\begin{tikzpicture}
\node[state]                               (0) {0};
\coordinate[draw=none,right=of 0]          (0-g);
\node[state,right=of 0-g]                    (1) {1};
\coordinate[draw=none,right=of 1]          (1-g);
\node[state,right=of 1-g]                    (2) {2};
\coordinate[draw=none,right=of 2]          (2-g);
\node[state,right=of {2-g},text depth=0pt] (g) {g};

\draw[
    >=latex,
    auto=right,                      %
    loop above/.style={out=75,in=105,loop},
    every loop,
    ]
     (2)   edge             node {$+1:0, P(.5)$}   (g)
     (1)   edge             node {$+1:0, P(.5)$}   (2)
     (0)   edge             node {$+1:0, P(.5)$} (1);
\end{tikzpicture}
    \caption{Illustration of the sample Markov Decision Process (MDP) used for the tabular case experiments.}
    \label{fig:samplemdp}
\end{figure}
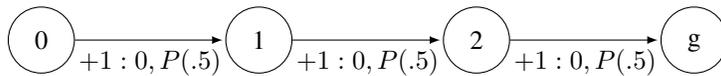

\subsection{MuJoCo Experiments}

We experiment on four different continuous control tasks in the MuJoCo simulator as provided by OpenAI Gym~\cite{brockman2016openai}: Reacher, Hopper, HalfCheetah, and Walker2d. In all cases we use the $-v2$ version of the MuJoCo benchmarks. We use the PyTorch PPO implementation with a clipping objective found in~\cite{pytorchrl} for the baseline, with modifications for reward estimation built directly on top of this. Further information about experimental setup can be found in Appendix~\ref{app:mujoco}. We compare against two baselines: the regular implementation of PPO using the sampled reward and an augmented value function network which trains on an additional auxiliary task to predict the reward as well as the value function (such that hidden layers encode information for both tasks) similarly to~\cite{jaderberg2017unreal}.

\subsubsection{Gaussian Noise}

The first type of reward noise which we add is a Gaussian noise centered around 0 with increasing variance. This noise is inspired by a reward signal which is sensory based, but where sensors exhibit a Gaussian noisy distribution as in~\citep{nguyen2012modeling}. That is, the resulting reward becomes $r_t^{new} = r_t + \psi$ where $\Psi \sim \mathcal{N}(0,\,\sigma^{2})$. The relative normalized baseline improvement of this experiment can be found in Table~\ref{mujoco-gaussian-table} with extended information in the Appendix. As can be seen under a zero-centered Gaussian noise, reward estimation improves results over the baseline in all cases except when no noise is added.

\begin{table}[H]
\centering

\label{mujoco-gaussian-table}
\small{
\begin{tabular}{|c|c|c|c|c|c|}
\hline
   & $\sigma=0.0$  & $\sigma=0.1$  & $\sigma=0.2$ & $\sigma=0.3$ & $\sigma=0.4$ \\
& (\% Gain) & (\% Gain) & (\% Gain) & (\% Gain) & (\% Gain)\\
\hline 
\hline 
Hopper     & -8.09 &\textbf{4.05} & \textbf{6.15} & \textbf{10.39}  & \textbf{33.42} \\ \hline
Walker     &   -8.09 & \textbf{63.67}&  \textbf{159.03}& \textbf{177.59} & \textbf{150.60}\\ \hline
Reacher     &   -1.79 & \textbf{10.41}  & \textbf{16.60} & \textbf{30.72} &  \textbf{24.73}\\ \hline
HalfCheetah &  -12.55 & \textbf{38.70}  & \textbf{115.21} & \textbf{139.52} & \textbf{493.61}\\ \hline
 \hline 
\hline 
Average       & -7.63   & \textbf{29.21}	&  \textbf{74.25}&    \textbf{89.55}&   \textbf{175.59} \\ 
\hline      
\end{tabular}
}

\caption{Gaussian reward noise ($\sigma=(0.0, 0.1, 0.2, 0.3, 0.4)$) comparison between our approach and the best of both baselines (PPO and PPO with the reward prediction auxiliary task). The score represents the relative improvement over the best baseline normalized with respect to the the average episode reward over the last 100 Episodes after training for 1M steps: $\frac{\text{Ours} - \text{Best Baseline}}{|\text{Best Baseline} - \text{Random Policy}|}$. Bold scores indicate an improvement over both baselines. The results are the average over $10$ runs using different random seeds. }
\end{table}

\subsubsection{Uniform Noise}

For the uniform noise experiments, we randomly replace the reward with an $\epsilon$ probability by a uniform reward between $-1$ and $1$. That is:
\begin{equation}
    r^{new}_t  = \begin{cases}
    \psi, & \text{with probability $\epsilon$}\\
    r_t, & \text{with probability $(1-\epsilon)$},
    \end{cases}
\end{equation}
where $\Psi \sim U(-1,1)$ and where $0 \leq \epsilon \leq 1$ is the probability of replacing the current reward. Table~\ref{mujoco-uniform-table} demonstrates the results of this type of added noise. Once again, using reward estimation increases results greatly with added noise up to a relative 500\% average gain in extremely noise scenarios (where the reward is replaced by a uniformly random reward 40\% of the time).

\begin{table}[H]
\centering

\label{mujoco-uniform-table}
\small{
\begin{tabular}{|c|c|c|c|c|c|}
\hline
   & $\epsilon=0.0$  & $\epsilon=0.1$  & $\epsilon=0.2$ & $\epsilon=0.3$ & $\epsilon=0.4$ \\
& (\% Gain) & (\% Gain) & (\% Gain) & (\% Gain) & (\% Gain)\\
\hline 
\hline 
Hopper     & -8.09 & \textbf{28.76} & \textbf{50.43} & \textbf{20.74}  & \textbf{110.45} \\ \hline
Walker      &  -8.09 & \textbf{180.86} & \textbf{105.78} & \textbf{125.84}  & \textbf{34.58} \\ \hline
Reacher          &  -1.79&  \textbf{15.60} & \textbf{24.62}  & \textbf{32.61}  & \textbf{40.31}  \\ \hline
HalfCheetah         & -12.55 & \textbf{110.99}  & \textbf{212.74} & \textbf{555.61}  & \textbf{2044.25} \\ \hline
 \hline 
\hline 
Average       & -7.63  & \textbf{84.05}	& \textbf{98.39}  & \textbf{183.70}   & \textbf{557.40}     \\ \hline             
\end{tabular}
}
\caption{Uniform reward noise ($\epsilon=(0.0, 0.1, 0.2, 0.3, 0.4)$) comparison between our approach to the best of both baselines (PPO and PPO with the reward prediction auxiliary task). The score represents the relative improvement as in Figure~\ref{mujoco-gaussian-table}. The results are the average over $10$ runs using different random seeds. }
\end{table}

\subsubsection{Sparsity}

We consider artificially making the reward sparser by replacing the true environmental reward with the zero reward with varying levels of probability. Specifically, the reward at time $t$ is defined as:
\begin{equation}
    r^{new}_t  = \begin{cases}
    0, & \text{with probability $\epsilon$}\\
    r_t, & \text{with probability $(1-\epsilon)$},
    \end{cases}
\end{equation}

This may reflect a scenario where there is a signal dropout either in a sensory-based reward signal as in \cite{potter2010sparsity} or in communication of the reward signal. This type of noise in particular provides insight into the robustness of $\hat{R}$ estimation to sparse rewards, 
while still preserving the optimal ordering of policies -- where the optimal ordering indicates that if $Q(s,a) > Q(s,a')$ under $R(s,a,s')$ then $Q'(s,a) > Q'(s,a')$ under $R(s,a,s') * c$ where $c > 0$. The latter is true because sparsity noise can be seen as simply multiplying the reward signal by a constant positive factor at every time step. Specifically, where $0 \leq \epsilon \leq 1$ is the probability of receiving reward $0$:
\begin{equation}
    E[r^{new}| s,a,s'] = (1-\epsilon) E[r | s,a,s'].
\end{equation}

As can be seen in Table~\ref{mujoco-sparse-table}, the reward estimation method does not always improve results in this case, except under extreme sparsity. This is likely due to the need to correct reward learning for distribution imbalance. It may be possible, since the baseline implementation uses Adam~\cite{kingma2014adam} for the optimization method, that the reward estimator was unable to learn an improved representation under sparsity for convergence properties presented in~\cite{amsgrad}, while a value function update would not encounter the sparsity issue as frequently. Nonetheless, in certain domains (Hopper and Walker) we still see improved performance.

\begin{table}[H]
\centering

\label{mujoco-sparse-table}
\small{
\begin{tabular}{|c|c|c|c|c|c|}
\hline
   & $\epsilon=0.6$  & $\epsilon=0.7$  & $\epsilon=0.8$ & $\epsilon=0.9$ & $\epsilon=0.95$ \\
& (\% Gain) & (\% Gain) & (\% Gain) & (\% Gain) & (\% Gain)\\
\hline 
\hline 
Hopper     & \textbf{16.31} & -8.0 & \textbf{2.00} & \textbf{72.54}  & \textbf{81.93} \\ \hline
Walker      & \textbf{6.19}  & \textbf{17.54} & \textbf{32.18}  & \textbf{205.12} & \textbf{130.63} \\ \hline
Reacher     &  -9.98 & -16.35 & -18.32 & -34.69 & \textbf{83.29} \\ \hline
HalfCheetah & -12.4 & 14.5  & -0.67 & -6.01 & \textbf{124.15} \\ \hline
 \hline 
\hline 
Average       & \textbf{0.03} &  -5.33 & \textbf{3.81}	& \textbf{59.24} & \textbf{105}        \\ \hline             
\end{tabular}}
\caption{Sparse reward noise ($\epsilon=(0.6, 0.7, 0.8, 0.9, 0.95)$) comparison between our approach to the best of both baselines (PPO and PPO with the reward prediction auxiliary task). The score represents the relative improvement as in Figure~\ref{mujoco-gaussian-table}. The results are the averaged over $10$ different experiment random seeds. }
\end{table}

\subsubsection{Analyzing the Empirical Advantage}

\begin{figure}[!htbp]
    \centering
    \includegraphics[width=.32\textwidth]{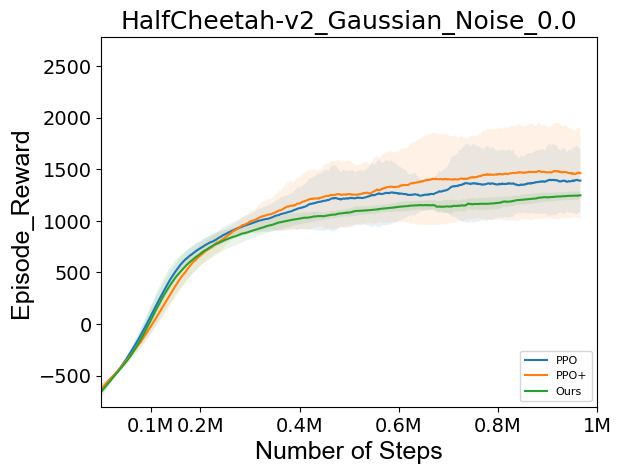}
    \includegraphics[width=.32\textwidth]{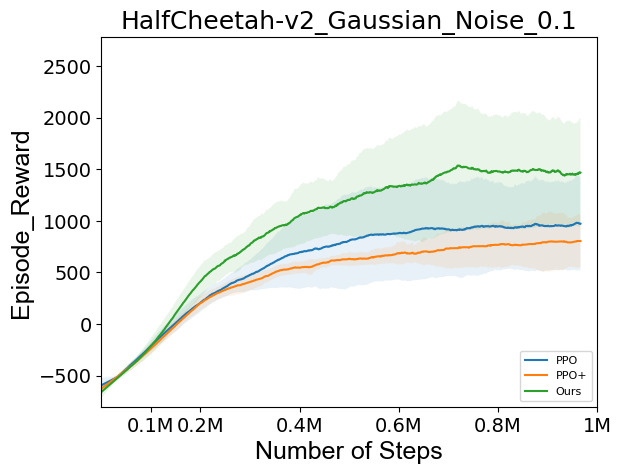}
    \includegraphics[width=.32\textwidth]{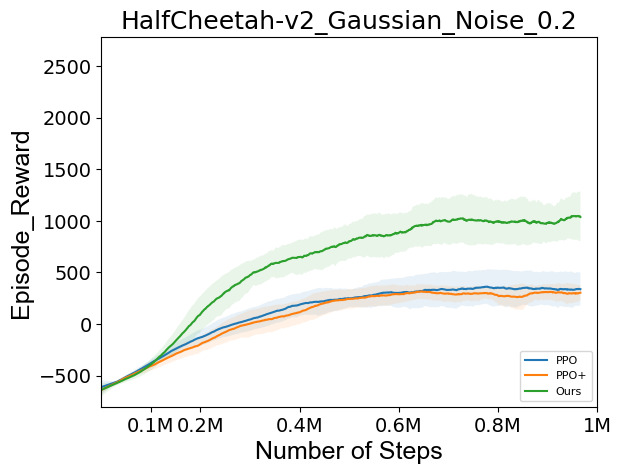}

    \includegraphics[width=.32\textwidth]{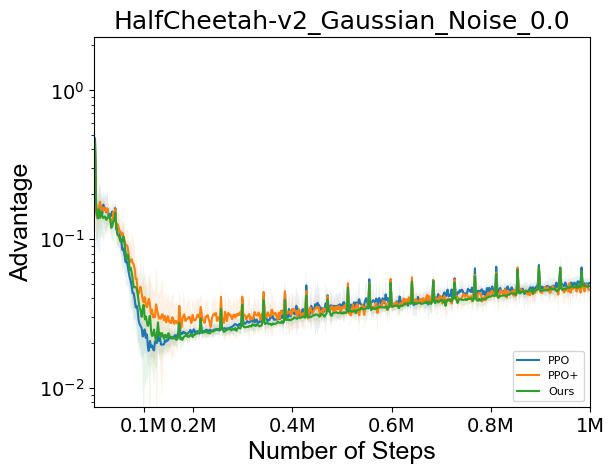}
    \includegraphics[width=.32\textwidth]{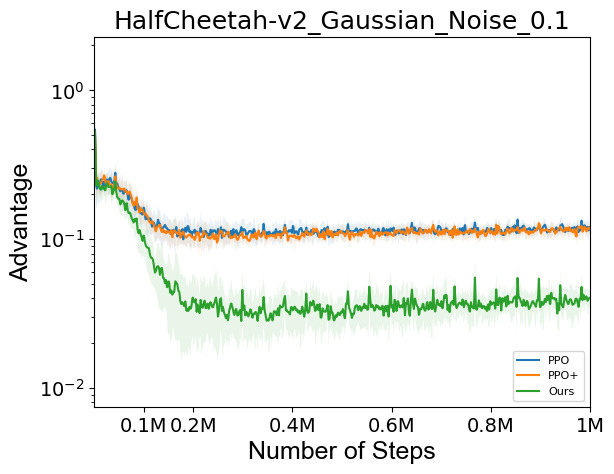}
    \includegraphics[width=.32\textwidth]{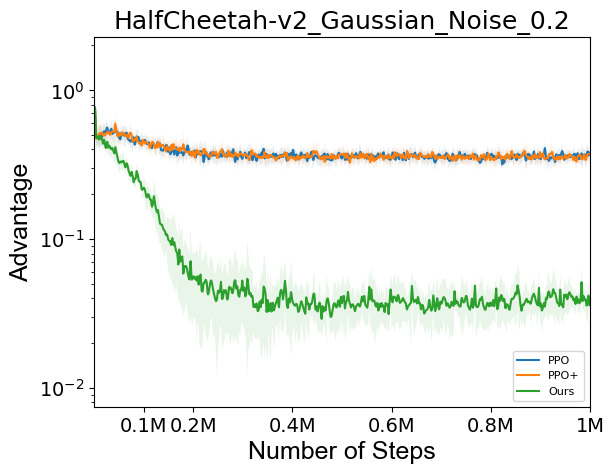}

    \caption{The recorded emperical advantage and performance of PPO on the HalfCheetah environment under Gaussian corruption of the reward.}
    \label{fig:advantage}
\end{figure}

We next attempt to uncover the exact source of the improved performances by analyzing the empirical advantage and TD-error. 
As mentioned in Section~\ref{sec:background}, the advantage function provides a signal for both the critic and the policy to use as part of their objective in actor-critic methods. For the policy, actions that lead to positive advantages are reinforced, while negative advantages result in a negative likelihood update for those actions. A high expected squared advantage - TD error - implies a large residual error that is not captured by the value function. As we can see in Figure~\ref{fig:advantage}, the reduction of this advantage (and thus TD error) directly correlates to performance gains in the episode return.

\subsubsection{Analyzing Reduction of Variance}
\label{sec:varianceexp}
To determine if our theoretical variance reduction properties found in Section~\ref{sec:proof} hold in the continuous control case with neural network function approximators, we empirically measure the variance of the Bellman operator itself with the corrupted stochastic rewards. With full experimental details and extended results in the Appendix, we find that across all environments, we see an average of 59.7\%, 76.7\%, and 62.5\% absolute reduction in variance under all tested levels of Gaussian, uniform, and sparsity inducing noise, respectively.

\subsection{Atari Experiments}

We extend our methodology in the Atari domain to demonstrate the extension to discrete action spaces (for example if a robot can only select macro actions). The experimental setup, details, and full results for this can be found in Appendix~\ref{app:atari}. Table~\ref{atari-table} shows a sampling of results under Gaussian noise where once again we see that using an estimator $\hat{R}$ improves performance under corrupted stochastic reward signals in most cases.

\begin{table}[!htbp]
\centering

\label{atari-table}
\small{
\begin{tabular}{|c|c|c|c|c|c|}
\hline
   & No Noise & Noise1  &Noise2 & Noise3 & Noise4 \\
\hline 
\hline 
Average Improvement (Gaussian)       & -8.69  & \textbf{12.6}	& \textbf{262.7} & \textbf{684.73}  & \textbf{632.82}    \\ \hline   
Average Improvement (Uniform)     & -8.69  & \textbf{20.86}  & \textbf{108.15}	& \textbf{285.59} & --   \\ \hline             
Average Improvement (Sparse)    & -8.69   & \textbf{9.95}  & \textbf{21.26} & \textbf{43.40} & --    \\ \hline
\end{tabular}
}
\caption{The relative average percentage improvement across 5 Atari games of the $\hat{R}$ estimator using the same metric as in Figure~\ref{mujoco-gaussian-table} under Gaussian noise with different standard deviations. See Appendix~\ref{app:atari} for more details and results. Noise1-4 correspond to $\sigma=0.1, 0.2, 0.3, 0.4$ for Gaussian noise, $\epsilon=0.1, 0.2, 0.3$ for uniform noise, and $\epsilon=0.3, 0.5, 0.75$ for sparsity inducing noise.}
\end{table}

\section{Conclusion}
Our work provides a simple yet effective method for addressing and improving performance under corrupted and stochastic rewards in model-free policy gradient methods. Future extensions of our work may involve learning a distributional reward estimator as in~\citep{DBLP:conf/icml/BellemareDM17}, off-policy experience replay for the reward estimator, or learn options for reward estimators as in~\citep{henderson2017optiongan}. These improvements may help improve the fidelity and accuracy of reward estimation to further improve results in settings such as induced sparsity.

More importantly, as reward generation moves toward data-driven~\cite{hadfield2017inverse,sermanet2016unsupervised} or human-in-the-loop~\cite{thomaz2006reinforcement,christiano2017deep} means, addressing the likely stochasticity or corruption in the reward signal -- particularly through the simple modification of existing methods as we show here -- is vital for successful learning of intended behaviours in complex robotic tasks. We hope that our work provides a simple foundation for which other methods can build on to address corrupted and stochastic rewards.

\appendix

\newpage

\section{Theoretical Proof Extensions}

\subsection{Theoretical Variance Reduction Extended Discussion}
\label{app:singlestepproof}
We note that we make an assumption that in most cases $\operatorname{cov}\left[ r_t,  \gamma V_\pi^\gamma (s_{t+1} ) \right] \geq 0$. 

When conditioned on $s_{t}$, $a_t$ and $'s_{t+1}$, $r_t$ and $r_{t+1}$ are independent variables and thus their covariance equal to zero. %
When the covariance is not zero and we must fall back to another formality. We note that as $N \rightarrow{\infty}$ the left most terms of the right hand side of Equation~\ref{eq:equality} tend to $0$, which gives us the following:
\begin{equation}
    \operatorname{var}\left[ \hat{G}_{t}^\gamma \right] - \operatorname{var}\left[ G_{t}^\gamma \right] =  - \operatorname{var}\left[ r_t \right] - 2\operatorname{cov}\left[ r_t,  \gamma V_\pi^\gamma (s_{t+1} ) \right], 
\end{equation}
which is less than $0$ when:
\begin{equation}
    \operatorname{var}\left[ r_t \right] > - 2\operatorname{cov}\left[ r_t,  \gamma V_\pi^\gamma (s_{t+1} ) \right].
    \label{eq:covinequality}
\end{equation}

Thus, when the variables are not independent and covariance equal to zero, Equation~\ref{eq:covinequality} must be followed for reductions in variance to occur using a reward estimator that is the expectation over the random variable.

\section{Tabular MDP Experiment}

\begin{figure}[H]
    \centering
    \includegraphics[width=.32\textwidth]{s5r1}
    \includegraphics[width=.32\textwidth]{s5r2}
    \includegraphics[width=.32\textwidth]{s5r5}
    \includegraphics[width=.32\textwidth]{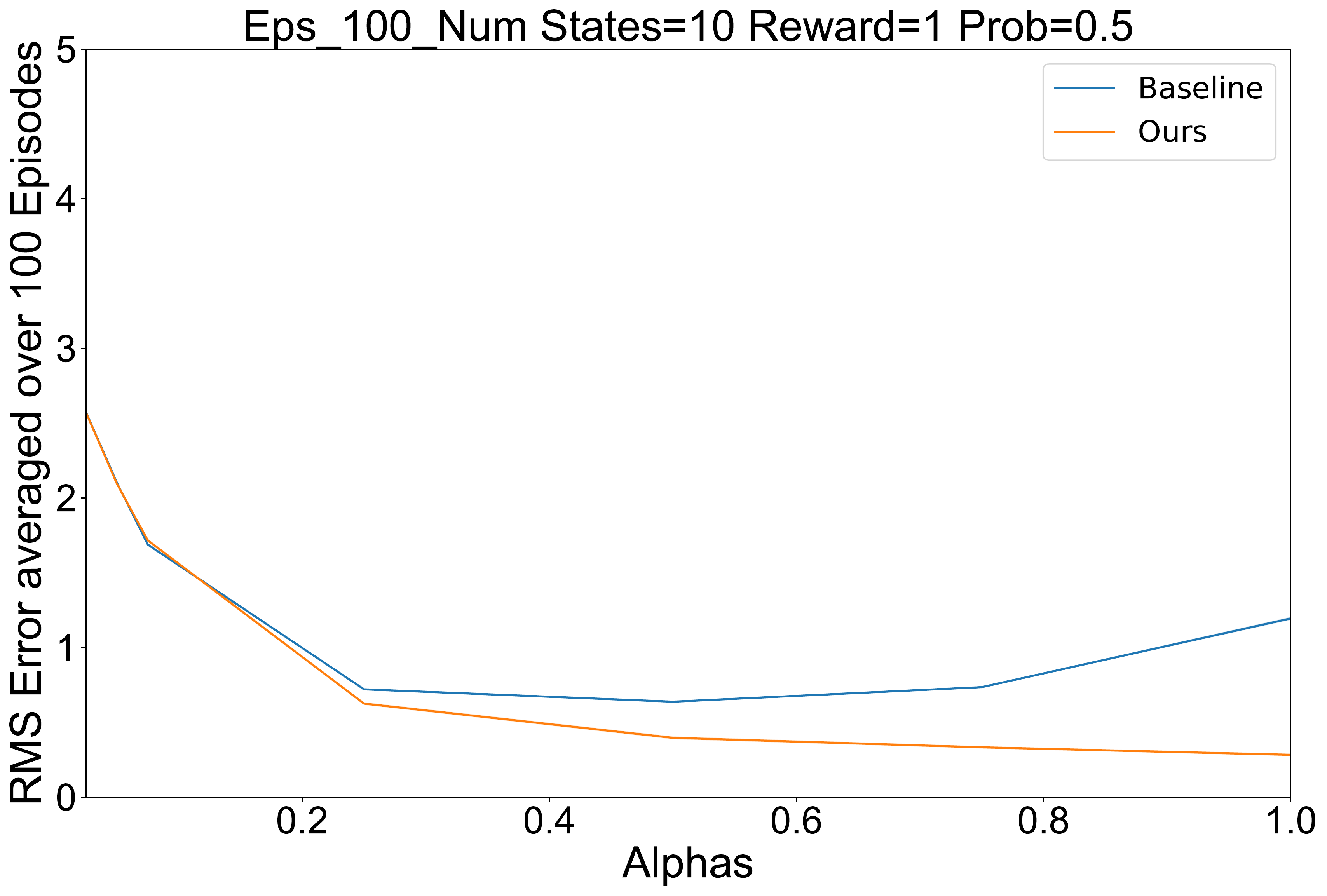}
    \includegraphics[width=.32\textwidth]{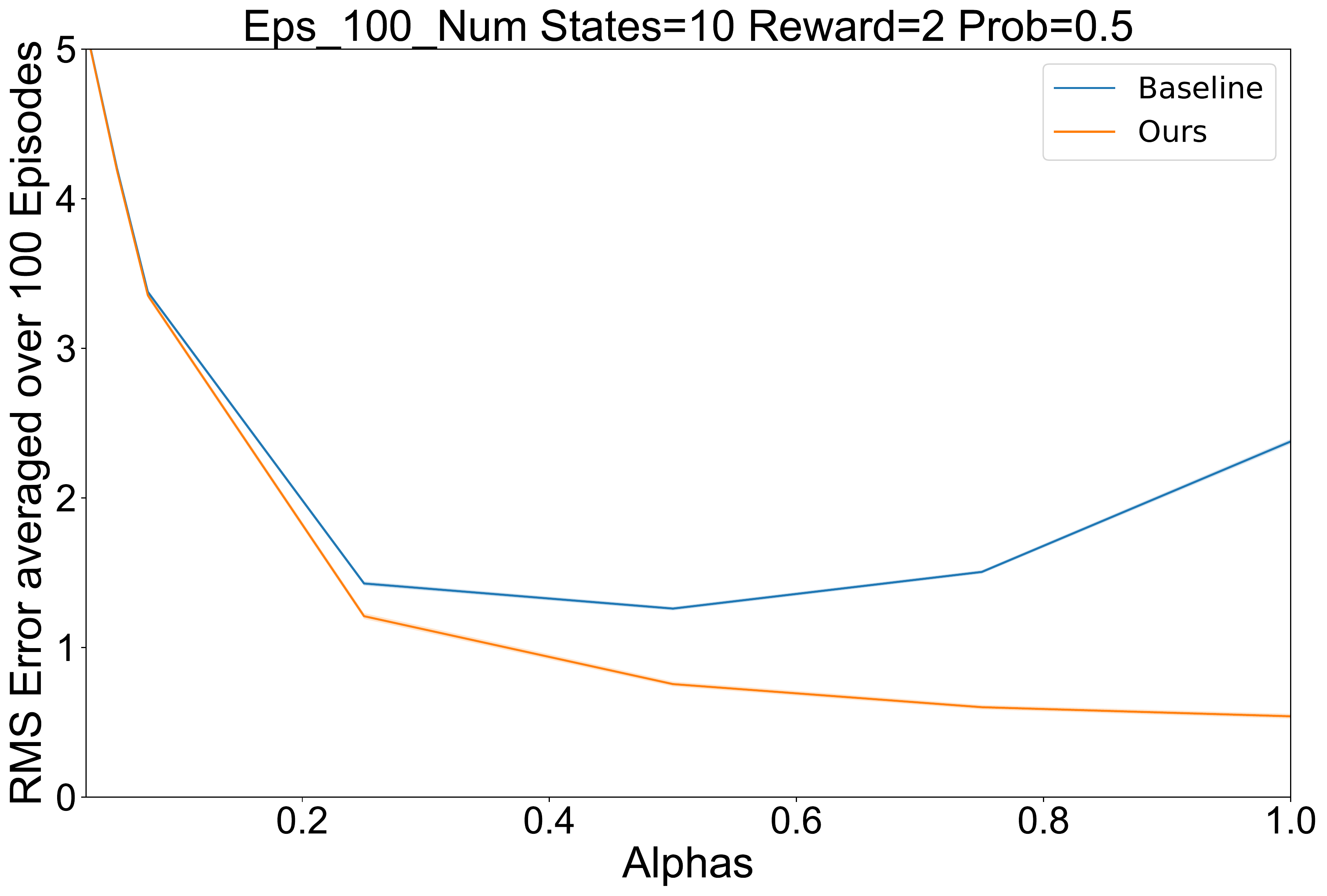}
    \includegraphics[width=.32\textwidth]{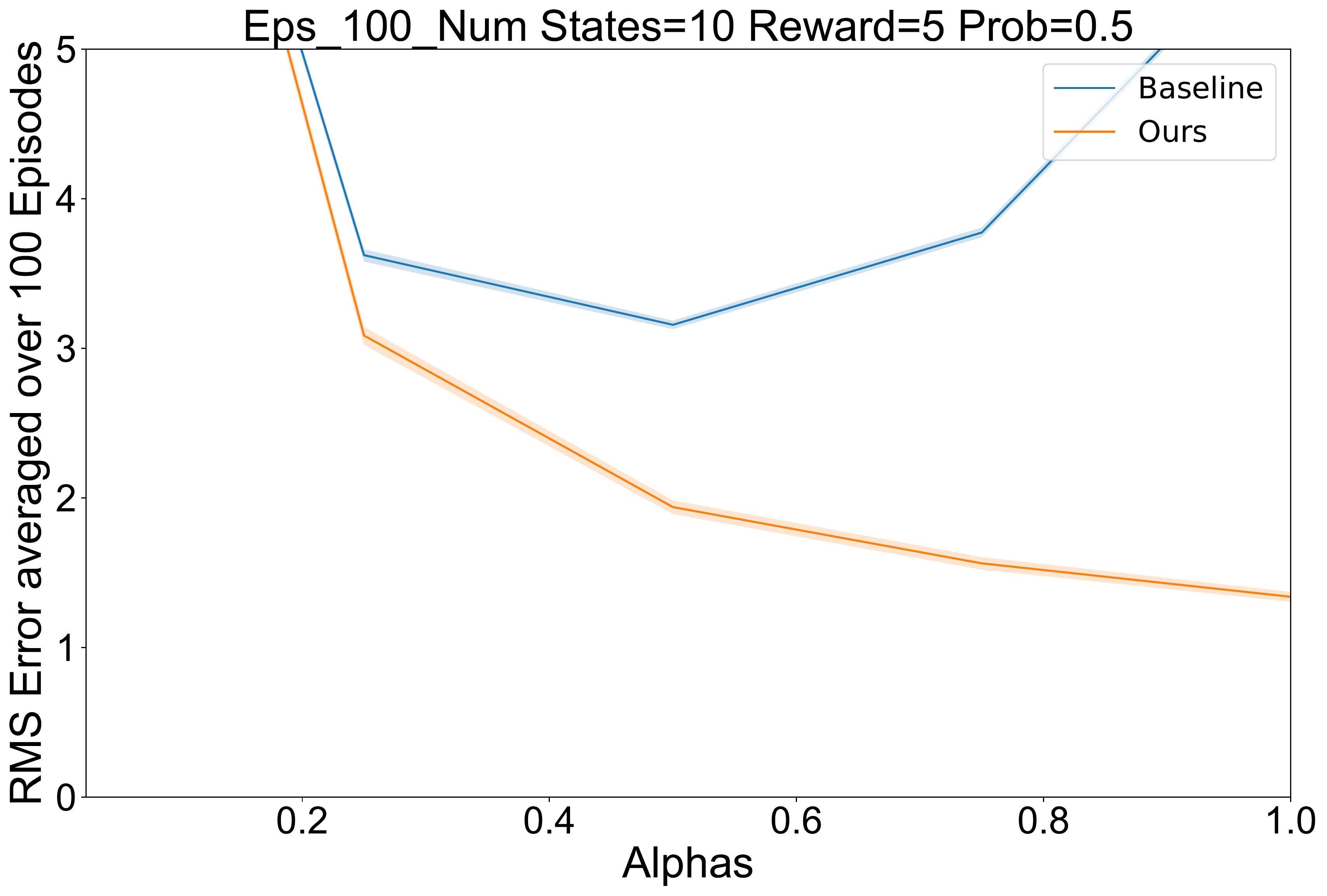}
    \caption{Tabular experiments with a $5$-state MDP (top row) and a $10$-state MDP (bottom row), with varying reward assignments at each states. In all cases, rewards are assigned with probability $0.5$ and, set to $0$ otherwise (rewards of $+1, +2, +5$, from left to right). The x-axis demonstrates various learning rates for the TD-update, as seen in similar variance analysis experiments~\citep{van2009theoretical}. As can be seen, using the sample mean in MDPs with stochastic processes greatly reduces the variance, allowing for higher learning rates to be used.}
    \label{fig:tabular_results_appendix}
\end{figure}

Figure~\ref{fig:tabular_results_appendix} shows the extended results for our tabular experiments. 

\label{app:tab}

\section{Atari Experiments}
\label{app:atari}

\begin{figure}[H]
    \centering
    \includegraphics[width=.19\textwidth]{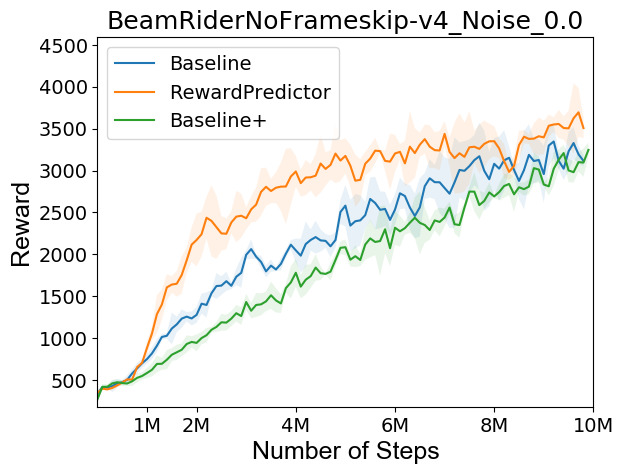}
    \includegraphics[width=.19\textwidth]{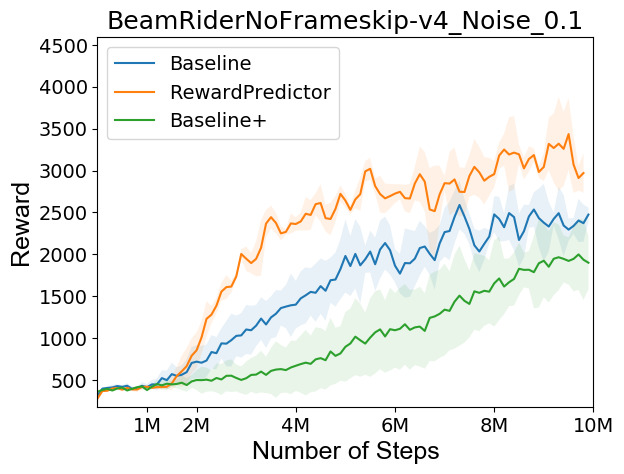}
    \includegraphics[width=.19\textwidth]{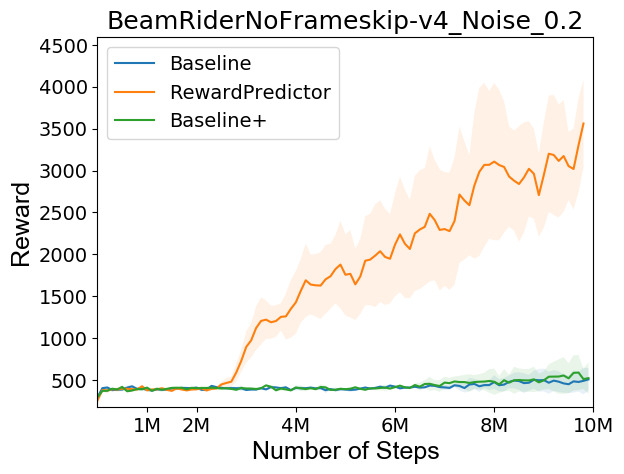}
    \includegraphics[width=.19\textwidth]{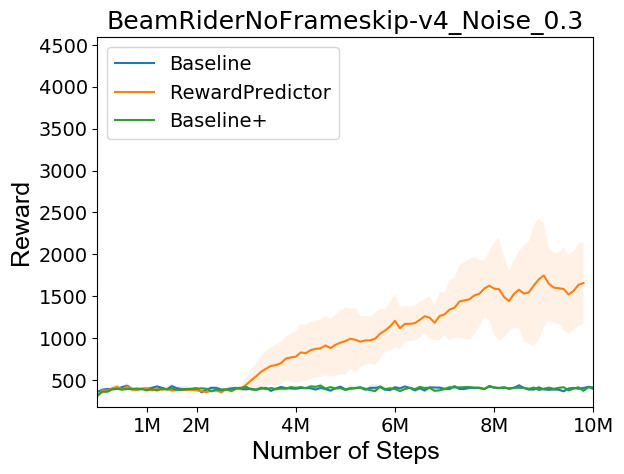}
    \includegraphics[width=.19\textwidth]{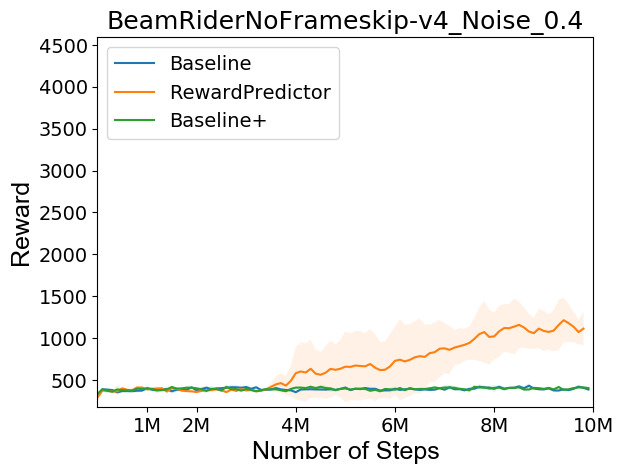}
    
    \includegraphics[width=.19\textwidth]{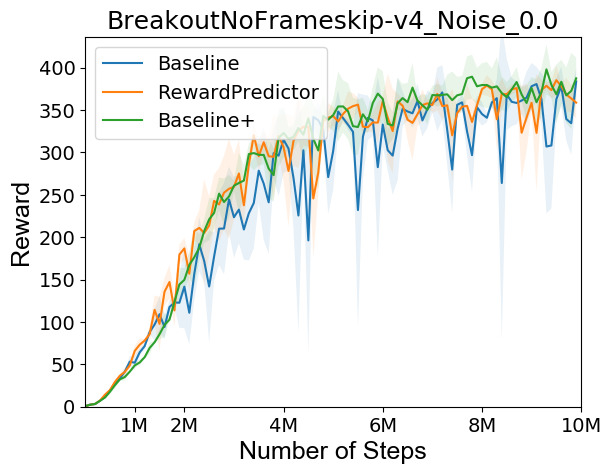}
    \includegraphics[width=.19\textwidth]{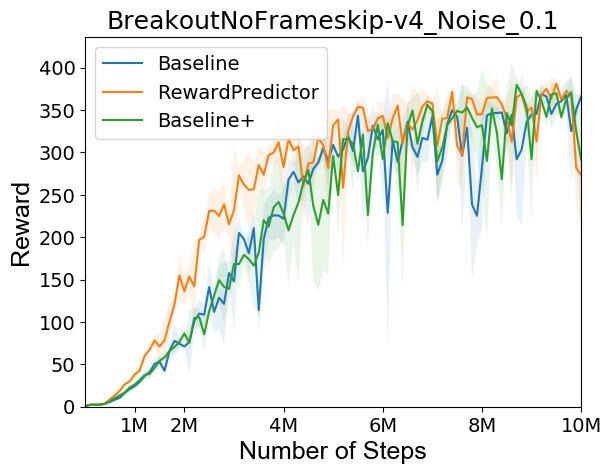}
    \includegraphics[width=.19\textwidth]{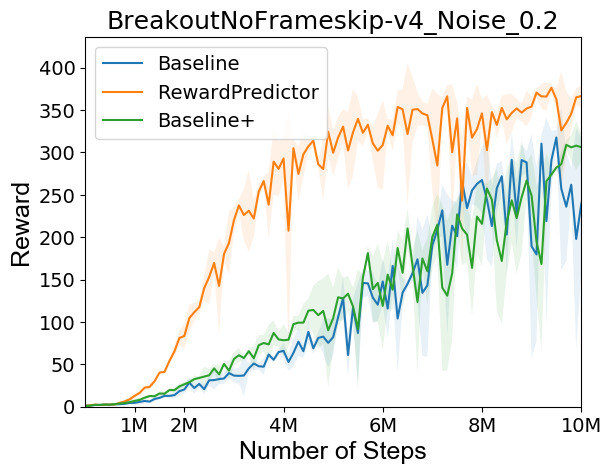}
    \includegraphics[width=.19\textwidth]{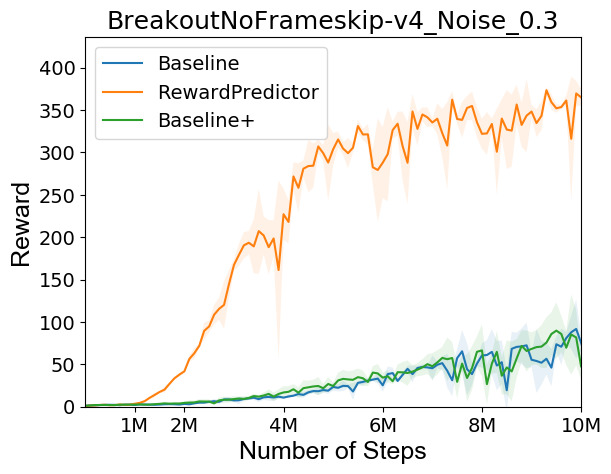}
    \includegraphics[width=.19\textwidth]{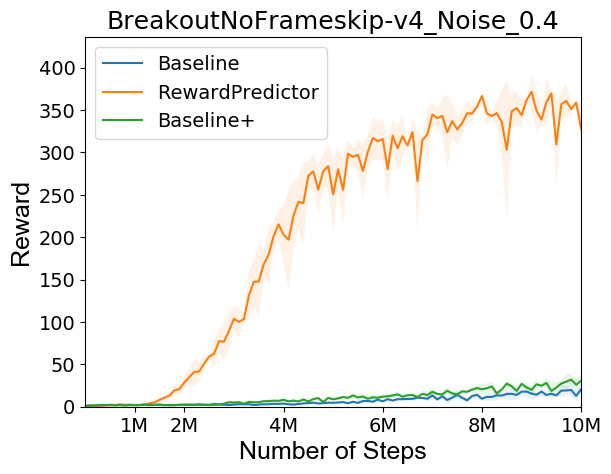}
    
    \includegraphics[width=.19\textwidth]{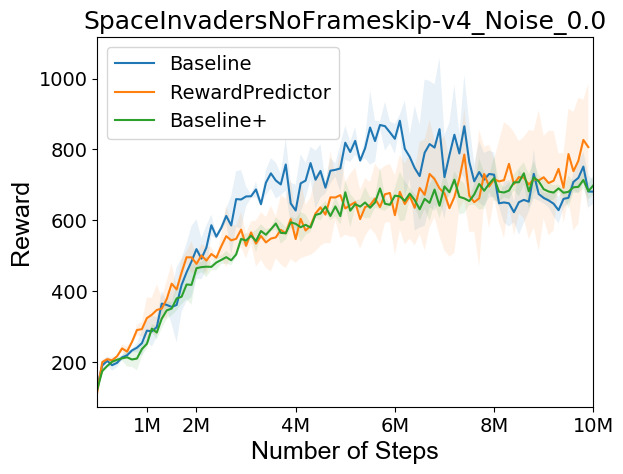}
    \includegraphics[width=.19\textwidth]{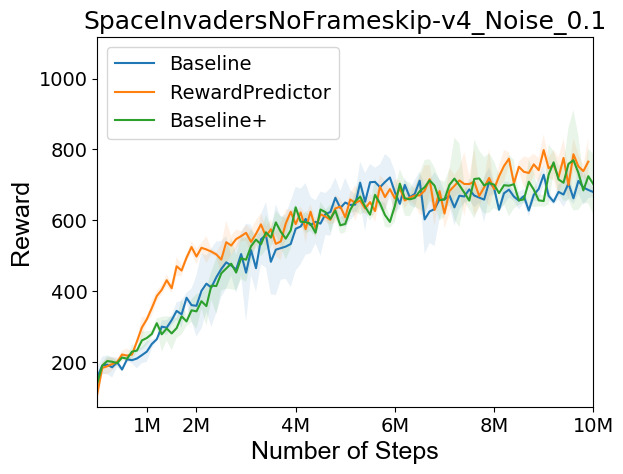}
    \includegraphics[width=.19\textwidth]{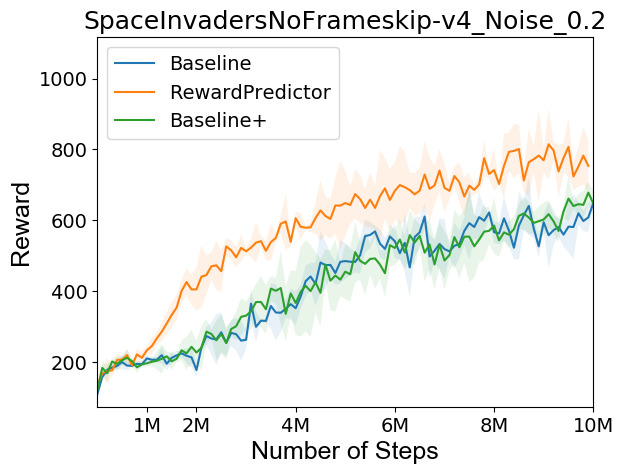}
    \includegraphics[width=.19\textwidth]{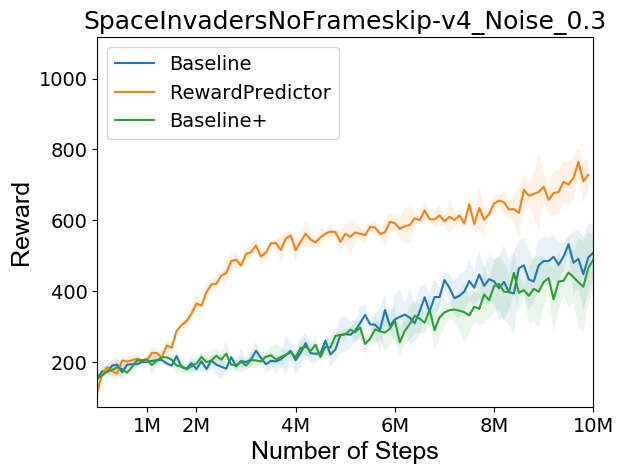}
    \includegraphics[width=.19\textwidth]{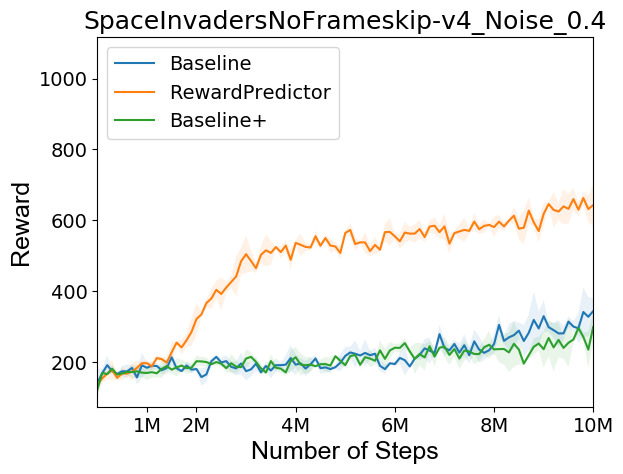}
    
    \includegraphics[width=.19\textwidth]{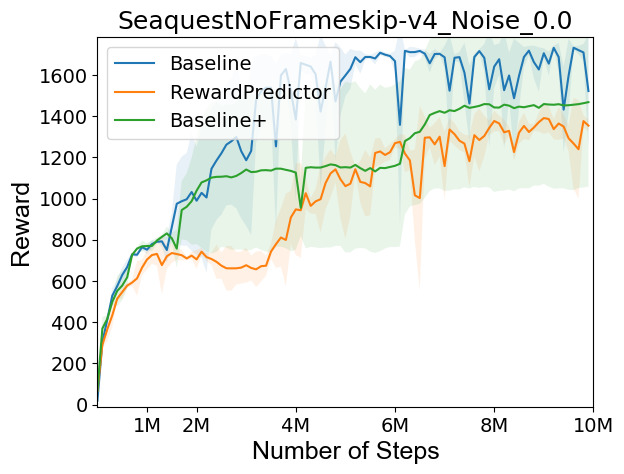}
    \includegraphics[width=.19\textwidth]{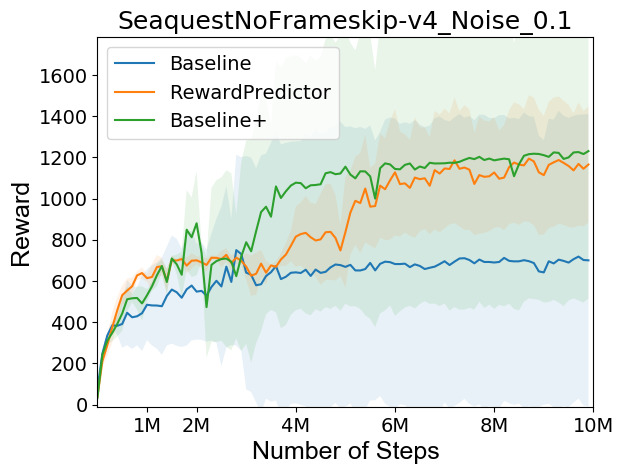}
    \includegraphics[width=.19\textwidth]{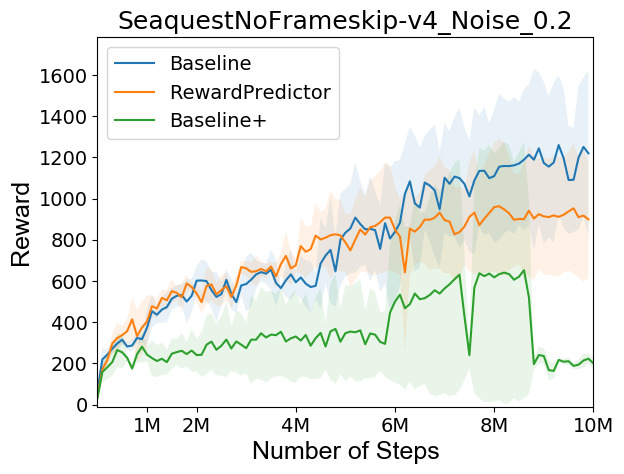}
    \includegraphics[width=.19\textwidth]{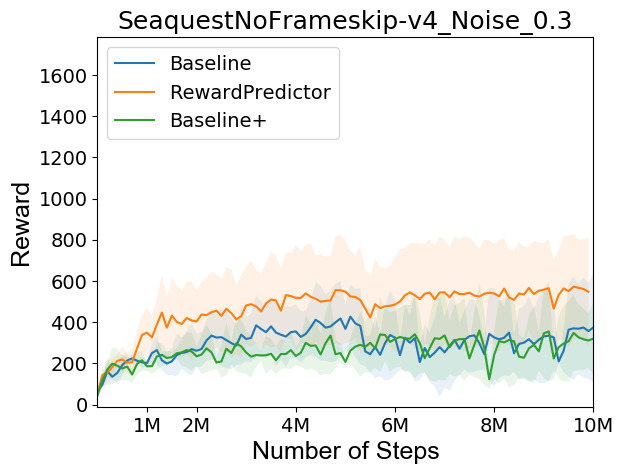}
    \includegraphics[width=.19\textwidth]{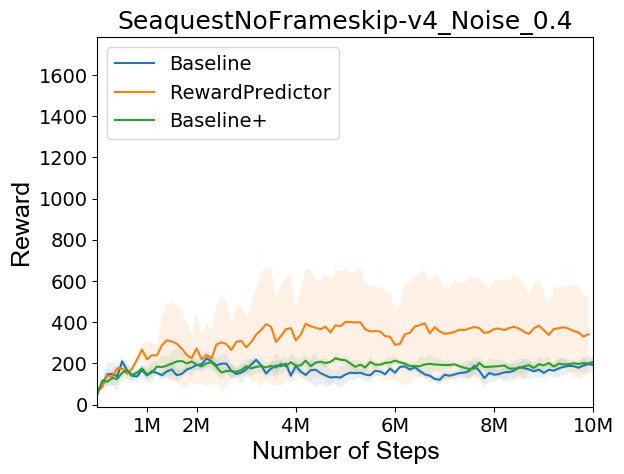}
    
    \includegraphics[width=.19\textwidth]{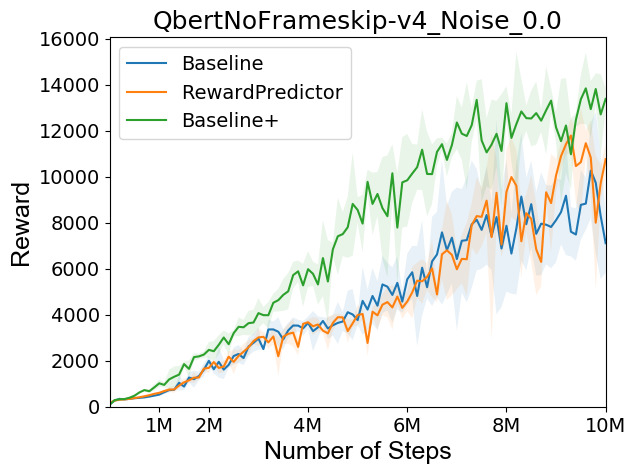}
    \includegraphics[width=.19\textwidth]{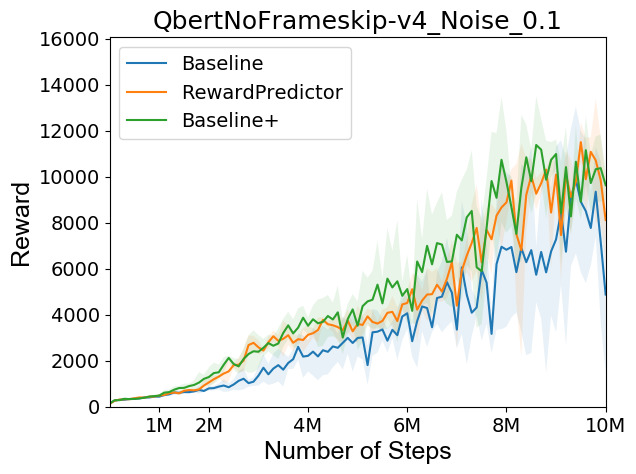}
    \includegraphics[width=.19\textwidth]{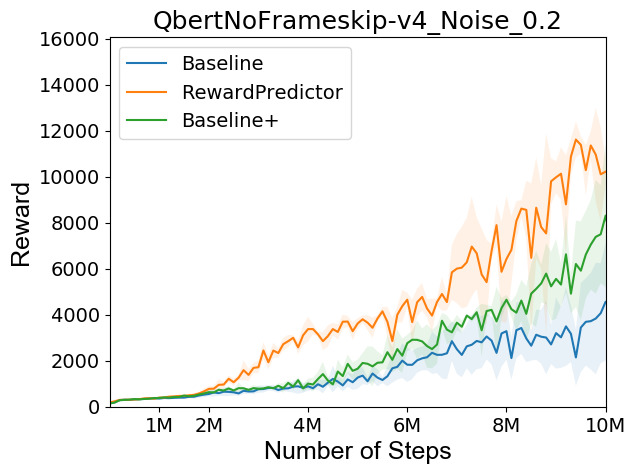}
    \includegraphics[width=.19\textwidth]{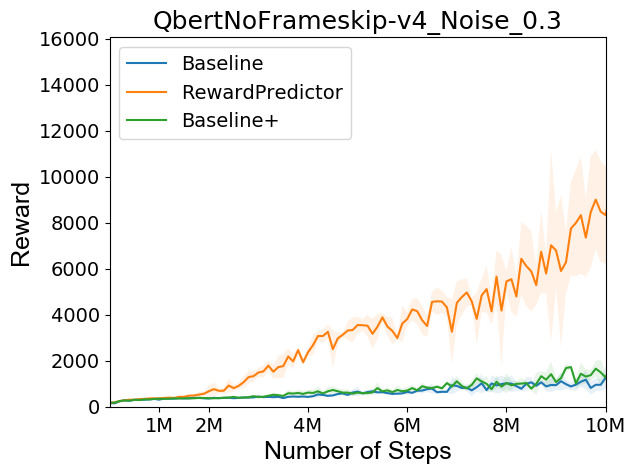}
    \includegraphics[width=.19\textwidth]{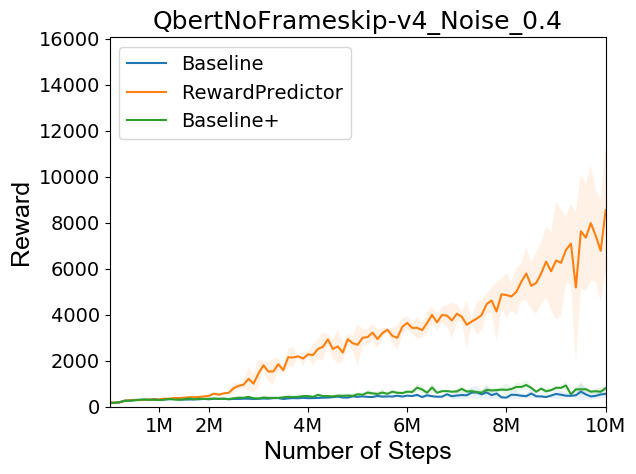}
    
    \includegraphics[width=.19\textwidth]{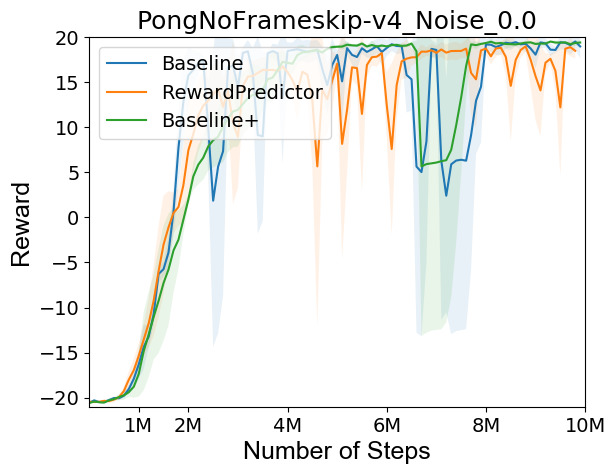}
    \includegraphics[width=.19\textwidth]{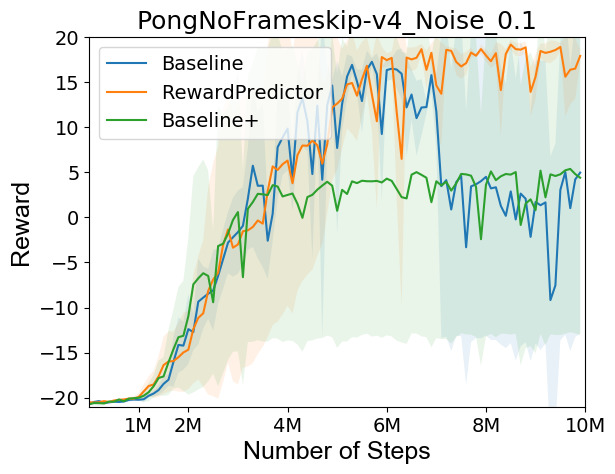}
    \includegraphics[width=.19\textwidth]{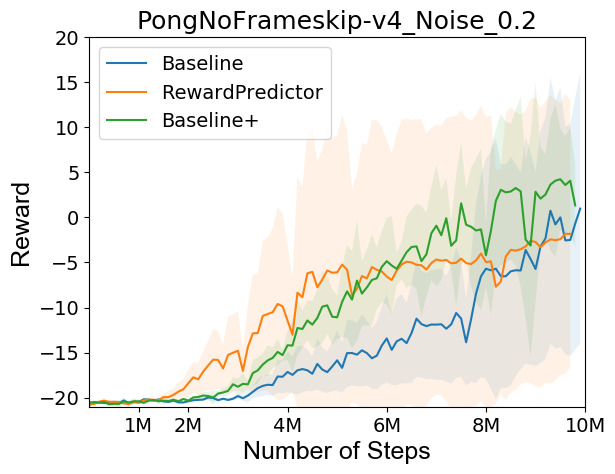}
    \includegraphics[width=.19\textwidth]{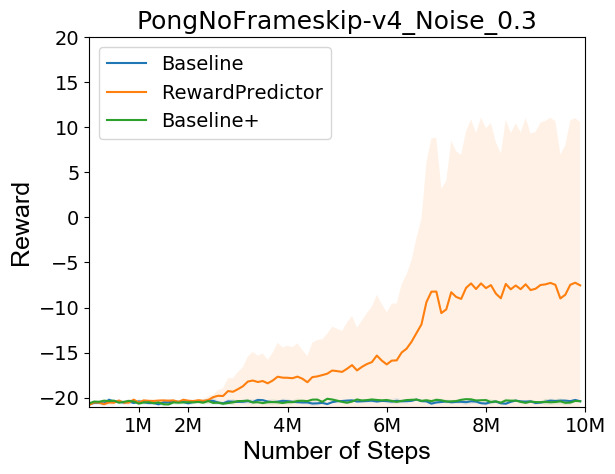}
    \includegraphics[width=.19\textwidth]{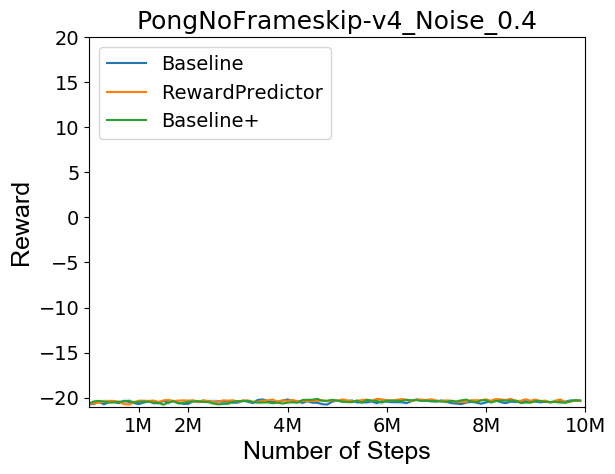}
    
    \caption{Full learning curves for $3$ runs for six Atari Games over $10$M training steps (corresponding with $40$M raw frames). From top to bottom: Beam Rider, Breakout, Space Invaders, Seaquest, Qbert, Pong. We compare five different noise levels that each correspond to adding Gaussian noise centered at zero with the labelled standard-deviation. From left to right we have: $0.0, 0.1, 0.2, 0.3, 0.4$. Our proposed method is labelled "Ours", while A2C and A2C with the reward prediction auxiliary task are labelled Baseline and Baseline+ respectively.}
    \label{fig:atari_results}
\end{figure}

For evaluating our method with function approximation, we use 5 Atari games from the Arcade Learning Environment (ALE)~\citep{bellemare13arcade}. We use the exact same hyperparameters used in OpenAI's Baselines implementation \citep{baselines} and modify the PyTorch A2C implementation for our codebase~\cite{pytorchrl}, but use an additional network (with the same architecture) as a reward predictor and use it to train our critic as described in Section~\ref{method}. We compare our approach to the standard A2C algorithm, as well as A2C with reward prediction as an auxiliary task, similar to \cite{jaderberg2017unreal}. We report results, averaged over $3$ random seeds, for rewards with varying levels of Gaussian noise in Table \ref{atari-table2}, \ref{tab:atari2} and Figure~\ref{fig:atari_results}. In most settings, we see that our proposed method performs relatively better once noise has been introduced. 

\subsection{Performance of Random Agent}

For the displayed results in the main text we normalize by the performance over a random agent. These performances were averaged over 100 episodes by uniformly sampling from the action space and can be found in Table~\ref{tab:atarirandom},

\begin{table}[H]
    \centering
    \begin{tabular}{|c|c|c|c|c|c|c|}
    \hline
         Game & BeamRider& Breakout & Pong & QBert & Seaquest & Space Invaders  \\
         \hline
         Return & 337 & 1.7 & -20.7 & 163.9 & 68.4 & 148 \\
         \hline
    \end{tabular}
    \caption{Average return of a random uniform sampling policy on Atari games across 100 episodes. }
    \label{tab:atarirandom}
\end{table}

\subsection{Extended Experimental Results}

\begin{table}[H]
\centering

\label{atari-table2}
\small{
\begin{tabular}{|c|c|c|c|c|c|}
\hline
   Environment & $\sigma=0.0$  & $\sigma=0.1$  & $\sigma=0.2$ & $\sigma=0.3$ & $\sigma=0.4$ \\
& (\%  Gain) & (\% Gain) & (\% Gain) & (\% Gain) & (\% Gain)\\
\hline 
\hline 
BeamRider     & \textbf{26.87}  & \textbf{49.45}  & \textbf{1350.95} & \textbf{876.43}  & \textbf{485.13}  \\ \hline
Breakout      & -1.24  & \textbf{15.40}  & \textbf{101.82} & \textbf{681.86}  & \textbf{2152.73} \\ \hline
Pong          & -0.22  & \textbf{21.66}  & -1.55  & \textbf{1882.6} & \textbf{32.05}  \\ \hline
Qbert         & -37.57 & -10.18  & \textbf{78.55} & \textbf{456.57}  & \textbf{646.32} \\ \hline
Seaquest      & -29.53  & -9.18  & -8.68  & \textbf{74.66}  & \textbf{115.86}  \\ \hline
SpaceInvaders & -10.48  & \textbf{8.46} & \textbf{55.10} & \textbf{136.29}  & \textbf{364.82} \\ \hline
 \hline 
\hline 
Average       & -8.69  & \textbf{12.6}	& \textbf{262.7} & \textbf{684.73}  & \textbf{632.82}    \\ \hline             
\end{tabular}
}
\caption{Comparison of the average episode reward over 10M steps of training between our approach to the best of both baselines (A2C and A2C with the reward prediction auxiliary task). The score represents the relative improvement over the best baseline normalized by the performance of the random policy: $\frac{\text{Ours} - \text{Best Baseline}}{|\text{Best Baseline} - \text{Random Policy}|}$. Bold scores indicate an improvement over both baselines. The results are the average over $3$ runs using different random seeds. Variance of the added Gaussian noise is $\sigma^2$.}
\end{table}

\begin{table}[H]
\centering
\begin{tabular}{|c|c|c|c|}
\hline
Environment                 & $\epsilon=0.1$ &  $\epsilon=0.2$ & $\epsilon=0.3$ \\
& (\%  Gain) & (\% Gain) & (\% Gain) \\
\hline
\hline
BeamRiderNoFrameskip-v4     & \textbf{86.72}      & \textbf{71.28}      & \textbf{38.76}      \\ \hline
BreakoutNoFrameskip-v4      & \textbf{44.93}      & \textbf{280.40}     & \textbf{1169.55}    \\ \hline
PongNoFrameskip-v4          & -13.97              & -0.54               & \textbf{0.17}       \\ \hline
QbertNoFrameskip-v4         & \textbf{13.74}      & \textbf{248.81}     & \textbf{351.26}     \\ \hline
SeaquestNoFrameskip-v4      & \textbf{-20.64}     & \textbf{12.64}      & \textbf{91.16}      \\ \hline
SpaceInvadersNoFrameskip-v4 & \textbf{14.39}      & \textbf{36.31}      & \textbf{62.64}      \\ \hline
\hline
Average                     & \textbf{20.86}      & \textbf{108.15}     & \textbf{285.59}     \\ \hline
\end{tabular}
\caption{Comparison of the average episode reward over 10M steps of training between our approach to the best of both baselines (A2C and A2C with the reward prediction auxiliary task). The score represents the relative improvement over the best baseline normalized by the performance of the random policy: $\frac{\text{Ours} - \text{Best Baseline}}{|\text{Best Baseline} - \text{Random Policy}|}$. Bold scores indicate an improvement over both baselines. The results are the average over $3$ runs using different random seeds. $\epsilon$ values indicate likelihood of uniform noise.}
\end{table}

\begin{table}[H]
\centering
\begin{tabular}{|c|c|c|c|}
\hline
Environment                 & $\epsilon=0.3$ & $\epsilon=0.5$ & $\epsilon=0.75$ \\ 
& (\%  Gain) & (\% Gain) & (\% Gain) \\
\hline
\hline
BeamRiderNoFrameskip-v4     & \textbf{44.11}       & \textbf{64.69}       & \textbf{16.27}        \\ \hline
BreakoutNoFrameskip-v4      & \textbf{3.22}        & \textbf{4.25}        & \textbf{69.70}        \\ \hline
PongNoFrameskip-v4          & \textbf{22.03}       & \textbf{16.27}       & -0.75                 \\ \hline
QbertNoFrameskip-v4         & \textbf{31.49}       & \textbf{70.61}       & \textbf{141.26}       \\ \hline
SeaquestNoFrameskip-v4      & -41.34               & -26.05               & \textbf{11.20}        \\ \hline
SpaceInvadersNoFrameskip-v4 & \textbf{0.18}        & -2.21                & \textbf{22.74}        \\ \hline
Average                     & \textbf{9.95}        & \textbf{21.26}       & \textbf{43.40}        \\ \hline
\end{tabular}
\caption{Comparison of the average episode reward over 10M steps of training between our approach to the best of both baselines (A2C and A2C with the reward prediction auxiliary task). The score represents the relative improvement over the best baseline normalized by the performance of the random policy: $\frac{\text{Ours} - \text{Best Baseline}}{|\text{Best Baseline} - \text{Random Policy}|}$. Bold scores indicate an improvement over both baselines. The results are the average over $3$ runs using different random seeds. $\epsilon$ values indicate likelihood of sparsity induction.}

\end{table}

\begin{table}[H]
\label{tab:atari2}
\centering
\begin{tabular}{|l|l|l|l|l|}
\hline
Environment                                  & Gaussian Noise & A2C         & A2C+        & $\hat{R}(s)$  \\ \hline
\multirow{5}{*}{BeamRiderNoFrameskip-v4}     & 0              & 2151.73          & 1836.51          & \textbf{2639.65} \\ \cline{2-5} 
                                             & 0.1            & 1555.06          & 991.56           & \textbf{2157.87} \\ \cline{2-5} 
                                             & 0.2            & 416.91           & 429.00           & \textbf{1685.37} \\ \cline{2-5} 
                                             & 0.3            & 398.13           & 396.06           & \textbf{942.69}  \\ \cline{2-5} 
                                             & 0.4            & 390.60           & 393.35           & \textbf{671.57}  \\ \hline
\multirow{5}{*}{BreakoutNoFrameskip-v4}      & 0              & 252.19           & \textbf{274.57}  & 271.19           \\ \cline{2-5} 
                                             & 0.1            & 222.99           & 223.14           & \textbf{257.24}  \\ \cline{2-5} 
                                             & 0.2            & 114.05           & 119.35           & \textbf{239.15}  \\ \cline{2-5} 
                                             & 0.3            & 27.51            & 29.83            & \textbf{221.64}  \\ \cline{2-5} 
                                             & 0.4            & 6.86             & 10.93            & \textbf{209.68}  \\ \hline
\multirow{5}{*}{PongNoFrameskip-v4}          & 0              & \textbf{9.95}    & 9.60             & 9.88             \\ \cline{2-5} 
                                             & 0.1            & \textbf{0.07}    & -2.10            & 4.57             \\ \cline{2-5} 
                                             & 0.2            & -14.63           & \textbf{-10.13}  & -10.29           \\ \cline{2-5} 
                                             & 0.3            & -20.46           & -20.42           & \textbf{-15.08}  \\ \cline{2-5} 
                                             & 0.4            & -20.47           & -20.43           & \textbf{-20.34}  \\ \hline
\multirow{5}{*}{QbertNoFrameskip-v4}         & 0              & 4539.80          & \textbf{7332.23} & 4638.91          \\ \cline{2-5} 
                                             & 0.1            & 3343.67          & \textbf{4867.65} & 4388.63          \\ \cline{2-5} 
                                             & 0.2            & 1584.20          & 2362.24          & \textbf{4089.10} \\ \cline{2-5} 
                                             & 0.3            & 608.56           & 712.62           & \textbf{3217.89} \\ \cline{2-5} 
                                             & 0.4            & 412.90           & 531.20           & \textbf{2905.16} \\ \hline
\multirow{5}{*}{SeaquestNoFrameskip-v4}      & 0              & \textbf{1381.11} & 1154.93          & 993.49           \\ \cline{2-5} 
                                             & 0.1            & 619.22           & \textbf{963.16}  & 881.00           \\ \cline{2-5} 
                                             & 0.2            & \textbf{785.69}  & 352.32           & 723.40           \\ \cline{2-5} 
                                             & 0.3            & 297.31           & 262.98           & \textbf{468.21}  \\ \cline{2-5} 
                                             & 0.4            & 162.95           & 185.23           & \textbf{320.59}  \\ \hline
\multirow{5}{*}{SpaceInvadersNoFrameskip-v4} & 0              & \textbf{630.36}  & 558.65           & 579.80           \\ \cline{2-5} 
                                             & 0.1            & 542.80           & 549.90           & \textbf{583.90}  \\ \cline{2-5} 
                                             & 0.2            & 420.60           & 422.60           & \textbf{573.91}  \\ \cline{2-5} 
                                             & 0.3            & 300.52           & 281.72           & \textbf{508.40}  \\ \cline{2-5} 
                                             & 0.4            & 218.09           & 209.49           & \textbf{473.78}  \\ \hline
\end{tabular}
\caption{The asymptotic average return across all episodes of training for the Atari experiments across 3 training runs with different random seeds (for both network initialization and environment seeding) under the Gaussian noise.}

\end{table}

\begin{table}[H]
\centering
\begin{tabular}{|l|l|l|l|l|}
\hline
Environment                                  & Uniform Noise & A2C             & A2C+             & Ours             \\ \hline
\multirow{3}{*}{BeamRiderNoFrameskip-v4}     & 0.1           & 402.36          & 532.39           & \textbf{1286.34} \\ \cline{2-5} 
                                             & 0.2           & 398.35          & 395.18           & \textbf{922.52}  \\ \cline{2-5} 
                                             & 0.3           & 390.33          & 393.20           & \textbf{676.20}  \\ \hline
\multirow{3}{*}{BreakoutNoFrameskip-v4}      & 0.1           & 149.31          & 160.98           & \textbf{234.07}  \\ \cline{2-5} 
                                             & 0.2           & 45.28           & 55.66            & \textbf{216.50}  \\ \cline{2-5} 
                                             & 0.3           & 12.79           & 13.43            & \textbf{190.33}  \\ \hline
\multirow{3}{*}{PongNoFrameskip-v4}          & 0.1           & -13.91          & \textbf{-8.94}   & -13.08           \\ \cline{2-5} 
                                             & 0.2           & \textbf{-20.08} & -20.37           & -20.30           \\ \cline{2-5} 
                                             & 0.3           & -20.41          & -20.38           & \textbf{-20.31}  \\ \hline
\multirow{3}{*}{QbertNoFrameskip-v4}         & 0.1           & 1778.70         & 3436.01          & \textbf{3930.60} \\ \cline{2-5} 
                                             & 0.2           & 752.14          & 589.16           & \textbf{3031.33} \\ \cline{2-5} 
                                             & 0.3           & 394.96          & 440.09           & \textbf{2561.66} \\ \hline
\multirow{3}{*}{SeaquestNoFrameskip-v4}      & 0.1           & 462.80          & \textbf{1005.19} & 783.56           \\ \cline{2-5} 
                                             & 0.2           & 384.28          & 184.19           & \textbf{441.49}  \\ \cline{2-5} 
                                             & 0.3           & 168.80          & 162.69           & \textbf{385.03}  \\ \hline
\multirow{3}{*}{SpaceInvadersNoFrameskip-v4} & 0.1           & 451.64          & 420.91           & \textbf{537.92}  \\ \cline{2-5} 
                                             & 0.2           & 319.46          & 279.20           & \textbf{489.20}  \\ \cline{2-5} 
                                             & 0.3           & 227.79          & 207.25           & \textbf{463.20}  \\ \hline
\end{tabular}
\caption{The asymptotic average return across all episodes episodes of training for the Atari experiments across 3 training runs with different random seeds (for both network initialization and environment seeding) under the uniform noise.}
\end{table}

\begin{table}[H]
\centering
\begin{tabular}{|l|l|l|l|l|}
\hline
Environment                                  & Sparsity Noise & A2C     & A2C+             & Ours             \\ \hline
\multirow{3}{*}{BeamRiderNoFrameskip-v4}     & 0.3            & 1555.07 & 1505.46          & \textbf{2389.58} \\ \cline{2-5} 
                                             & 0.5            & 918.60  & 931.14           & \textbf{1751.48} \\ \cline{2-5} 
                                             & 0.75           & 492.38  & 405.17           & \textbf{627.33}  \\ \hline
\multirow{3}{*}{BreakoutNoFrameskip-v4}      & 0.3            & 256.43  & 259.22           & \textbf{267.63}  \\ \cline{2-5} 
                                             & 0.5            & 223.10  & 234.60           & \textbf{244.65}  \\ \cline{2-5} 
                                             & 0.75           & 101.99  & 88.84            & \textbf{174.25}  \\ \hline
\multirow{3}{*}{PongNoFrameskip-v4}          & 0.3            & -0.53   & -0.01            & \textbf{4.55}    \\ \cline{2-5} 
                                             & 0.5            & -6.73   & -20.28           & \textbf{-2.27}   \\ \cline{2-5} 
                                             & 0.75           & -20.28  & \textbf{-19.38}  & -19.68           \\ \hline
\multirow{3}{*}{QbertNoFrameskip-v4}         & 0.3            & 2534.81 & 3232.79          & \textbf{4302.55} \\ \cline{2-5} 
                                             & 0.5            & 1666.17 & 1986.04          & \textbf{3504.11} \\ \cline{2-5} 
                                             & 0.75           & 564.04  & 587.49           & \textbf{1648.87} \\ \hline
\multirow{3}{*}{SeaquestNoFrameskip-v4}      & 0.3            & 1195.62 & \textbf{1319.00} & 745.50           \\ \cline{2-5} 
                                             & 0.5            & 938.44  & \textbf{957.43}  & 690.18           \\ \cline{2-5} 
                                             & 0.75           & 399.54  & 354.25           & \textbf{451.95}  \\ \hline
\multirow{3}{*}{SpaceInvadersNoFrameskip-v4} & 0.3            & 562.61  & 588.71           & \textbf{590.05}  \\ \cline{2-5} 
                                             & 0.5            & 496.33  & \textbf{510.24}  & 495.69           \\ \cline{2-5} 
                                             & 0.75           & 348.18  & 328.75           & \textbf{461.00}  \\ \hline
\end{tabular}
\caption{The asymptotic average return across all episodes of training for the Atari experiments across 3 training runs with different random seeds (for both network initialization and environment seeding) under the sparsity inducing noise.}

\end{table}

\textbf{Additional details:} Our architecture and hyper-parameters are identical to the standard A2C parameters used in \cite{baselines}.
To learn the reward-predictor, we used a completely separate network with the same overall structure as the value/policy network. 

We tuned the learning rate for the reward-predictor roughly through a coarse grid-search between $[0.0001, 0.00025, 0.0005, 0.00075, 0.001]$ on a single game Pong and then used the best one ($0.0001$) on all other games.

Additionally, we found that occasionally our algorithm diverged completely due to poor initialization of the reward-predictor. To alleviate this issue, we provided a convex combination between our estimate $\hat{r}$ and the stochastic corrupted environment reward (for all environments) - which we linearly decayed over the first $25000$ network updates (out of the total $125000$ updates).

For Atari experiments we model $\hat{R}(s)$ which we find to be sufficient to improve performance since rewards are not directly related to the actions themselves and are often delayed by several steps.

\section{Mujoco Experiments}
\label{app:mujoco}

\textbf{Additional details:} Our architecture and hyper-parameters are identical to the standard PPO parameters used in \cite{baselines}, our implementation is directly taken from \cite{pytorchrl}.
To learn the reward-predictor, we used a completely separate network with the same overall structure as the value network. 

We used the same learning rate as was standard for the policy / value network $0.0003$.

Additionally, we found that occasionally our algorithm diverged completely due to poor initialization of the reward-predictor. To alleviate this issue, we provided a convex combination between our estimate $\hat{r}$ and the stochastic corrupted environment reward (for all environments) - which we linearly decayed over the first $100$ network updates (out of the total $\sim 500$ updates).

For MuJoCo experiments we model $\hat{R}(s,a,s')$. We find that adding an expectation over states alone does not provide enough fidelity on the reward function since several of the tasks provide a reward for the action itself (e.g., Reacher). 

\subsection{Performance of Random Agent}

For the displayed results in the main text we normalize by the performance over a random agent. These performances were averaged over 100 episodes by uniformly sampling from the action space and can be found in Table~\ref{tab:mujocorandom},

\begin{table}[H]
    \centering
    \begin{tabular}{|c|c|c|c|c|}
    \hline
         Env & Hopper& Walker2d & HalfCheetah & Reacher  \\
         \hline
         Return & 16.97 & 1.54 & -272 & -43.1 \\
         \hline
    \end{tabular}
    \caption{Average return of a random uniform sampling policy on MuJoCo tasks across 100 episodes. }
    \label{tab:mujocorandom}
\end{table}

\subsection{Result Tables}

\begin{table}[H]
\centering
\begin{tabular}{|l|l|l|l|l|l|l|}
\hline
Environment                     & Gaussian Noise & PPO     & PPO+    & $\hat{R}$(s) & $\hat{R}$(s,a) & $\hat{R}$(s,a,s') \\ \hline
\multirow{5}{*}{HalfCheetah-v2} & 0.0              & 1392.40 & 1462.89 & 174.38                      & \textbf{1638.44}                       & 1245.21                          \\ \cline{2-7} 
                                & 0.1            & 977.12  & 803.66  & 404.80                      & \textbf{1595.80}                       & 1460.57                          \\ \cline{2-7} 
                                & 0.2            & 339.41  & 300.59  & 105.42                      & \textbf{1095.05}                       & 1043.79                          \\ \cline{2-7} 
                                & 0.3            & 122.36  & 56.86   & 115.28                      & \textbf{1148.32}                       & 672.59                           \\ \cline{2-7} 
                                & 0.4            & -131.29 & -149.76 & -98.19                      & \textbf{868.31}                        & 563.25                           \\ \hline
\multirow{5}{*}{Hopper-v2}      & 0.0              & 1854.69 & \textbf{1991.64} & 1820.39                     & 1828.26                       & 1831.93                        \\ \cline{2-7} 
                                & 0.1            & 1689.57 & 1809.31 & 1857.21                     & 1801.71                       & \textbf{1881.85 }                         \\ \cline{2-7} 
                                & 0.2            & 1170.60 & 1687.88 & 1873.61                     & \textbf{1922.32}                       & 1790.57                          \\ \cline{2-7} 
                                & 0.3            & 1420.17 & 1184.10 & 1542.23                     & \textbf{1767.25}                       & 1565.90                          \\ \cline{2-7} 
                                & 0.4            & 843.86  & 1188.19 & \textbf{1793.98}                     & 1697.48                       & 1579.56                          \\ \hline
\multirow{5}{*}{Reacher-v2}     & 0.0              & -6.53   & \textbf{-6.41}   & -99.90                      & -6.54                         & -7.07                            \\ \cline{2-7} 
                                & 0.1            & -17.28  & -17.11  & -127.82                     & -15.94                        & \textbf{-14.40}                           \\ \cline{2-7} 
                                & 0.2            & -22.43  & -22.88  & -161.28                     & -20.87                        & \textbf{-19.00}                           \\ \cline{2-7} 
                                & 0.3            & -26.34  & -27.33  & -173.28                     & -22.57                        & \textbf{-21.20}                           \\ \cline{2-7} 
                                & 0.4            & -29.14  & -32.53  & -175.05                     & -25.70                        & \textbf{-25.68}                           \\ \hline
\multirow{5}{*}{Walker2d-v2}    & 0.0              & 2340.91 & \textbf{2672.04} & 2346.76                     & 2447.12                       & 2455.96                          \\ \cline{2-7} 
                                & 0.1            & 1412.13 & 1211.24 & 2153.51                     & \textbf{2473.58}                       & 2310.20                          \\ \cline{2-7} 
                                & 0.2            & 677.54  & 589.28  & 1571.95                     & \textbf{1966.97}                       & 1752.62                          \\ \cline{2-7} 
                                & 0.3            & 393.01  & 540.58  & 1292.29                     & 1091.87                       & \textbf{1497.89}                          \\ \cline{2-7} 
                                & 0.4            & 400.04  & 352.48  & 1160.65                     & \textbf{1362.22}                       & 1000.16                          \\ \hline
\end{tabular}
\caption{The asymptotic average return across the last 100 episodes of training for the MuJoCo experiments across 10 training runs with different random seeds (for both network initialization and environment seeding). We also include results for varying features provided to the $\hat{R}$ estimator such that this demonstrates the selection criteria in the main text. As can be seen, in several tasks where the reward has a component related to the action or transition itself, the use of action as a feature is required.}
\end{table}

\begin{table}[H]
\centering
\begin{tabular}{|l|l|l|l|l|l|l|}
\hline
Environment                     & Sparsity Noise & PPO     & PPO+    & $\hat{R}$(s)    & $\hat{R}$(s,a)  & $\hat{R}$(s,a,s') \\ \hline
\multirow{5}{*}{HalfCheetah-v2} & 0.6            & \textbf{1612.95} & 1222.08 & 361.60  & 1233.47 & 1379.16   \\ \cline{2-7} 
                                & 0.7            & 1232.03 & \textbf{1305.56} & 198.67  & 1200.28 & 1076.82   \\ \cline{2-7} 
                                & 0.8            & \textbf{1293.07} & 1113.19 & 30.52   & 1142.94 & 1282.56   \\ \cline{2-7} 
                                & 0.9            & \textbf{1238.62} & 810.83  & 152.05  & 648.19  & 1147.88   \\ \cline{2-7} 
                                & 0.95           & 98.35   & 113.87  & -375.79 & \textbf{783.76}  & 592.91    \\ \hline
\multirow{5}{*}{Hopper-v2}      & 0.6            & 1675.30 & 1618.70 & 1788.76 & \textbf{2058.57} & 1945.85   \\ \cline{2-7} 
                                & 0.7            & 1883.82 & \textbf{1944.51} & 1864.50 & 1753.45 & 1790.25   \\ \cline{2-7} 
                                & 0.8            & 1720.89 & 1661.29 & \textbf{1954.20} & 1700.17 & 1755.76   \\ \cline{2-7} 
                                & 0.9            & 1023.55 & 678.48  & \textbf{1778.44} & 1560.97 & 1753.77   \\ \cline{2-7} 
                                & 0.95           & 1031.92 & 598.93  & \textbf{1497.67} & 1487.81 & 1863.43   \\ \hline
\multirow{5}{*}{Reacher-v2}     & 0.6            & -8.43   & \textbf{-8.29}   & -118.82 & -12.18  & -11.76    \\ \cline{2-7} 
                                & 0.7            & \textbf{-8.51}   & -11.09  & -138.26 & -14.10  & -14.16    \\ \cline{2-7} 
                                & 0.8            & \textbf{-10.75}  & -12.24  & -157.20 & -17.77  & -16.68    \\ \cline{2-7} 
                                & 0.9            & \textbf{-15.64}  & -20.07  & -188.48 & -27.33  & -25.17    \\ \cline{2-7} 
                                & 0.95           & -53.25  & -50.36  & -241.73 & \textbf{-36.47}  & -44.31    \\ \hline
\multirow{5}{*}{Walker2d-v2}    & 0.6            & 1692.40 & 2126.73 & 2053.24 & \textbf{2581.97} & 2258.18   \\ \cline{2-7} 
                                & 0.7            & 1357.61 & 1949.10 & 2270.63 & 2090.89 & \textbf{2290.78}   \\ \cline{2-7} 
                                & 0.8            & 1655.21 & 1141.59 & 2059.51 & \textbf{2477.55} & 2187.35   \\ \cline{2-7} 
                                & 0.9            & 679.50  & 731.16  & 1525.32 & 1828.97 & \textbf{2227.78}   \\ \cline{2-7} 
                                & 0.95           & 600.89  & 456.80  & 966.98  & \textbf{1727.84} & 1383.83   \\ \hline
\end{tabular}
\caption{The asymptotic average return across the last 100 episodes of training for the MuJoCo experiments across 10 training runs with different random seeds (for both network initialization and environment seeding). We also include results for varying features provided to the $\hat{R}$ estimator such that this demonstrates the selection criteria in the main text. As can be seen, in several tasks where the reward has a component related to the action or transition itself, the use of action as a feature is required. Sparsity inducing noise did not always yield improvements over the baseline except for in higher noise conditions.}

\end{table}

\begin{table}[H]
\centering
\begin{tabular}{|l|l|l|l|l|l|l|}
\hline
Environment                     & Uniform Noise & PPO     & PPO+    & $\hat{R}$(s)      & $\hat{R}$(s,a)            & $\hat{R}$(s,a,s')         \\ \hline
\multirow{4}{*}{HalfCheetah-v2} & 0.1           & 293.01  & 391.48  & 99.28    & 990.88           & \textbf{1127.91} \\ \cline{2-7} 
                                & 0.2           & 55.01   & 74.90   & -44.54   & \textbf{933.78}  & 812.91           \\ \cline{2-7} 
                                & 0.3           & -164.12 & -141.44 & -178.28  & \textbf{641.36}  & 583.93           \\ \cline{2-7} 
                                & 0.4           & -432.13 & -304.93 & -1017.45 & 273.59           & \textbf{368.28}  \\ \hline
\multirow{4}{*}{Hopper-v2}      & 0.1           & 1579.85 & 1629.17 & 1799.82  & 1842.31          & \textbf{2092.86} \\ \cline{2-7} 
                                & 0.2           & 1303.62 & 1101.23 & 1494.78  & 1622.64          & \textbf{1952.41} \\ \cline{2-7} 
                                & 0.3           & 1085.40 & 935.10  & 1262.49  & \textbf{1323.28} & 1307.01          \\ \cline{2-7} 
                                & 0.4           & 559.75  & 758.62  & 1081.88  & 1109.00          & \textbf{1577.78} \\ \hline
\multirow{4}{*}{Reacher-v2}     & 0.1           & -21.66  & -23.61  & -148.80  & \textbf{-18.16}  & -18.31           \\ \cline{2-7} 
                                & 0.2           & -26.20  & -27.87  & -167.59  & -23.58           & \textbf{-22.04}  \\ \cline{2-7} 
                                & 0.3           & -31.29  & -33.03  & -176.22  & \textbf{-24.93}  & -27.44           \\ \cline{2-7} 
                                & 0.4           & -37.97  & -37.79  & -201.12  & \textbf{-30.95}  & -35.65           \\ \hline
\multirow{4}{*}{Walker2d-v2}    & 0.1           & 720.39  & 639.01  & 1861.84  & \textbf{2176.73} & 2020.49          \\ \cline{2-7} 
                                & 0.2           & 430.20  & 355.24  & 1071.72  & \textbf{1235.42} & 883.65           \\ \cline{2-7} 
                                & 0.3           & 350.06  & 343.81  & 687.34   & \textbf{986.12}  & 788.61           \\ \cline{2-7} 
                                & 0.4           & 287.41  & 320.70  & 582.23   & \textbf{752.07}  & 431.05           \\ \hline
\end{tabular}
\caption{The asymptotic average return across the last 100 episodes of training for the MuJoCo experiments across 10 training runs with different random seeds (for both network initialization and environment seeding). We also include results for varying features provided to the $\hat{R}$ estimator such that this demonstrates the selection criteria in the main text. As can be seen, in several tasks where the reward has a component related to the action or transition itself, the use of action as a feature is required.}

\end{table}

\subsection{Extended Experiments}

\begin{sidewaysfigure}
    \centering
    \includegraphics[width=.19\textwidth]{HalfCheetah-v2_Gaussian_Noise_0-0_Episode_Reward}
    \includegraphics[width=.19\textwidth]{HalfCheetah-v2_Gaussian_Noise_0-1_Episode_Reward}
    \includegraphics[width=.19\textwidth]{HalfCheetah-v2_Gaussian_Noise_0-2_Episode_Reward}
    \includegraphics[width=.19\textwidth]{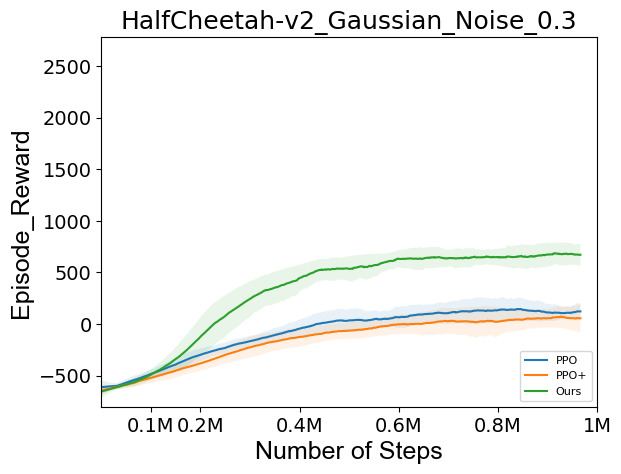}
    \includegraphics[width=.19\textwidth]{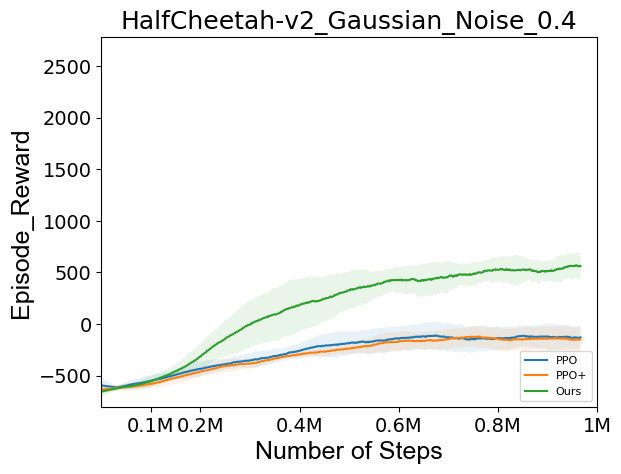}
    
    \includegraphics[width=.19\textwidth]{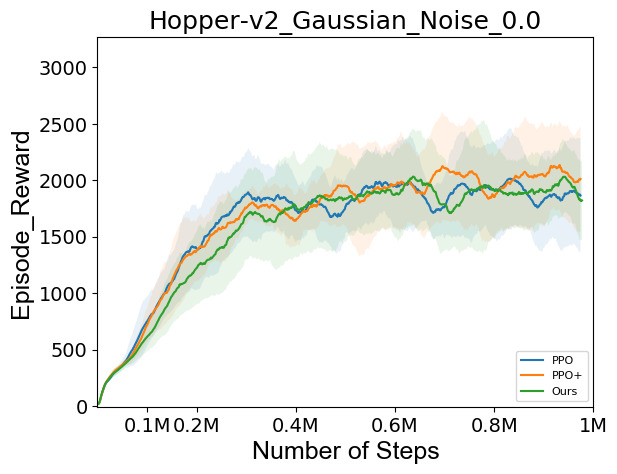}
    \includegraphics[width=.19\textwidth]{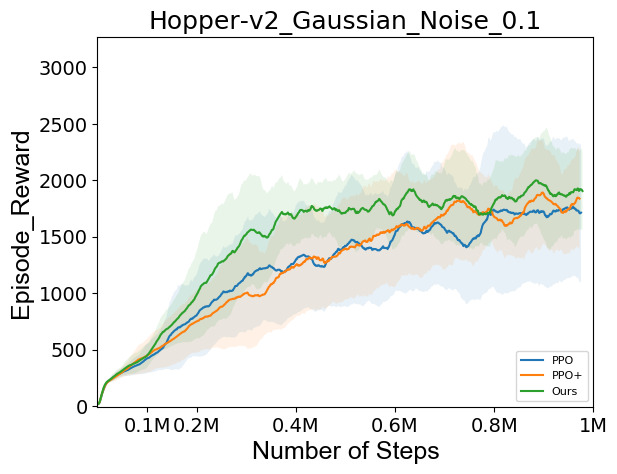}
    \includegraphics[width=.19\textwidth]{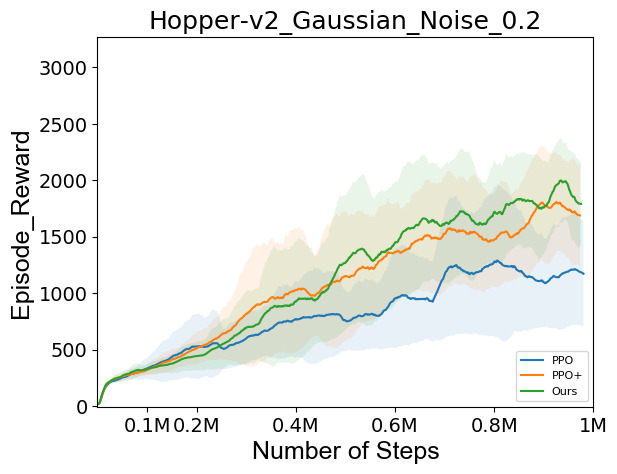}
    \includegraphics[width=.19\textwidth]{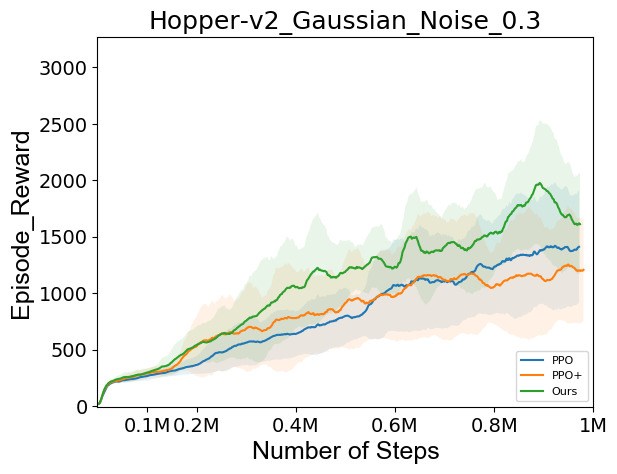}
    \includegraphics[width=.19\textwidth]{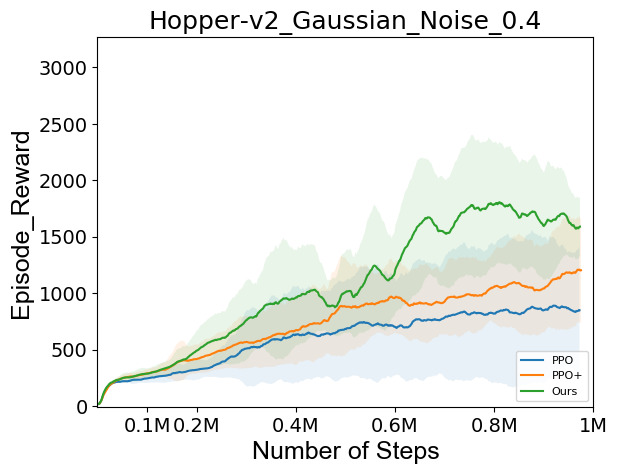}
    
    \includegraphics[width=.19\textwidth]{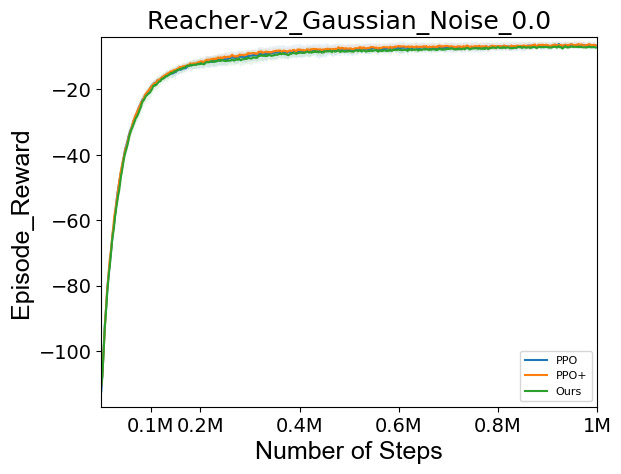}
    \includegraphics[width=.19\textwidth]{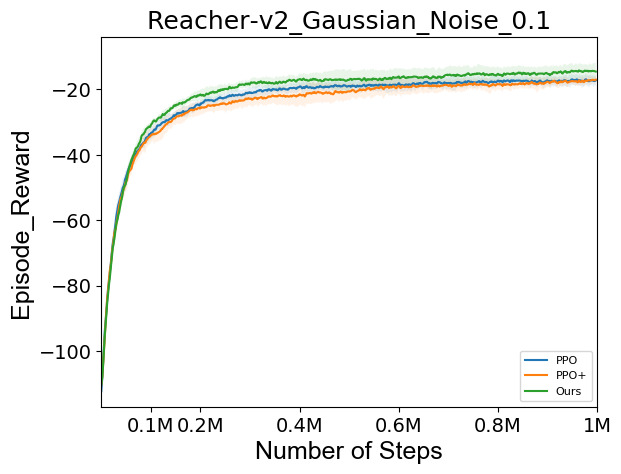}
    \includegraphics[width=.19\textwidth]{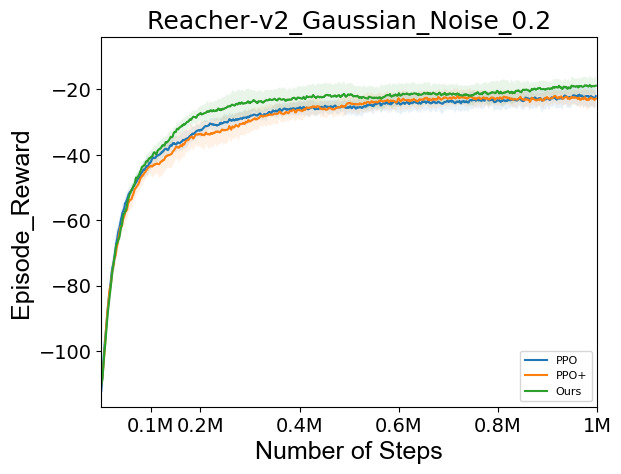}
    \includegraphics[width=.19\textwidth]{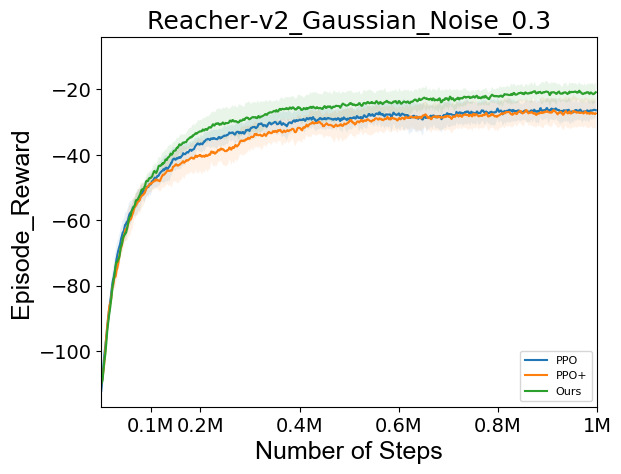}
    \includegraphics[width=.19\textwidth]{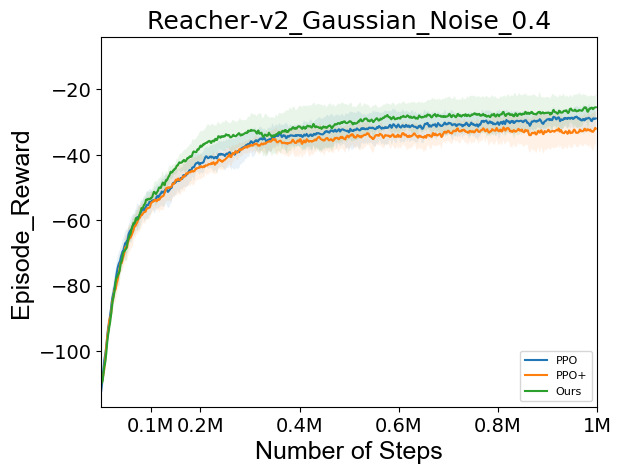}
    
    \includegraphics[width=.19\textwidth]{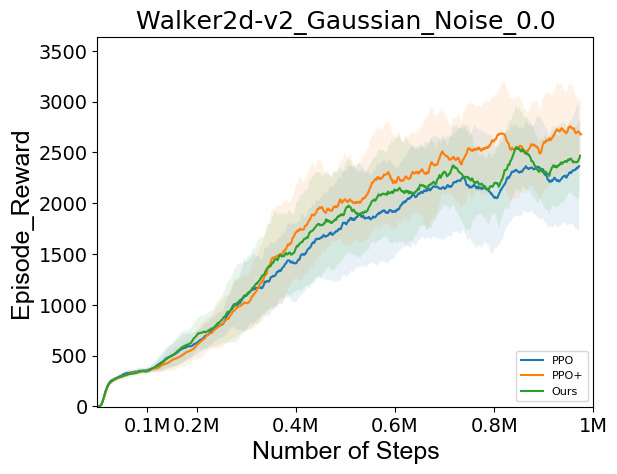}
    \includegraphics[width=.19\textwidth]{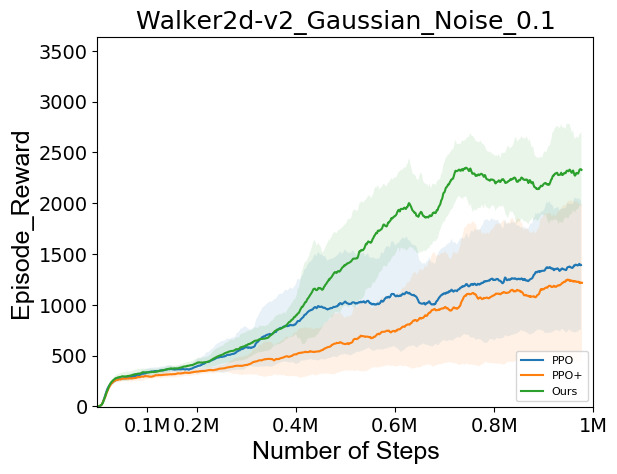}
    \includegraphics[width=.19\textwidth]{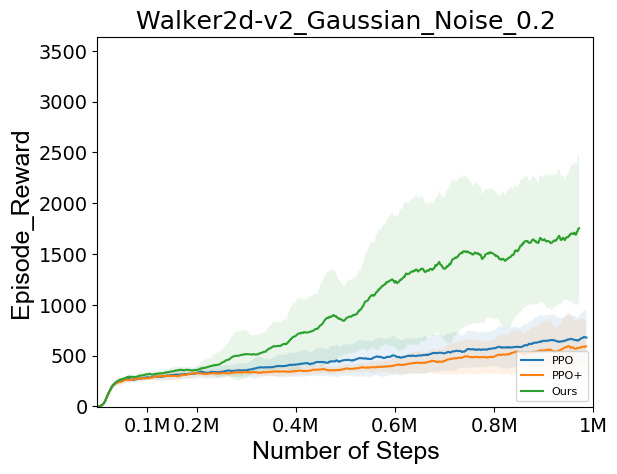}
    \includegraphics[width=.19\textwidth]{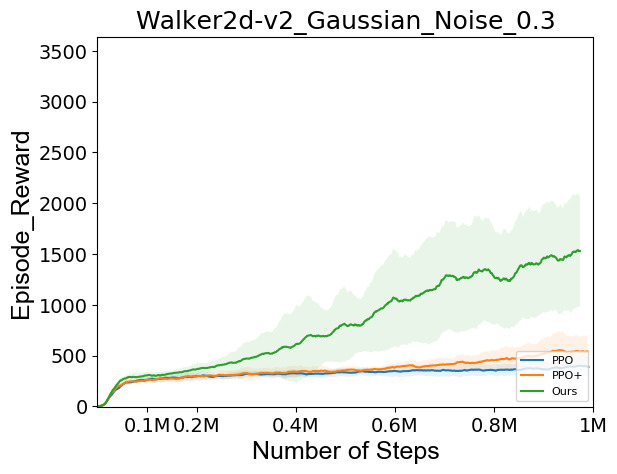}
    \includegraphics[width=.19\textwidth]{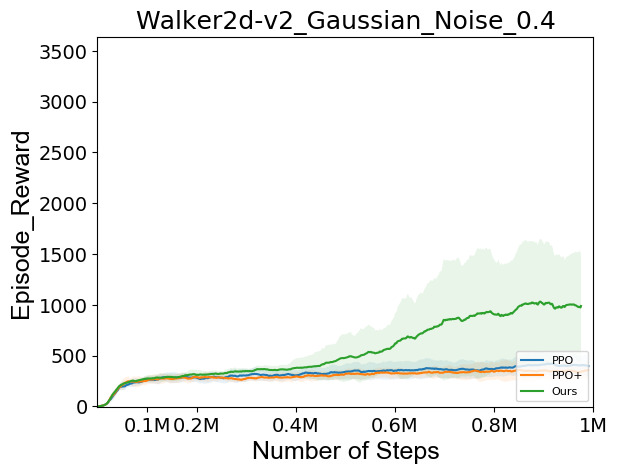}
    
    \caption{Gaussian noise}
\end{sidewaysfigure}

\begin{sidewaysfigure}
    \centering
    \includegraphics[width=.19\textwidth]{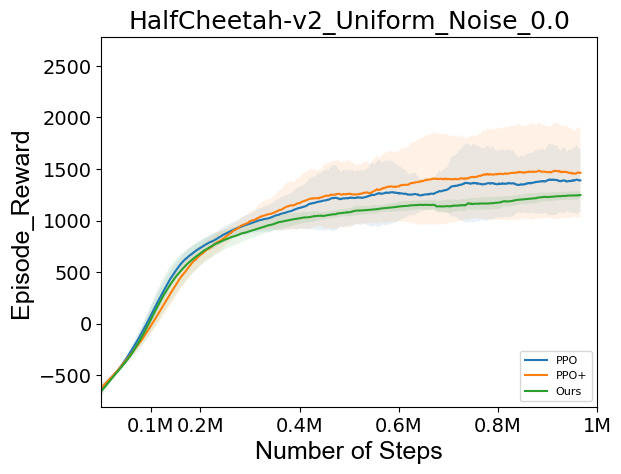}
    \includegraphics[width=.19\textwidth]{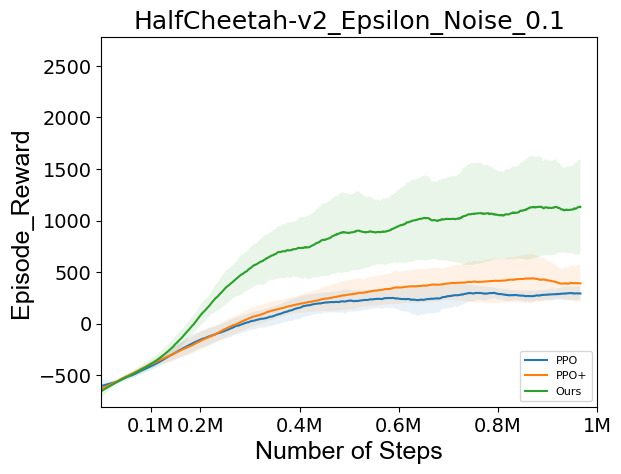}
    \includegraphics[width=.19\textwidth]{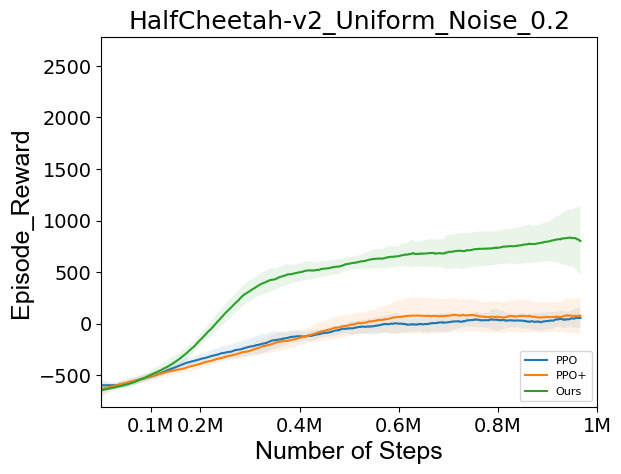}
    \includegraphics[width=.19\textwidth]{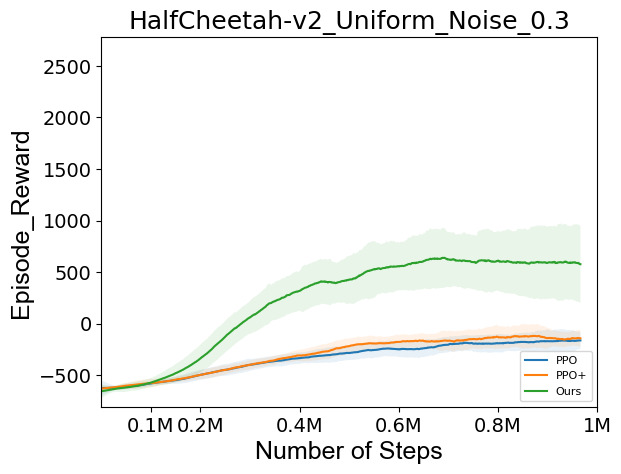}
    \includegraphics[width=.19\textwidth]{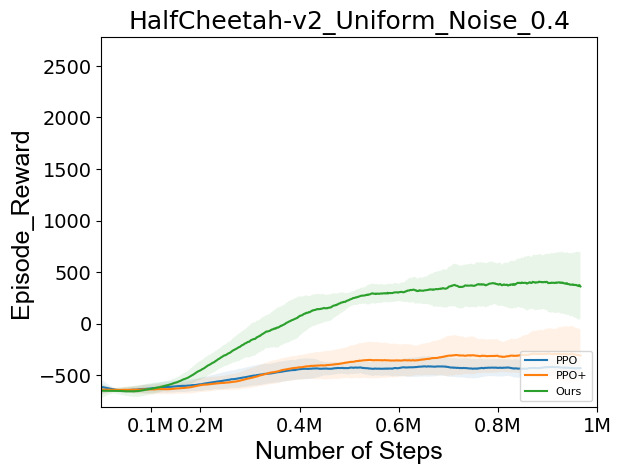}
    
    \includegraphics[width=.19\textwidth]{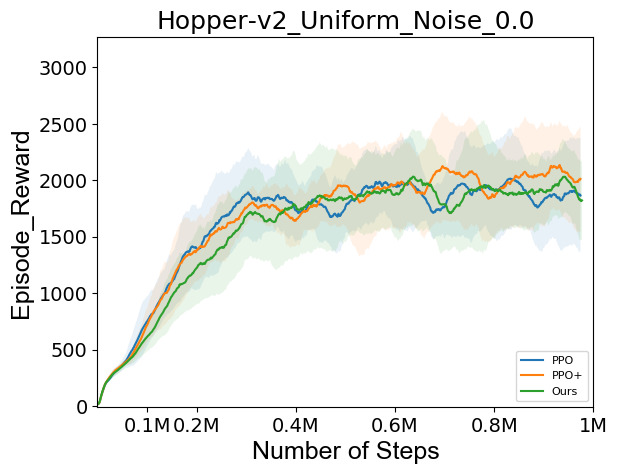}
    \includegraphics[width=.19\textwidth]{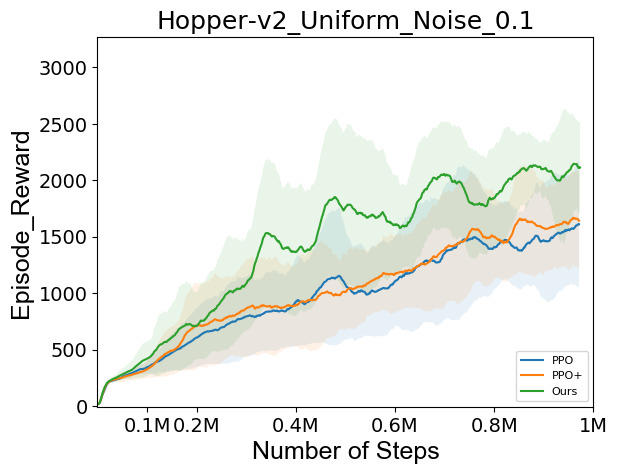}
    \includegraphics[width=.19\textwidth]{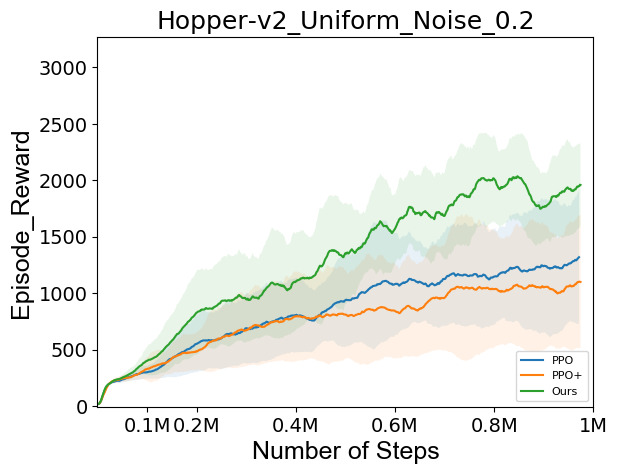}
    \includegraphics[width=.19\textwidth]{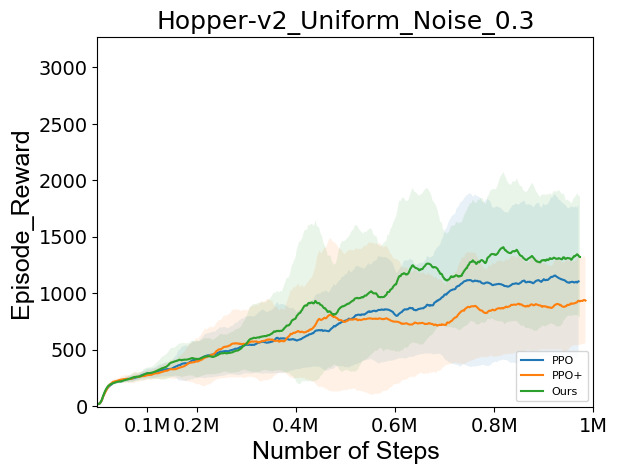}
    \includegraphics[width=.19\textwidth]{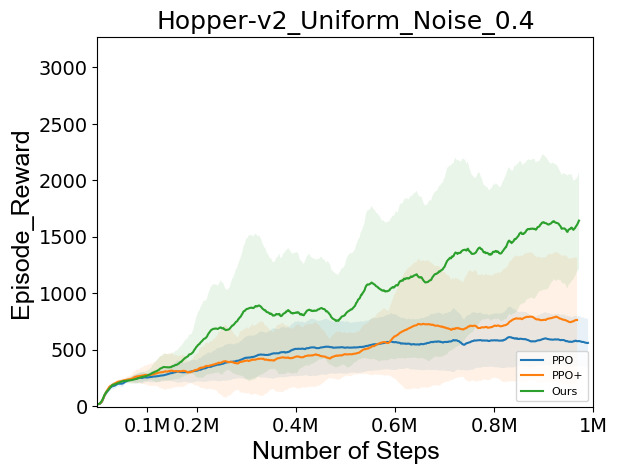}
    
    \includegraphics[width=.19\textwidth]{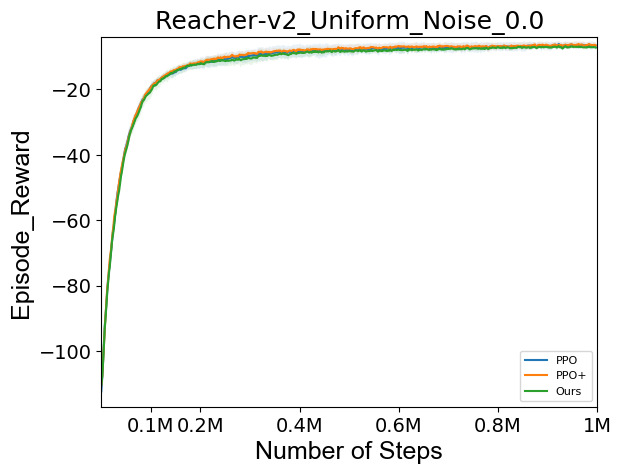}
    \includegraphics[width=.19\textwidth]{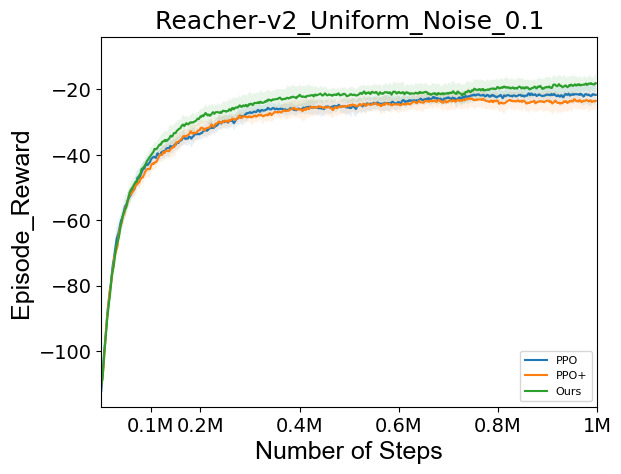}
    \includegraphics[width=.19\textwidth]{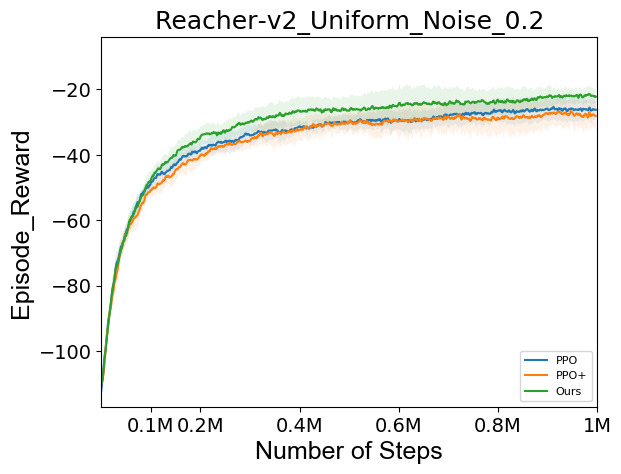}
    \includegraphics[width=.19\textwidth]{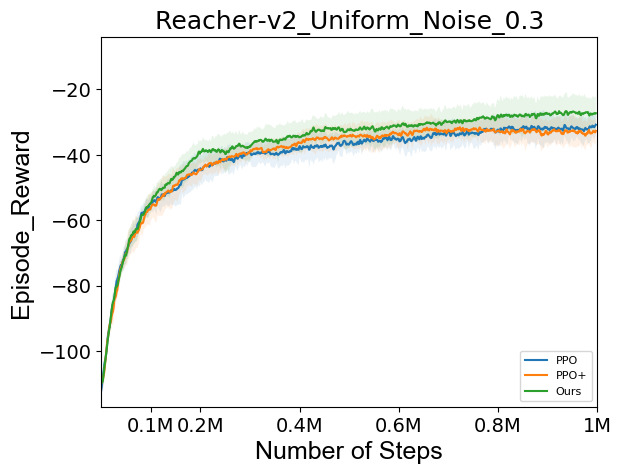}
    \includegraphics[width=.19\textwidth]{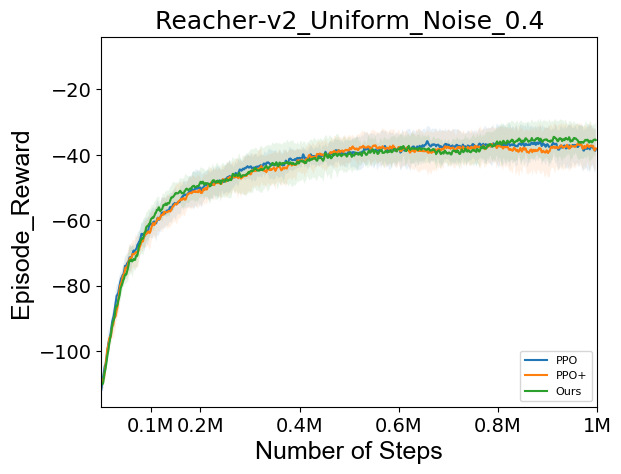}
    
    \includegraphics[width=.19\textwidth]{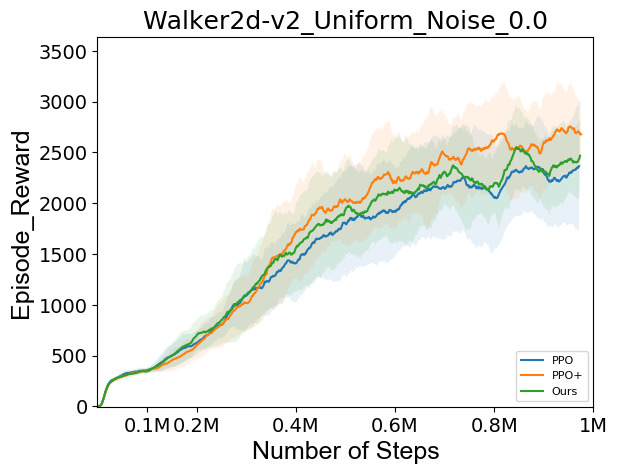}
    \includegraphics[width=.19\textwidth]{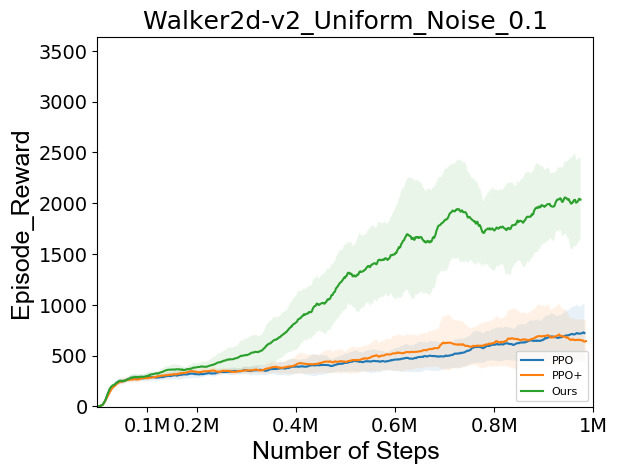}
    \includegraphics[width=.19\textwidth]{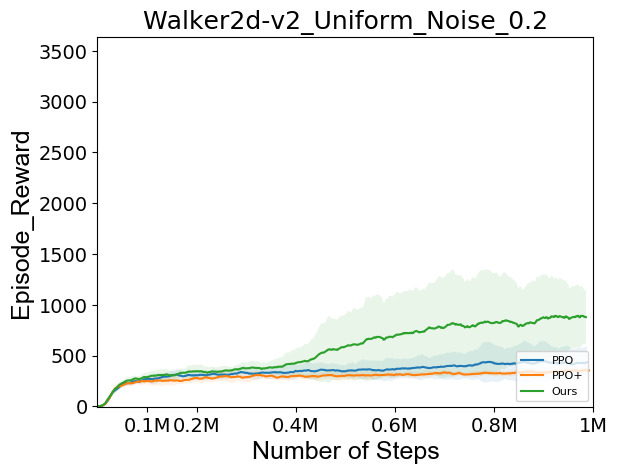}
    \includegraphics[width=.19\textwidth]{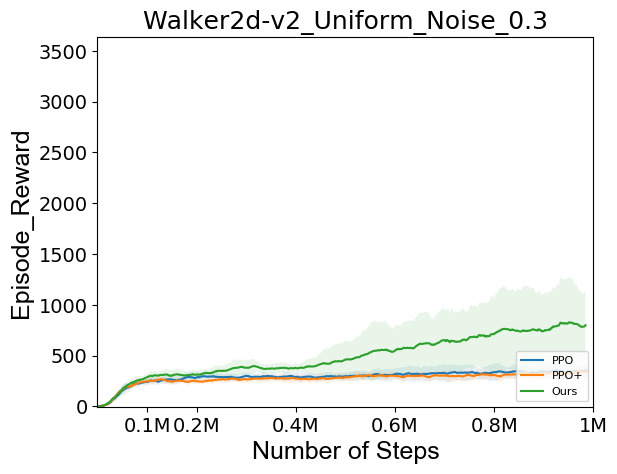}
    \includegraphics[width=.19\textwidth]{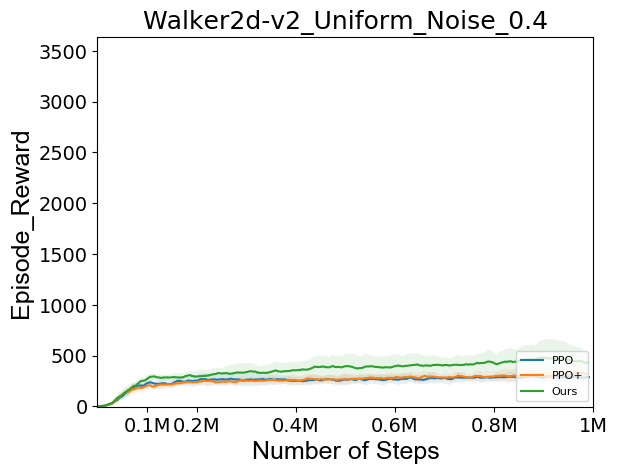}
    
    \caption{Uniform noise - with probability p the reward is replaced with a sampled reward from U[-1,1]}
\end{sidewaysfigure}

\begin{sidewaysfigure}
    \centering
    \includegraphics[width=.19\textwidth]{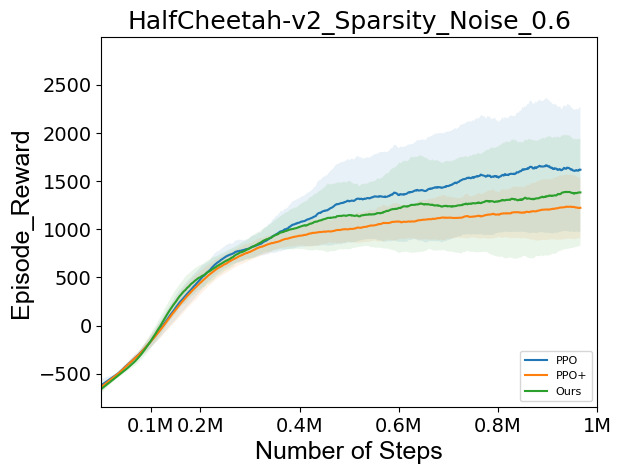}
    \includegraphics[width=.19\textwidth]{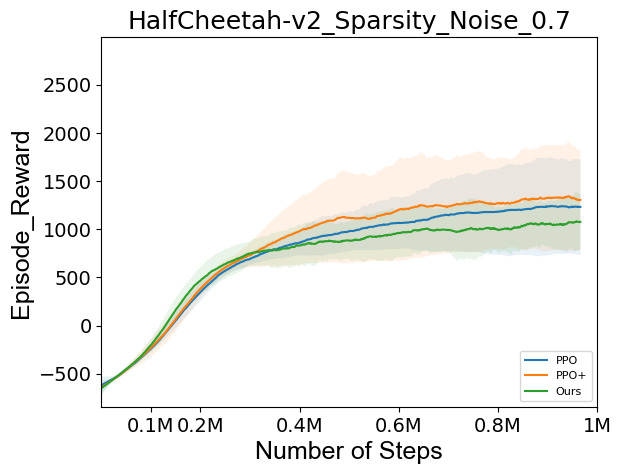}
    \includegraphics[width=.19\textwidth]{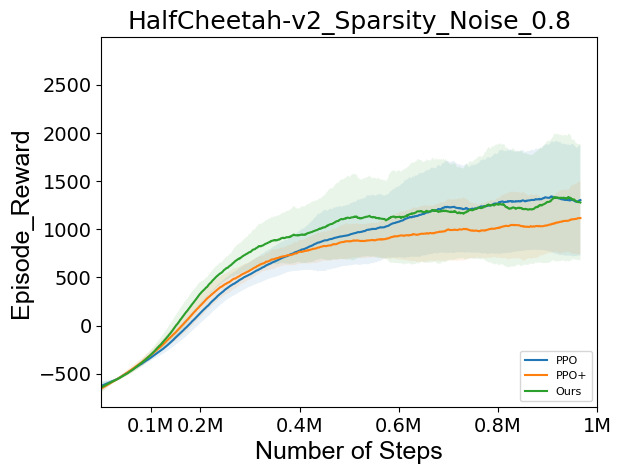}
    \includegraphics[width=.19\textwidth]{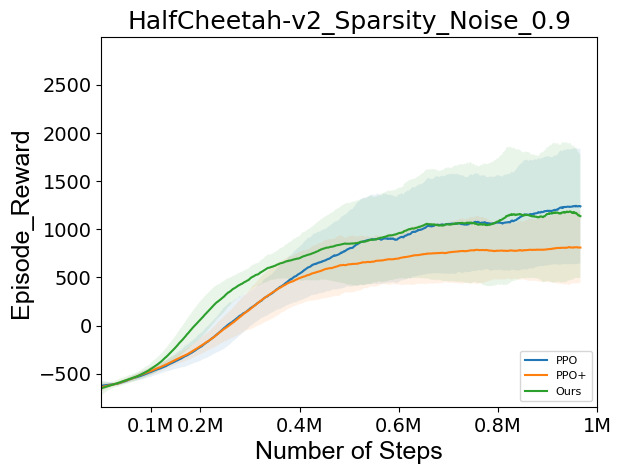}
    \includegraphics[width=.19\textwidth]{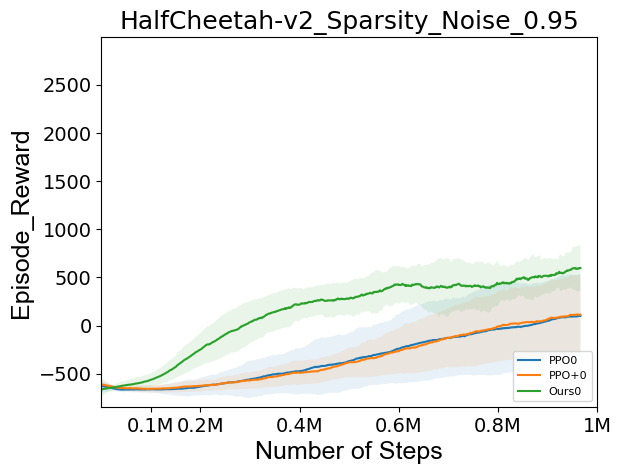}
    
    \includegraphics[width=.19\textwidth]{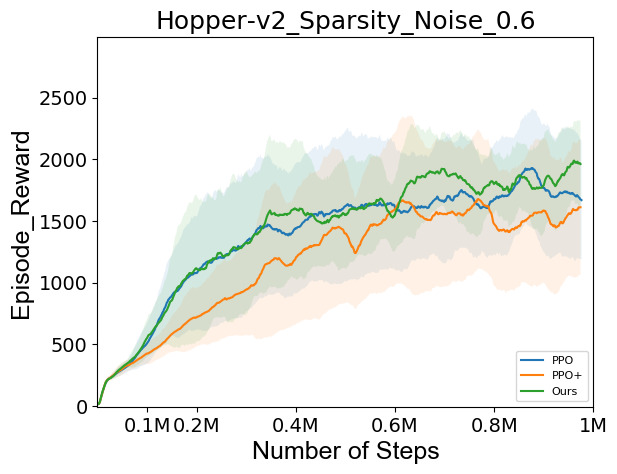}
    \includegraphics[width=.19\textwidth]{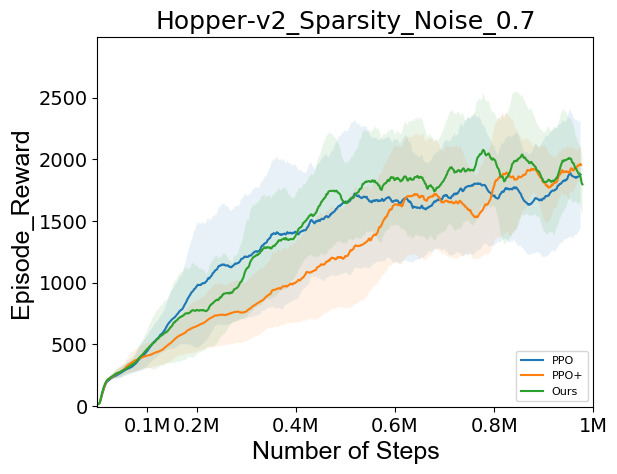}
    \includegraphics[width=.19\textwidth]{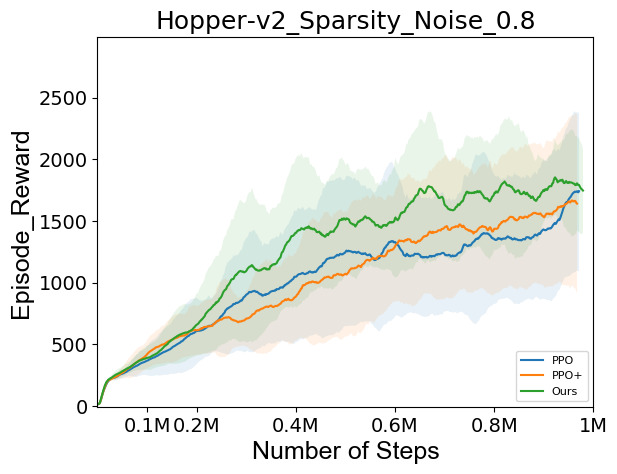}
    \includegraphics[width=.19\textwidth]{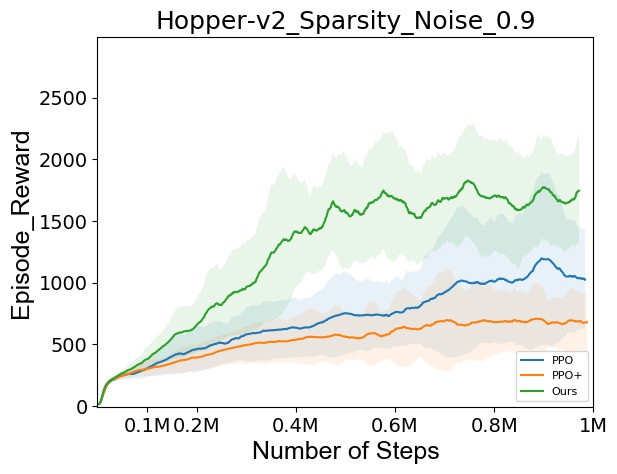}
    \includegraphics[width=.19\textwidth]{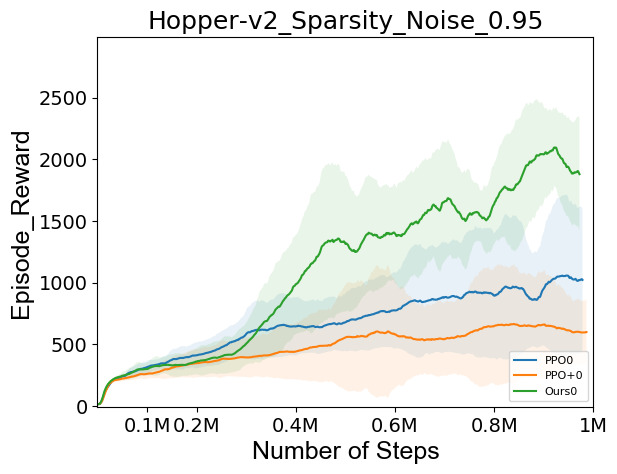}
    
    \includegraphics[width=.19\textwidth]{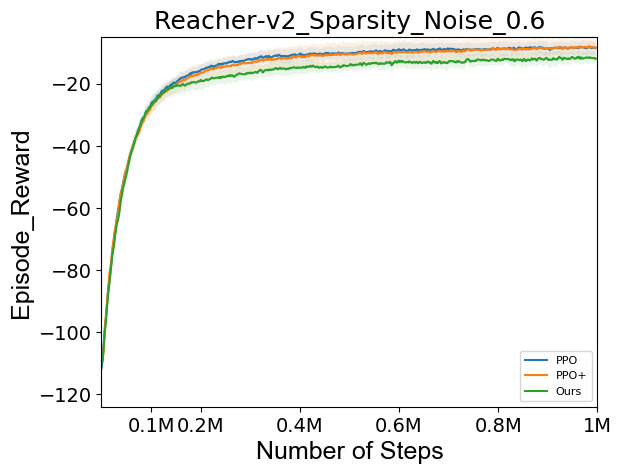}
    \includegraphics[width=.19\textwidth]{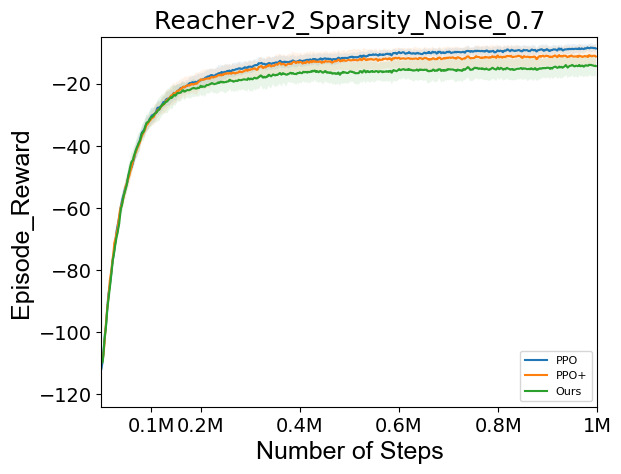}
    \includegraphics[width=.19\textwidth]{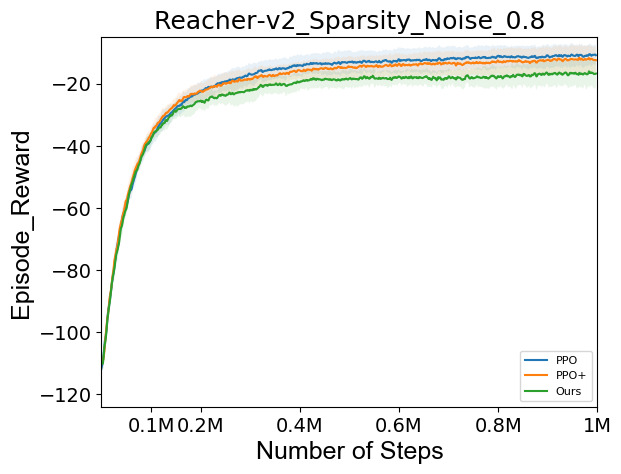}
    \includegraphics[width=.19\textwidth]{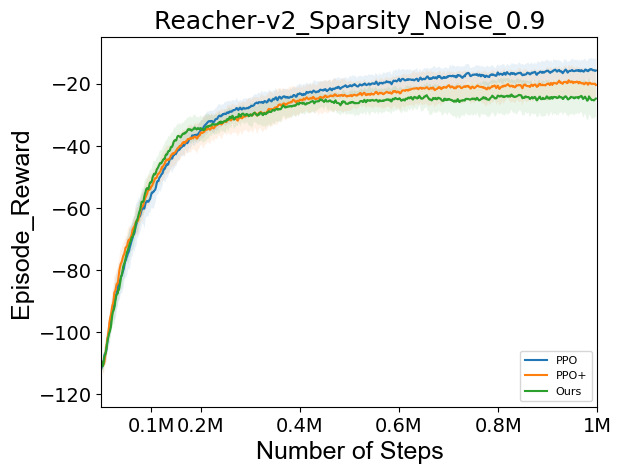}
    \includegraphics[width=.19\textwidth]{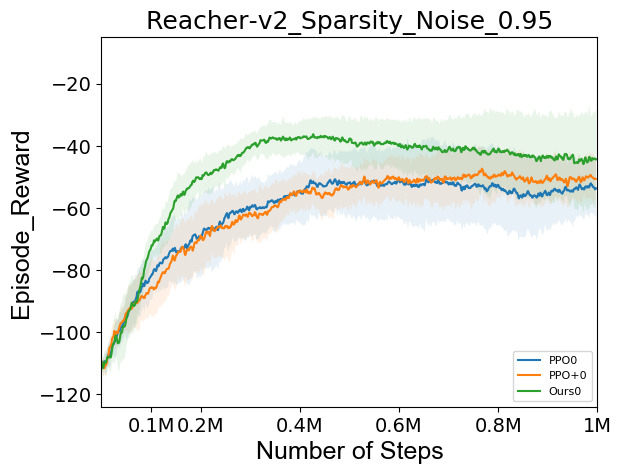}
    
    \includegraphics[width=.19\textwidth]{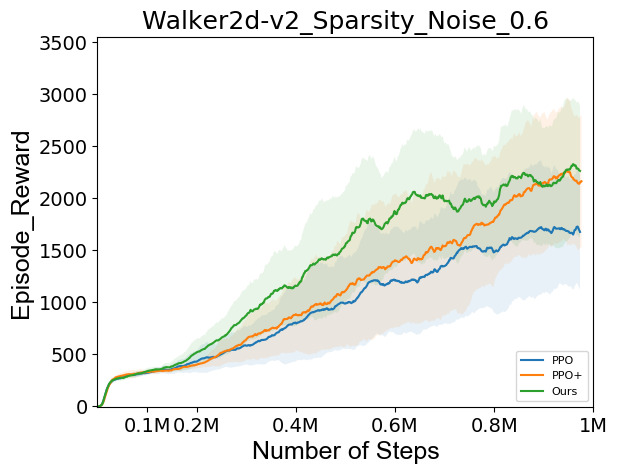}
    \includegraphics[width=.19\textwidth]{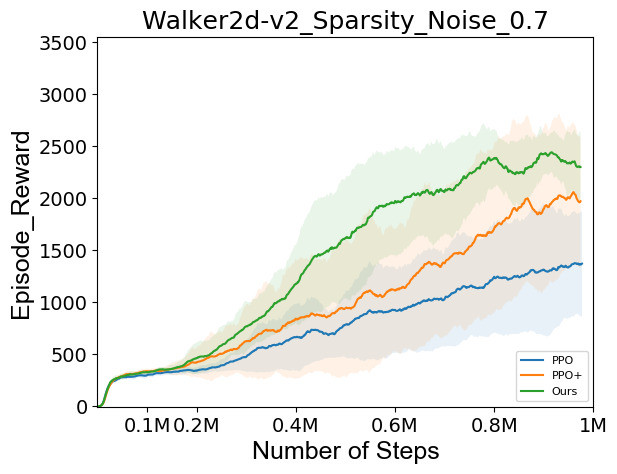}
    \includegraphics[width=.19\textwidth]{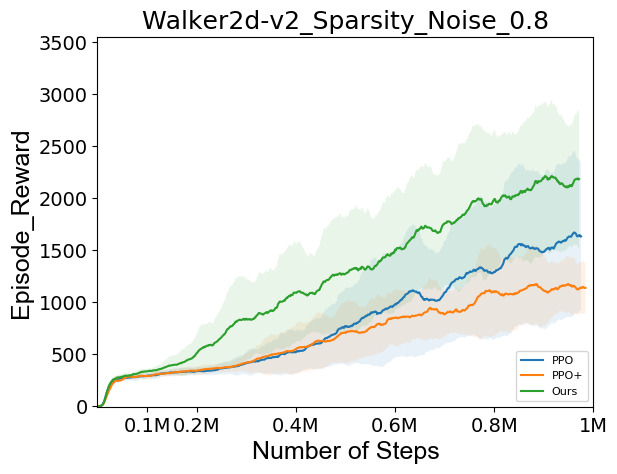}
    \includegraphics[width=.19\textwidth]{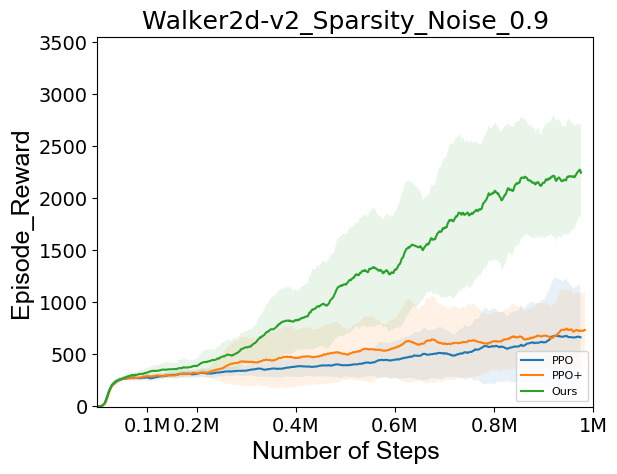}
    \includegraphics[width=.19\textwidth]{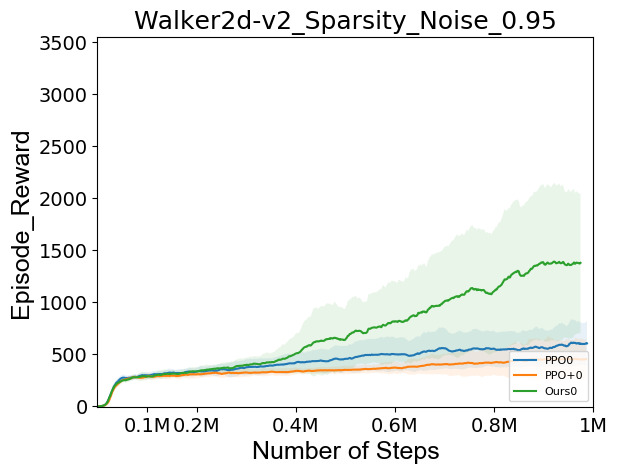}
    
    \caption{Sparsity noise - with probability p the reward is replaced with $0$}
\end{sidewaysfigure}

\begin{sidewaysfigure}
    \centering
    \includegraphics[width=.19\textwidth]{HalfCheetah-v2_Gaussian_Noise_0-0_Advantage}
    \includegraphics[width=.19\textwidth]{HalfCheetah-v2_Gaussian_Noise_0-1_Advantage}
    \includegraphics[width=.19\textwidth]{HalfCheetah-v2_Gaussian_Noise_0-2_Advantage}
    \includegraphics[width=.19\textwidth]{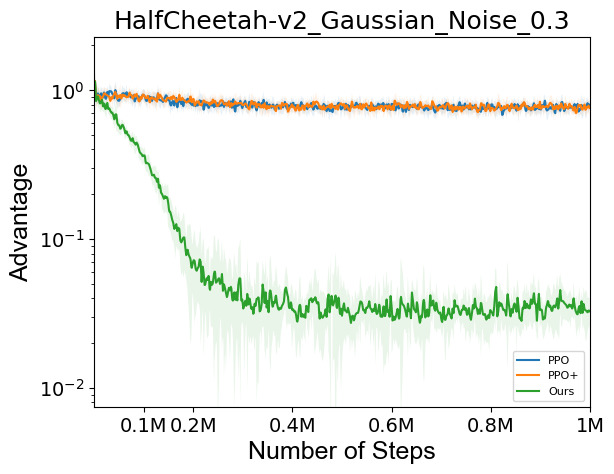}
    \includegraphics[width=.19\textwidth]{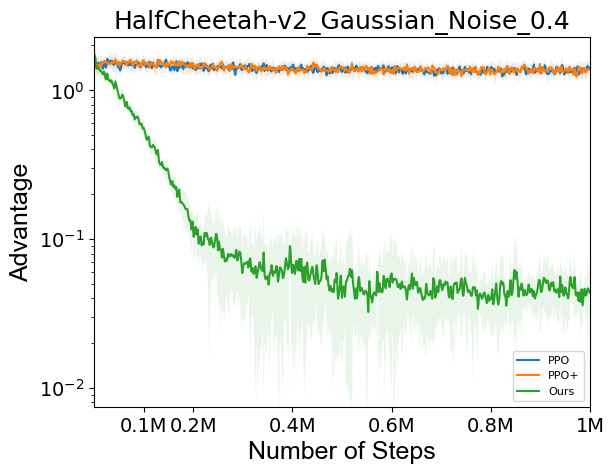}
    
    \includegraphics[width=.19\textwidth]{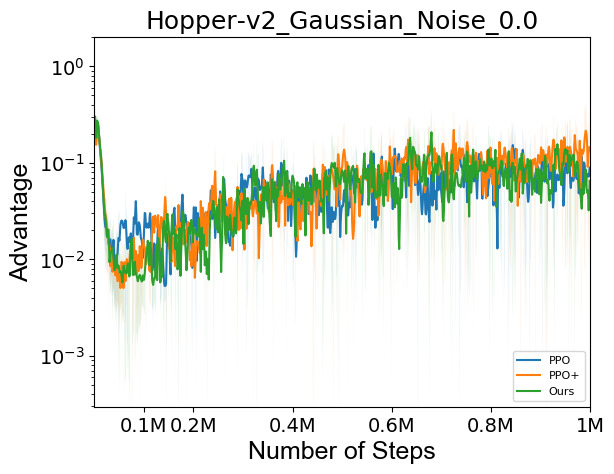}
    \includegraphics[width=.19\textwidth]{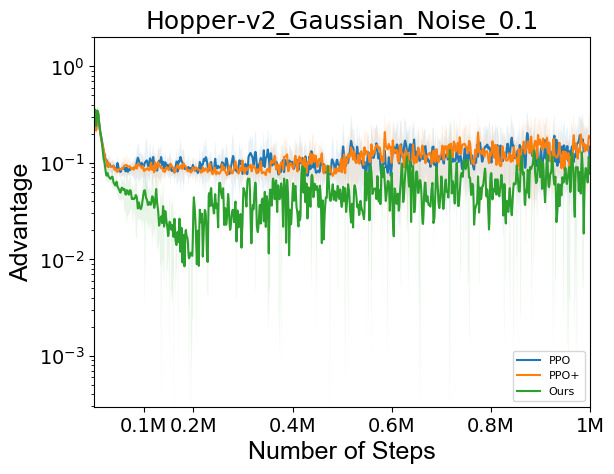}
    \includegraphics[width=.19\textwidth]{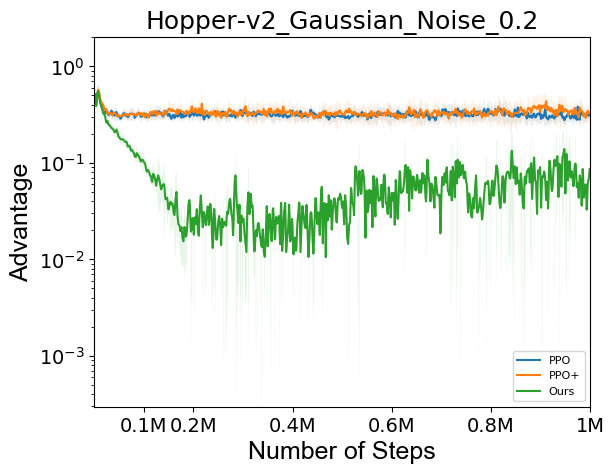}
    \includegraphics[width=.19\textwidth]{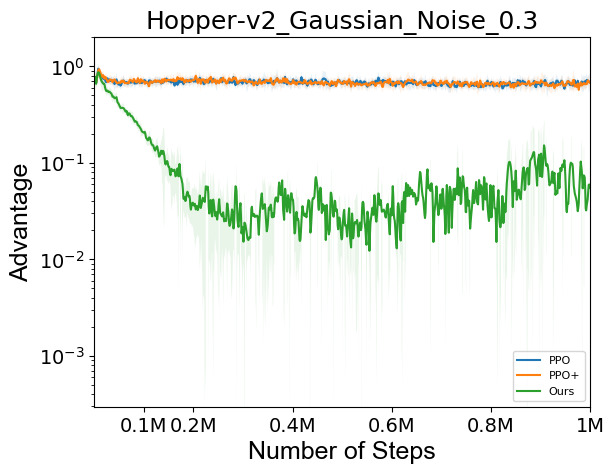}
    \includegraphics[width=.19\textwidth]{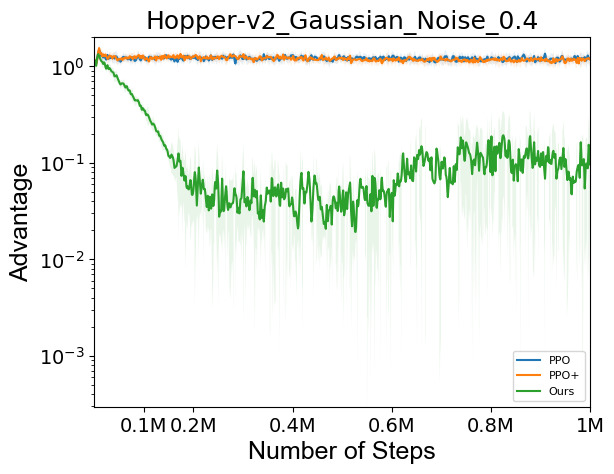}
    
    \includegraphics[width=.19\textwidth]{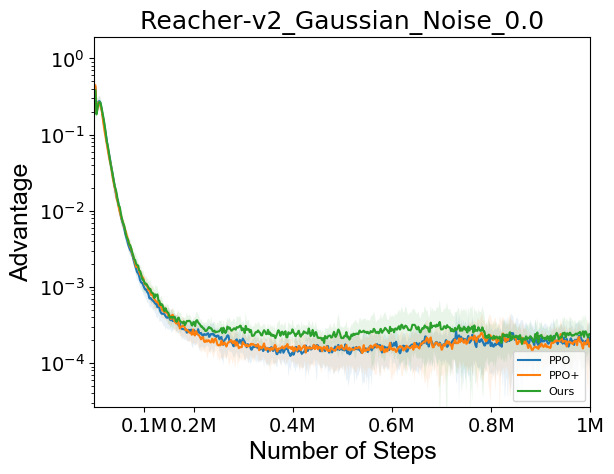}
    \includegraphics[width=.19\textwidth]{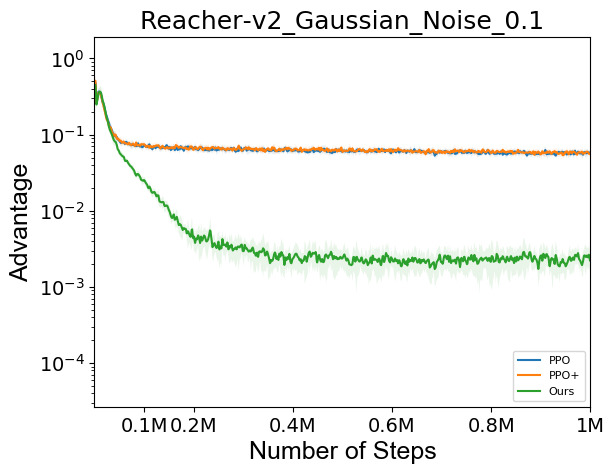}
    \includegraphics[width=.19\textwidth]{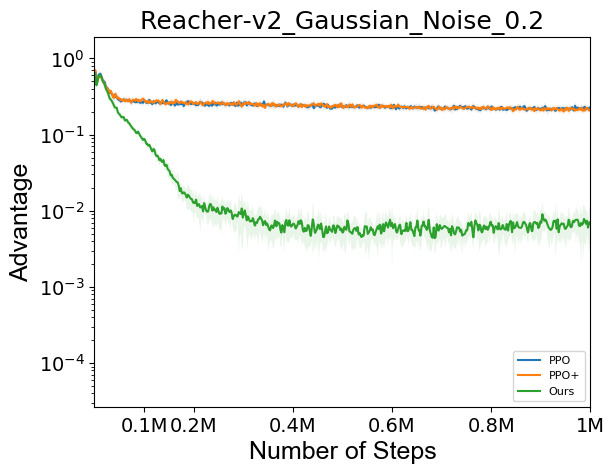}
    \includegraphics[width=.19\textwidth]{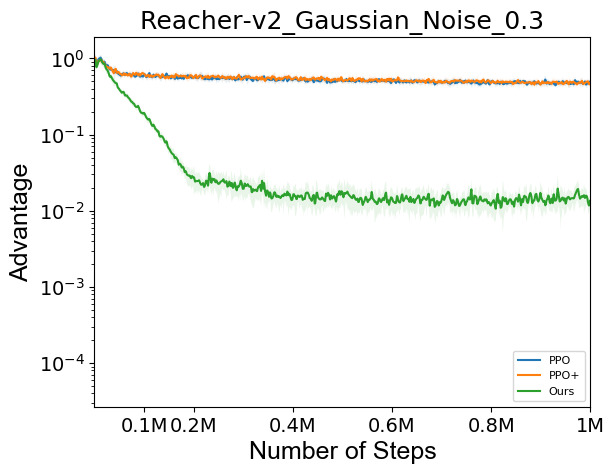}
    \includegraphics[width=.19\textwidth]{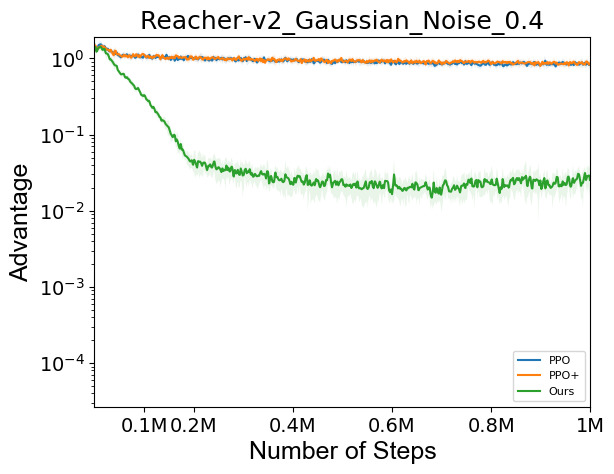}
    
    \includegraphics[width=.19\textwidth]{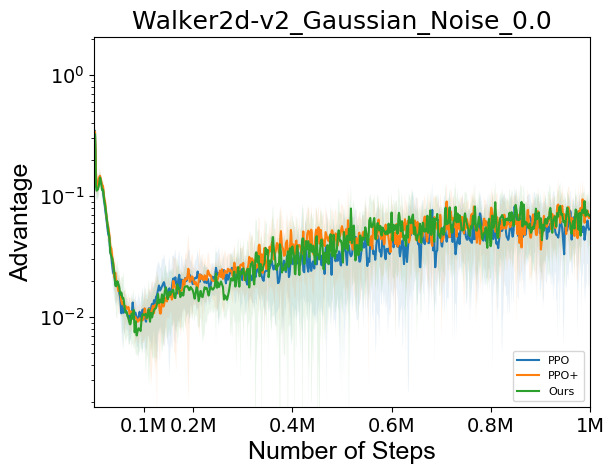}
    \includegraphics[width=.19\textwidth]{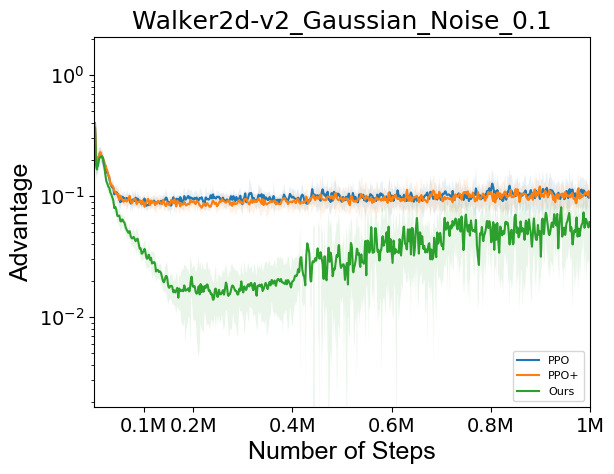}
    \includegraphics[width=.19\textwidth]{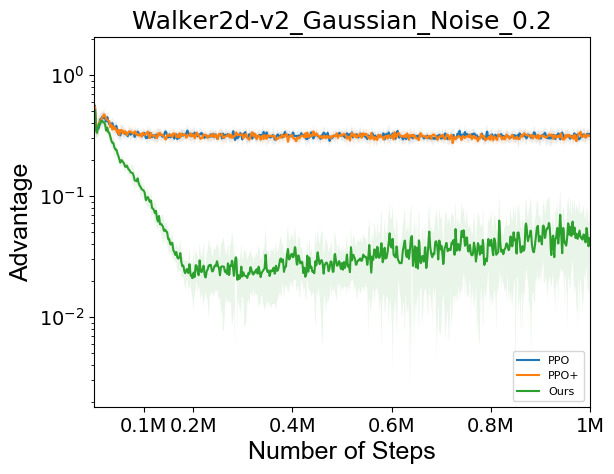}
    \includegraphics[width=.19\textwidth]{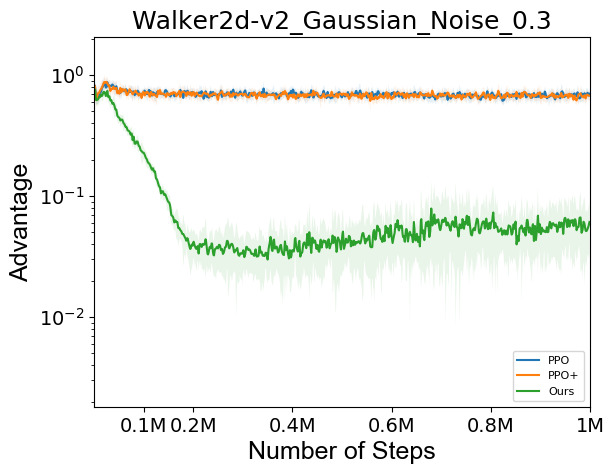}
    \includegraphics[width=.19\textwidth]{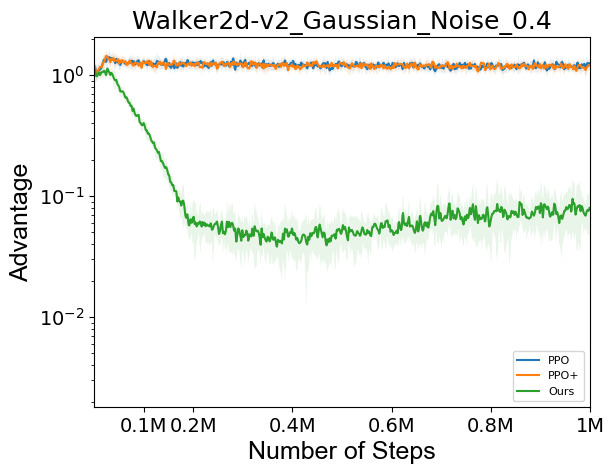}
    
    \caption{TD error with Gaussian noise}
\end{sidewaysfigure}

\begin{sidewaysfigure}
    \centering
    \includegraphics[width=.19\textwidth]{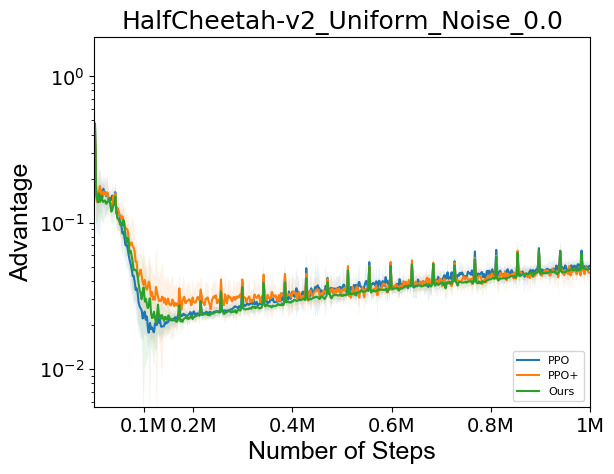}
    \includegraphics[width=.19\textwidth]{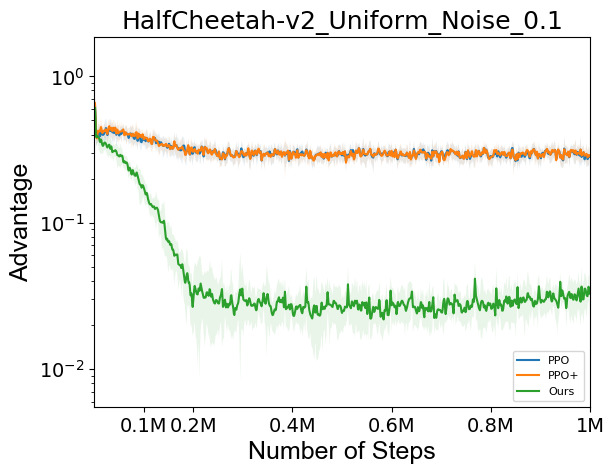}
    \includegraphics[width=.19\textwidth]{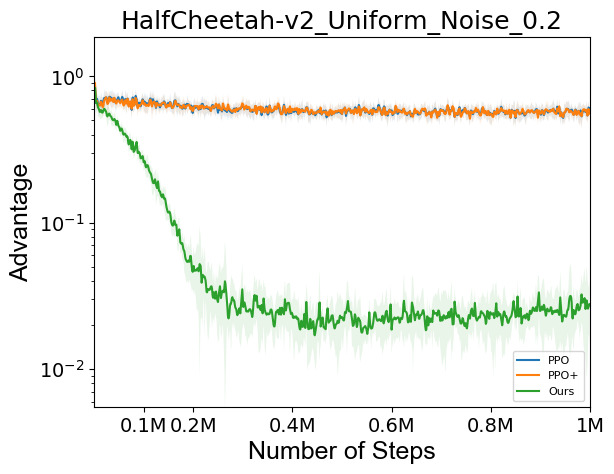}
    \includegraphics[width=.19\textwidth]{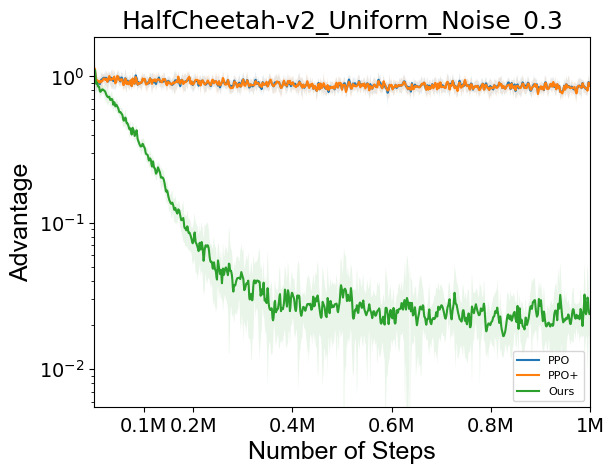}
    \includegraphics[width=.19\textwidth]{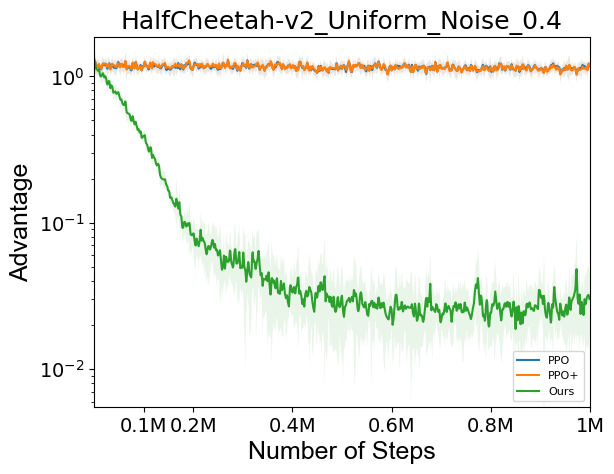}
    
    \includegraphics[width=.19\textwidth]{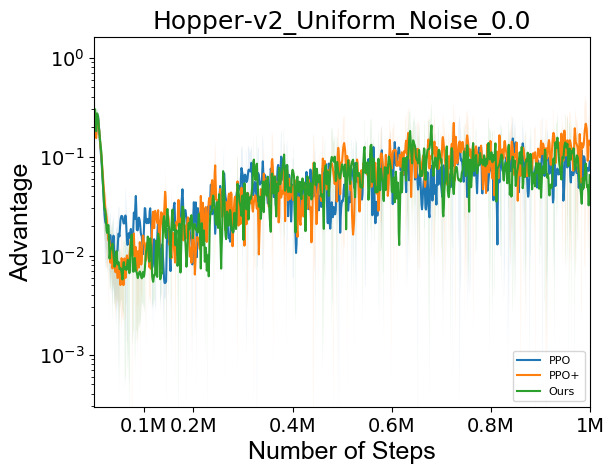}
    \includegraphics[width=.19\textwidth]{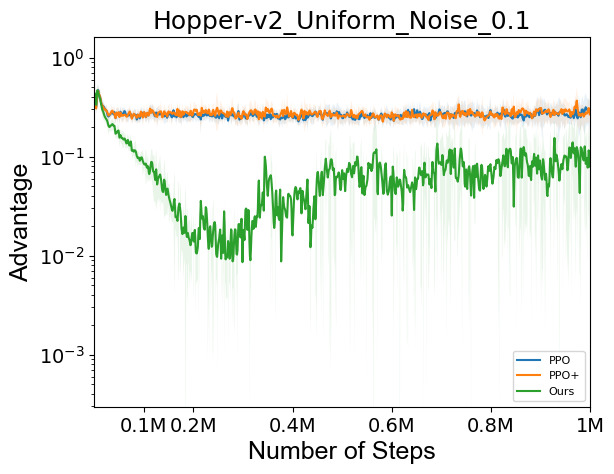}
    \includegraphics[width=.19\textwidth]{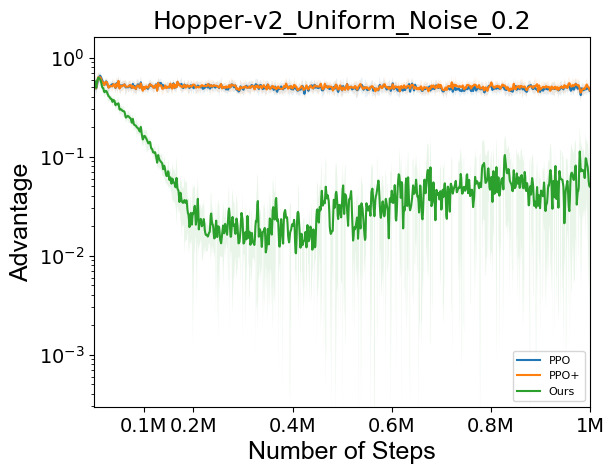}
    \includegraphics[width=.19\textwidth]{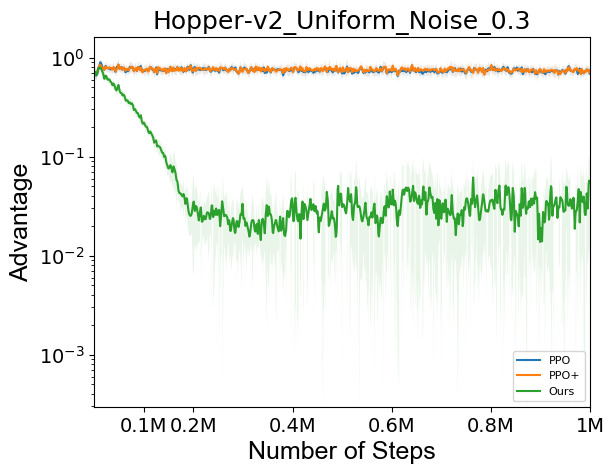}
    \includegraphics[width=.19\textwidth]{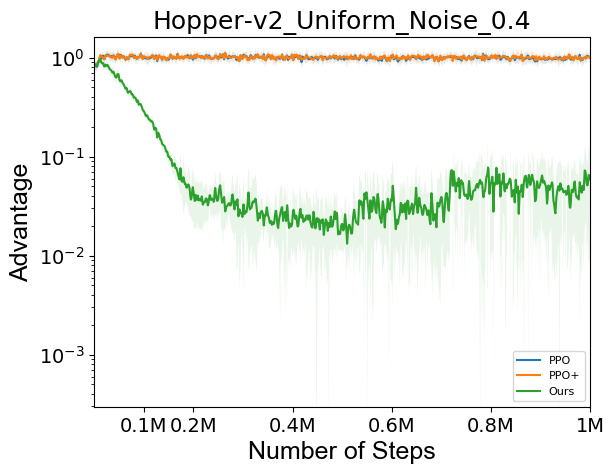}
    
    \includegraphics[width=.19\textwidth]{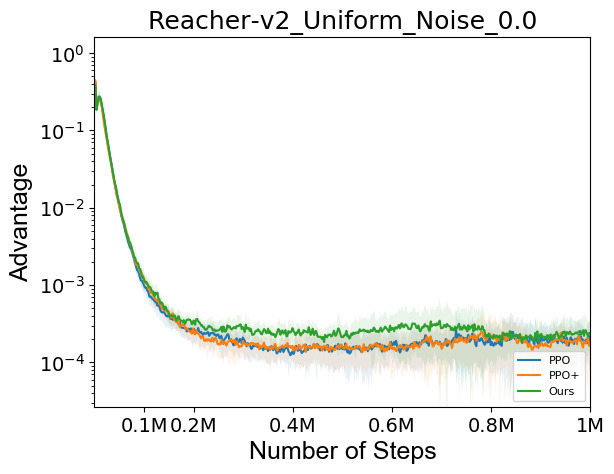}
    \includegraphics[width=.19\textwidth]{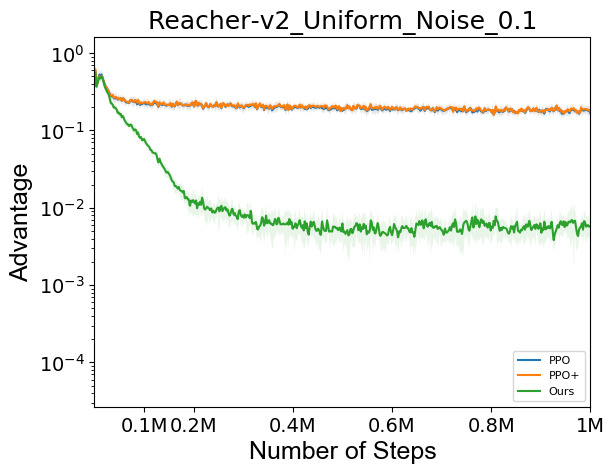}
    \includegraphics[width=.19\textwidth]{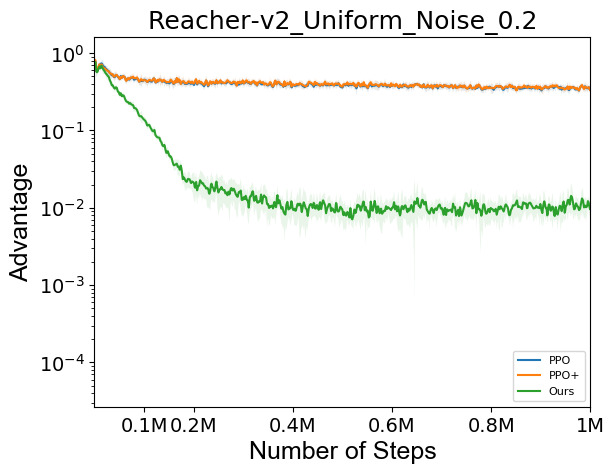}
    \includegraphics[width=.19\textwidth]{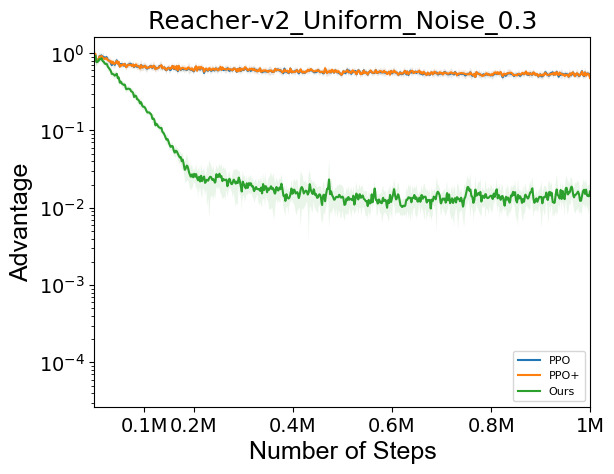}
    \includegraphics[width=.19\textwidth]{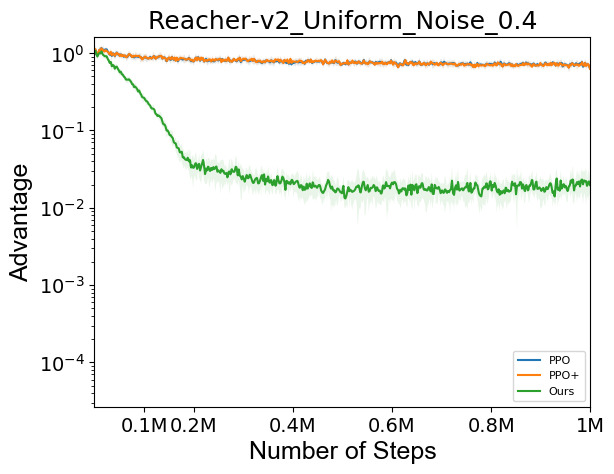}
    
    \includegraphics[width=.19\textwidth]{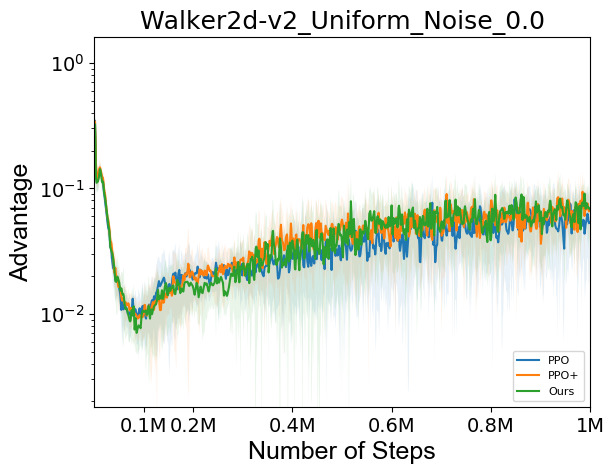}
    \includegraphics[width=.19\textwidth]{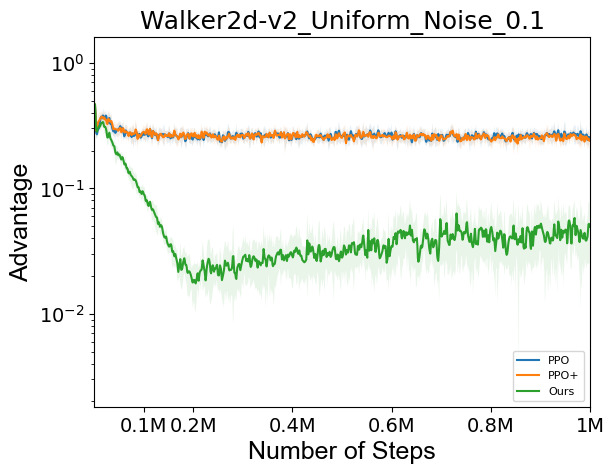}
    \includegraphics[width=.19\textwidth]{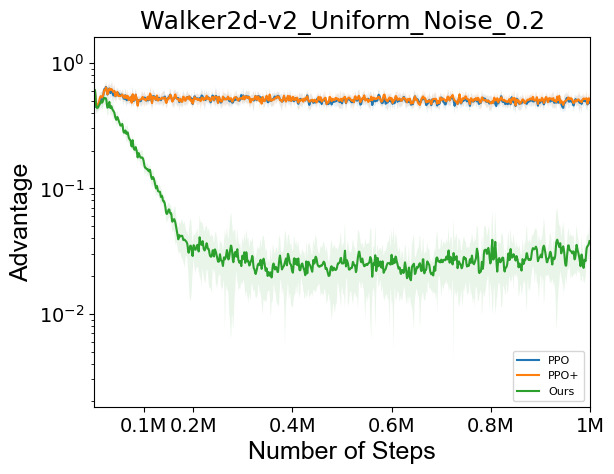}
    \includegraphics[width=.19\textwidth]{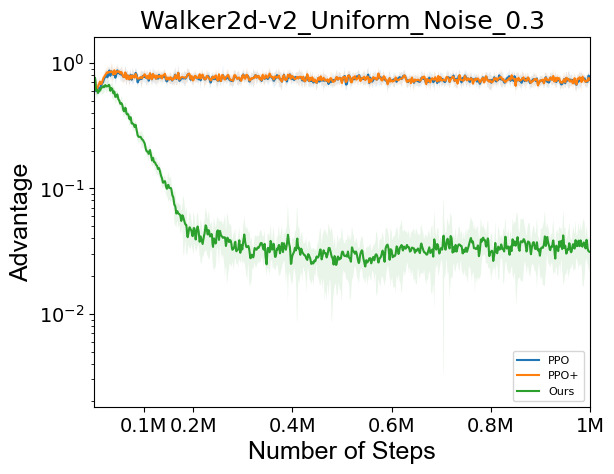}
    \includegraphics[width=.19\textwidth]{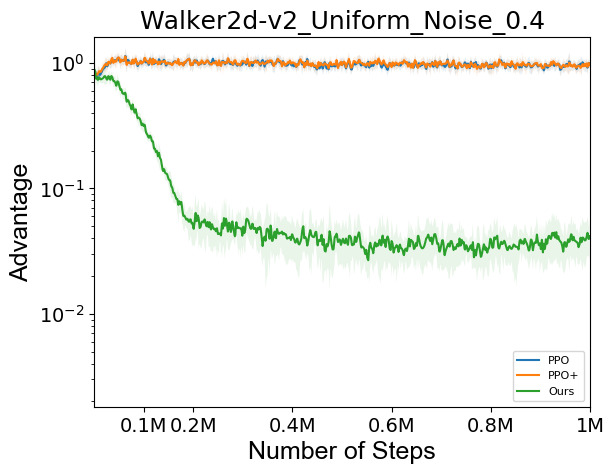}
    
    \caption{TD error with Uniform noise}
\end{sidewaysfigure}

\begin{sidewaysfigure}
    \centering
    \includegraphics[width=.19\textwidth]{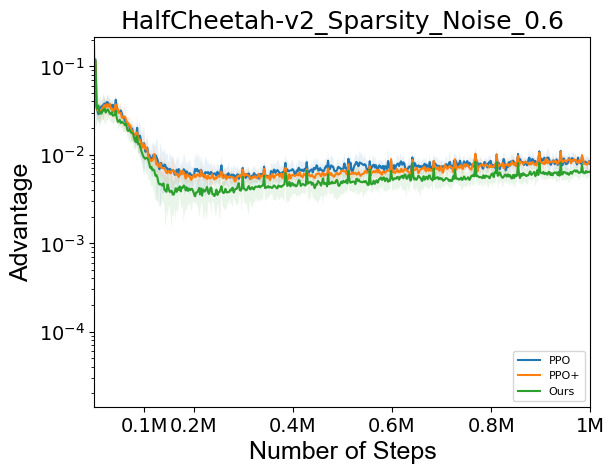}
    \includegraphics[width=.19\textwidth]{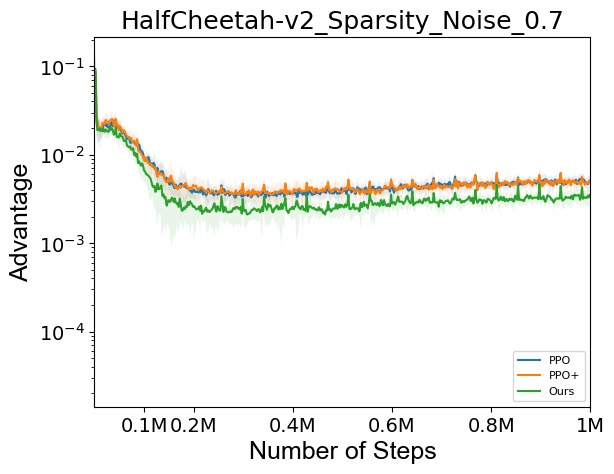}
    \includegraphics[width=.19\textwidth]{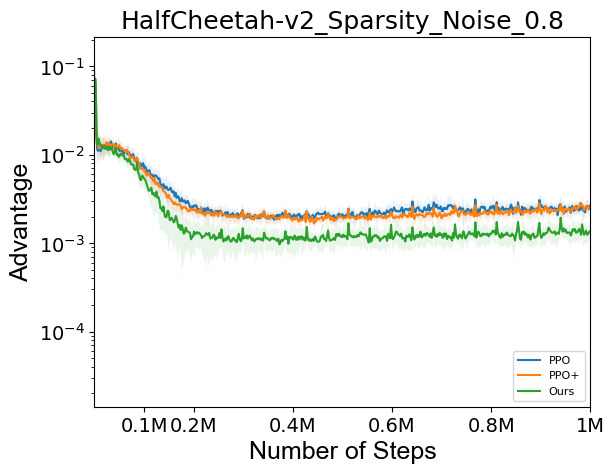}
    \includegraphics[width=.19\textwidth]{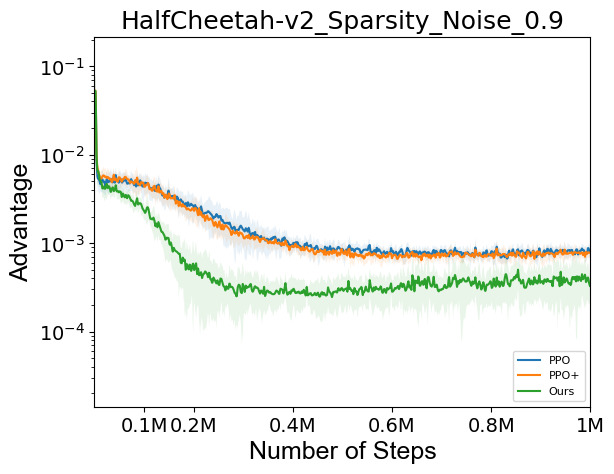}
    \includegraphics[width=.19\textwidth]{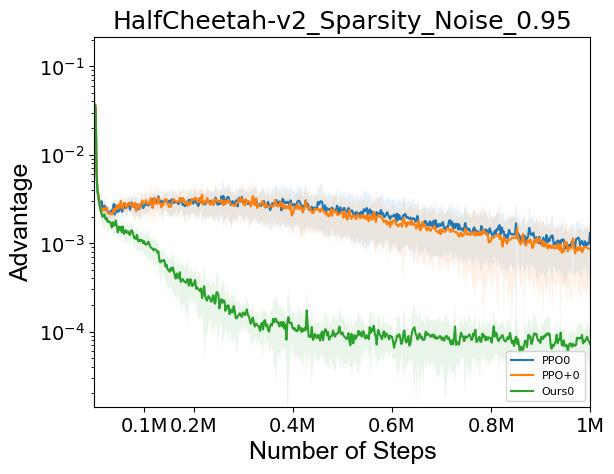}
    
    \includegraphics[width=.19\textwidth]{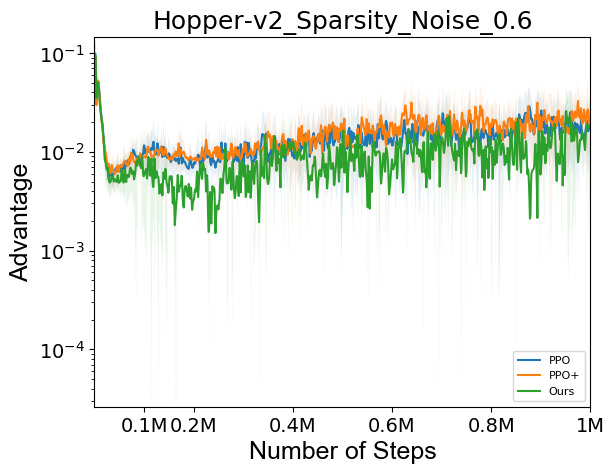}
    \includegraphics[width=.19\textwidth]{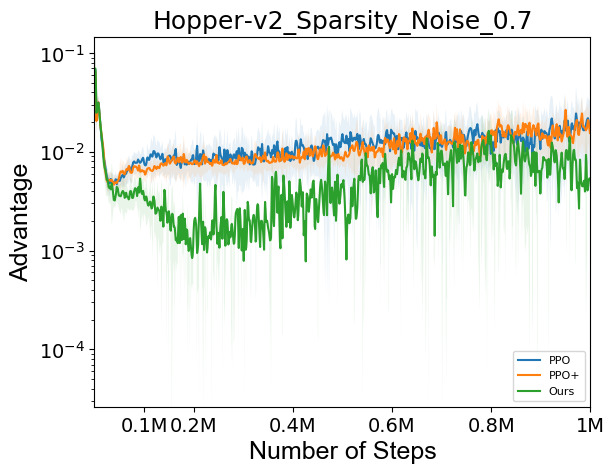}
    \includegraphics[width=.19\textwidth]{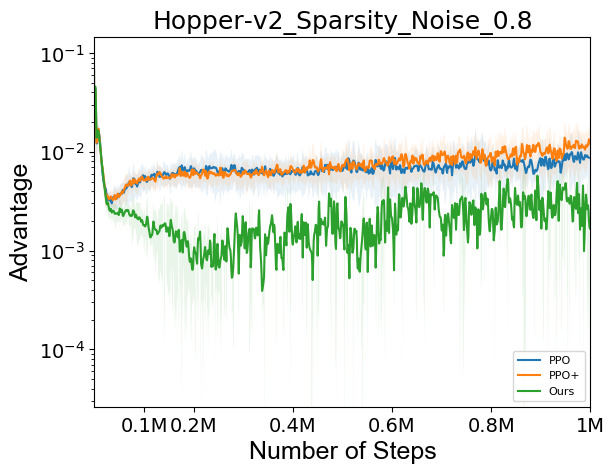}
    \includegraphics[width=.19\textwidth]{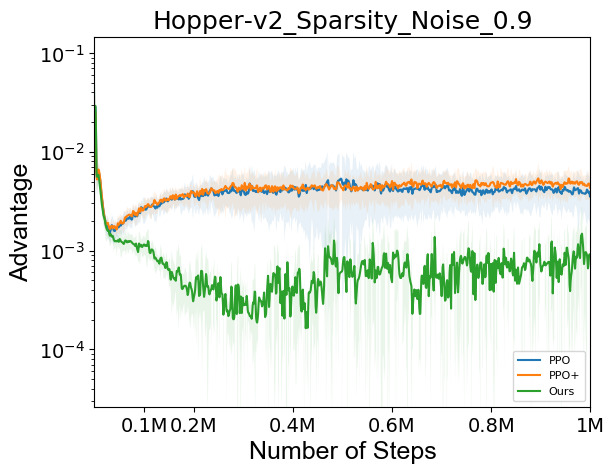}
    \includegraphics[width=.19\textwidth]{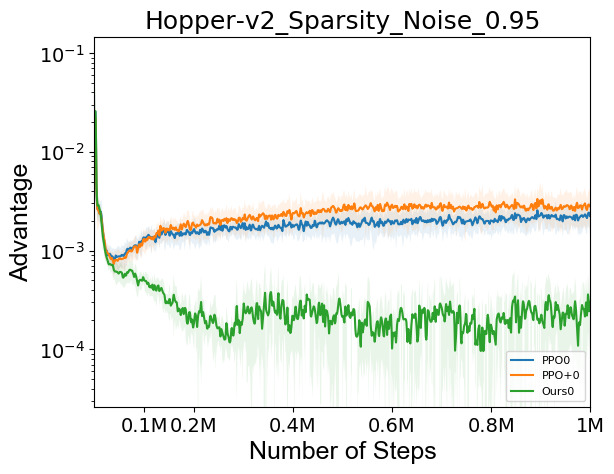}
    
    \includegraphics[width=.19\textwidth]{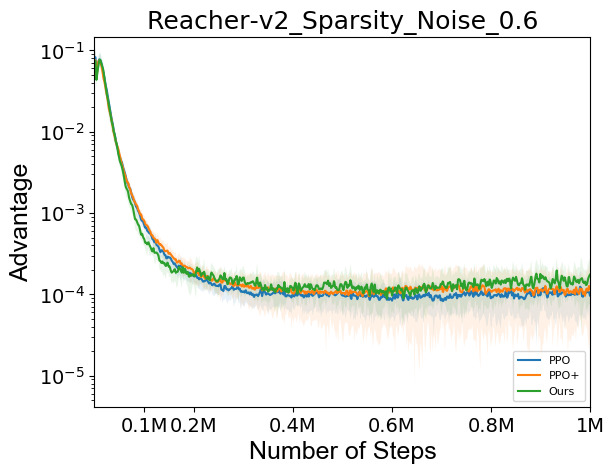}
    \includegraphics[width=.19\textwidth]{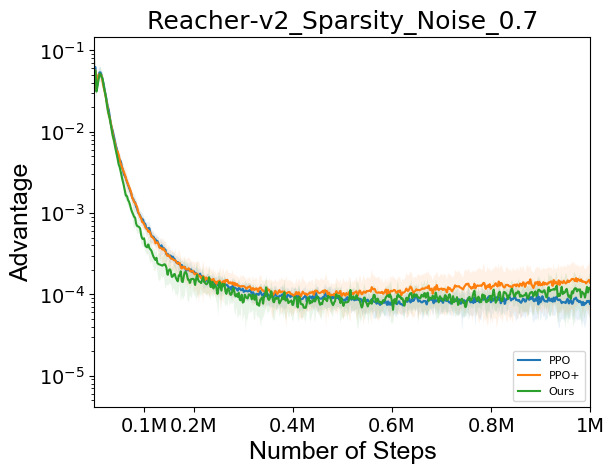}
    \includegraphics[width=.19\textwidth]{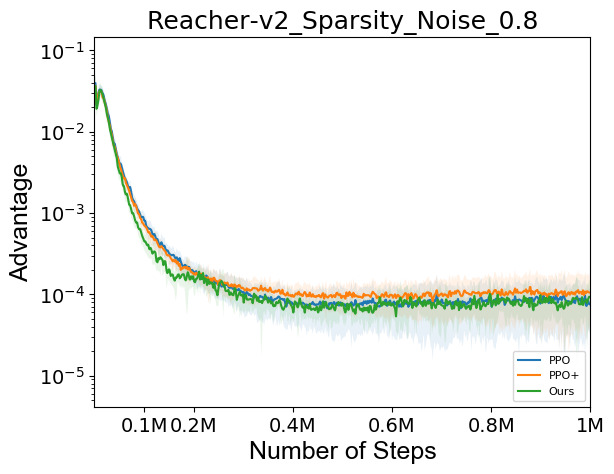}
    \includegraphics[width=.19\textwidth]{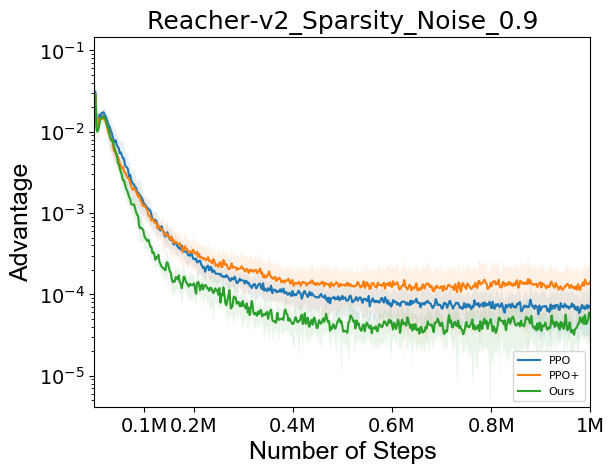}
    \includegraphics[width=.19\textwidth]{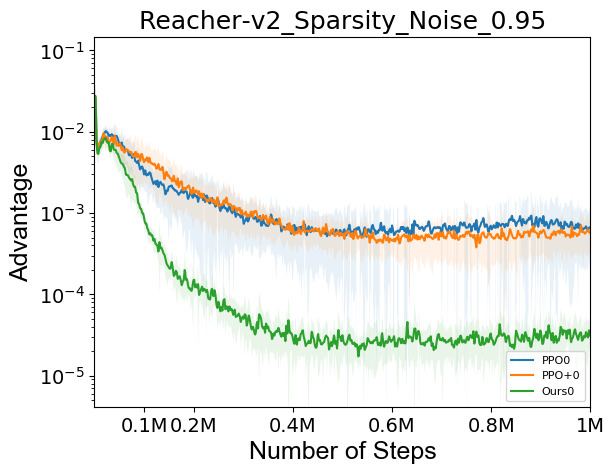}
    
    \includegraphics[width=.19\textwidth]{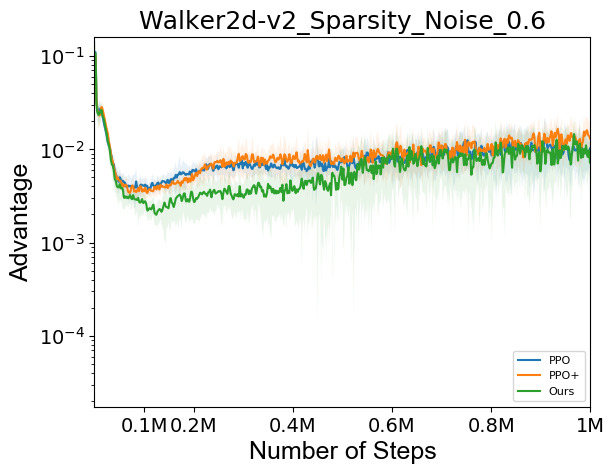}
    \includegraphics[width=.19\textwidth]{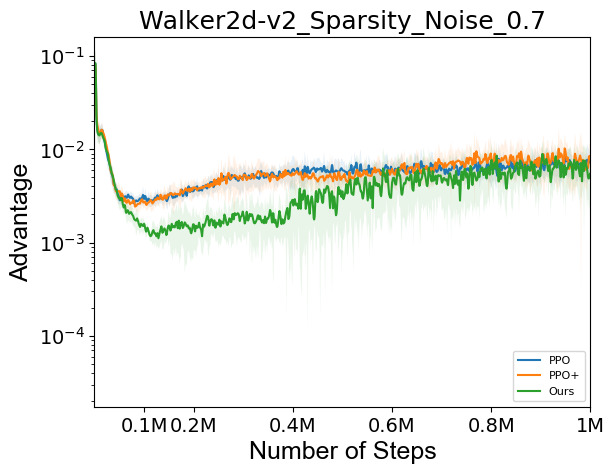}
    \includegraphics[width=.19\textwidth]{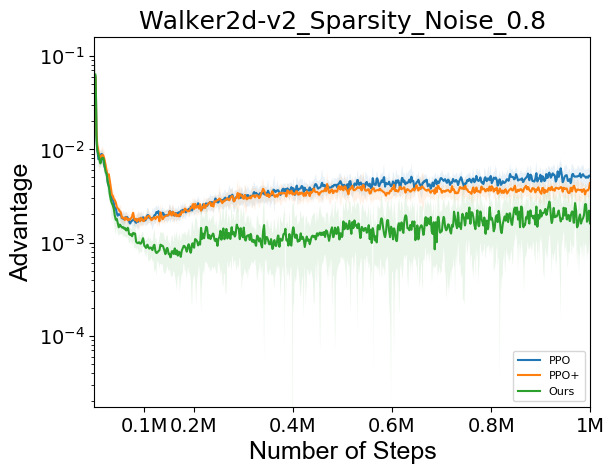}
    \includegraphics[width=.19\textwidth]{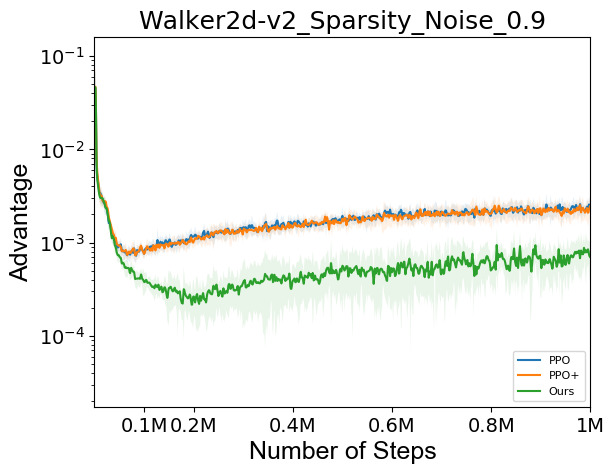}
    \includegraphics[width=.19\textwidth]{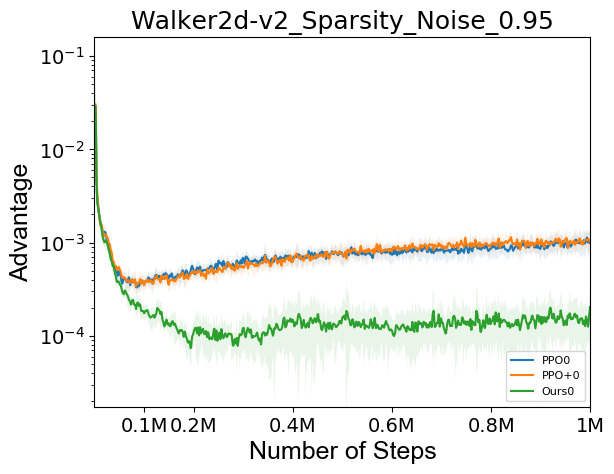}
    
    \caption{TD error with Sparsity noise}
\end{sidewaysfigure}

\subsubsection{Variance}

To discover whether there is indeed a variance reduction in the Bellman backup operator, we use a pretrained policy and reward estimator and run 100 more training episodes to retrain a new value function. We do this for 10 trials under all noise conditions and environments. The results of this can be seen in Tables~\ref{tab:var1},~\ref{tab:var2}, and~\ref{tab:var3}.

\begin{table}[H]
\label{tab:var1}
\centering
\small{
\begin{tabular}{|l|l|l|l|l|l|l|}
\hline
Environment                     & $\epsilon$& $\operatorname{var} (R_{true})$ & $\operatorname{var} (R_{corr})$ & $\operatorname{var} (\hat{R})$       & $\operatorname{MSE}(R_{corr},R_{true})$ & $\operatorname{MSE}(\hat{R},R_{true})$ \\ \hline
\multirow{4}{*}{HalfCheetah-v2} & 0.1           & 54.770     & 78.743        & \textbf{19.112} & 0.034               & \textbf{0.004}  \\ \cline{2-7} 
                                & 0.2           & 49.468     & 101.663       & \textbf{44.717} & 0.067               & \textbf{0.006}  \\ \cline{2-7} 
                                & 0.3           & 83.261     & 149.241       & \textbf{30.471} & 0.101               & \textbf{0.004}  \\ \cline{2-7} 
                                & 0.4           & 57.554     & 163.790       & \textbf{29.989} & 0.134               & \textbf{0.005}  \\ \hline
\multirow{4}{*}{Hopper-v2}      & 0.1           & 74.558     & 64.372        & \textbf{37.278} & 0.033               & \textbf{0.017}  \\ \cline{2-7} 
                                & 0.2           & 370.582    & 243.092       & \textbf{93.033} & 0.070               & \textbf{0.029}  \\ \cline{2-7} 
                                & 0.3           & 151.449    & 85.673        & \textbf{13.354} & 0.105               & \textbf{0.016}  \\ \cline{2-7} 
                                & 0.4           & 149.258    & 72.268        & \textbf{4.146}  & 0.142               & \textbf{0.020}  \\ \hline
\multirow{4}{*}{Reacher-v2}     & 0.1           & 4.947      & 6.047         & \textbf{0.209}  & 0.034               & \textbf{0.026}  \\ \cline{2-7} 
                                & 0.2           & 5.058      & 6.675         & \textbf{0.309}  & 0.071               & \textbf{0.027}  \\ \cline{2-7} 
                                & 0.3           & 5.110      & 7.692         & \textbf{0.622}  & 0.106               & \textbf{0.027}  \\ \cline{2-7} 
                                & 0.4           & 4.917      & 8.709         & \textbf{0.778}  & 0.143               & \textbf{0.027}  \\ \hline
\multirow{4}{*}{Walker2d-v2}    & 0.1           & 14.739     & 16.633        & \textbf{9.202}  & 0.034               & \textbf{0.014}  \\ \cline{2-7} 
                                & 0.2           & 14.710     & 16.499        & \textbf{4.345}  & 0.069               & \textbf{0.014}  \\ \cline{2-7} 
                                & 0.3           & 15.603     & 31.481        & \textbf{7.402}  & 0.102               & \textbf{0.013}  \\ \cline{2-7} 
                                & 0.4           & 17.224     & 29.401        & \textbf{5.587}  & 0.140               & \textbf{0.016}  \\ \hline
\end{tabular}}
\caption{The variance and mean squared error averaged over 10 trials of 100 episodes under uniform noise at varying $\epsilon$ probabilities where the likelihood of replacement with a random reward is $\epsilon$.}
\end{table}

\begin{table}[H]
\label{tab:var2}

\centering
\small{
\begin{tabular}{|l|l|l|l|l|l|l|}
\hline
Environment                     & $\epsilon$ & $\operatorname{var} (R_{true})$ & $\operatorname{var} (R_{corr})$ & $\operatorname{var} (\hat{R})$       & $\operatorname{MSE}(R_{corr},R_{true})$ & $\operatorname{MSE}(\hat{R},R_{true})$ \\ \hline
\multirow{5}{*}{HalfCheetah-v2} & 0.6           & 66.947     & 11.796         & \textbf{5.087}  & \textbf{0.003}      & 0.005           \\ \cline{2-7} 
                                & 0.7           & 67.277     & 6.051          & \textbf{2.461}  & \textbf{0.003}      & \textbf{0.003}  \\ \cline{2-7} 
                                & 0.8           & 67.176     & 3.368          & \textbf{1.107}  & \textbf{0.004}      & 0.005           \\ \cline{2-7} 
                                & 0.9           & 59.593     & 0.705          & \textbf{0.698}  & \textbf{0.002}      & 0.003           \\ \cline{2-7} 
                                & 0.95          & 69.234     & 0.316          & \textbf{0.183}  & \textbf{0.004}      & \textbf{0.004}  \\ \hline
\multirow{5}{*}{Hopper-v2}      & 0.6           & 343.477    & 54.577         & \textbf{14.768} & \textbf{0.011}      & 0.017           \\ \cline{2-7} 
                                & 0.7           & 505.215    & 46.451         & \textbf{10.550} & \textbf{0.014}      & 0.018           \\ \cline{2-7} 
                                & 0.8           & 10.416     & 0.483          & \textbf{0.218}  & \textbf{0.016}      & 0.017           \\ \cline{2-7} 
                                & 0.9           & 239.583    & 3.014          & \textbf{0.256}  & \textbf{0.018}      & \textbf{0.018}  \\ \cline{2-7} 
                                & 0.95          & 407.326    & 1.161          & \textbf{0.060}  & \textbf{0.021}      & \textbf{0.021}  \\ \hline
\multirow{5}{*}{Reacher-v2}     & 0.6           & 5.068      & 0.721          & \textbf{0.007}  & \textbf{0.020}      & 0.025           \\ \cline{2-7} 
                                & 0.7           & 4.811      & 0.406          & \textbf{0.004}  & \textbf{0.022}      & 0.026           \\ \cline{2-7} 
                                & 0.8           & 5.249      & 0.260          & \textbf{0.004}  & \textbf{0.023}      & 0.026           \\ \cline{2-7} 
                                & 0.9           & 5.008      & 0.054          & \textbf{0.002}  & \textbf{0.026}      & 0.027           \\ \cline{2-7} 
                                & 0.95          & 4.811      & 0.019          & \textbf{0.002}  & \textbf{0.027}      & \textbf{0.027}  \\ \hline
\multirow{5}{*}{Walker2d-v2}    & 0.6           & 28.327     & \textbf{4.720} & 4.929           & \textbf{0.010}      & 0.013           \\ \cline{2-7} 
                                & 0.7           & 84.195     & 7.686          & \textbf{4.380}  & \textbf{0.008}      & \textbf{0.011}  \\ \cline{2-7} 
                                & 0.8           & 11.193     & 0.518          & \textbf{0.285}  & \textbf{0.010}      & 0.011           \\ \cline{2-7} 
                                & 0.9           & 45.122     & 0.501          & \textbf{0.268}  & \textbf{0.013}      & \textbf{0.013}  \\ \cline{2-7} 
                                & 0.95          & 38.299     & 0.118          & \textbf{0.096}  & \textbf{0.011}      & \textbf{0.011}  \\ \hline
\end{tabular}}
\caption{Variance and mean squared error under sparsity inducing noise with $\epsilon$ indicating the probability that a reward is replaced with $0$.}
\end{table}

\begin{table}[H]
\label{tab:var3}
\centering
\small{
\begin{tabular}{|l|l|l|l|l|l|l|}
\hline
Environment                     & $\sigma$ & $\operatorname{var} (R_{true})$ & $\operatorname{var} (R_{corr})$ & $\operatorname{var} (\hat{R})$       & $\operatorname{MSE}(R_{corr},R_{true})$ & $\operatorname{MSE}(\hat{R},R_{true})$ \\ \hline
\multirow{5}{*}{HalfCheetah-v2} & 0              & 90.984          & 90.984          & \textbf{47.689}  & \textbf{0.000}      & 0.004           \\ \cline{2-7} 
                                & 0.1            & 58.407          & 69.364          & \textbf{27.151}  & 0.010               & \textbf{0.005}  \\ \cline{2-7} 
                                & 0.2            & 81.375          & 117.149         & \textbf{45.172}  & 0.040               & \textbf{0.004}  \\ \cline{2-7} 
                                & 0.3            & 48.259          & 134.517         & \textbf{42.478}  & 0.090               & \textbf{0.005}  \\ \cline{2-7} 
                                & 0.4            & \textbf{67.252} & 241.708         & 95.358           & 0.160               & \textbf{0.006}  \\ \hline
\multirow{5}{*}{Hopper-v2}      & 0              & 577.269         & 577.269         & \textbf{108.070} & \textbf{0.000}      & 0.019           \\ \cline{2-7} 
                                & 0.1            & 254.271         & 254.137         & \textbf{176.411} & \textbf{0.010}      & 0.012           \\ \cline{2-7} 
                                & 0.2            & 77.429          & 84.184          & \textbf{45.178}  & 0.040               & \textbf{0.016}  \\ \cline{2-7} 
                                & 0.3            & 1180.103        & 1213.711        & \textbf{67.992}  & 0.091               & \textbf{0.029}  \\ \cline{2-7} 
                                & 0.4            & 288.084         & 333.094         & \textbf{87.369}  & 0.160               & \textbf{0.026}  \\ \hline
\multirow{5}{*}{Reacher-v2}     & 0              & 4.947           & 4.947           & \textbf{0.020}   & \textbf{0.000}      & 0.024           \\ \cline{2-7} 
                                & 0.1            & 4.845           & 5.383           & \textbf{0.124}   & \textbf{0.010}      & 0.025           \\ \cline{2-7} 
                                & 0.2            & 4.930           & 6.983           & \textbf{0.371}   & 0.040               & \textbf{0.025}  \\ \cline{2-7} 
                                & 0.3            & 4.871           & 9.672           & \textbf{0.605}   & 0.090               & \textbf{0.026}  \\ \cline{2-7} 
                                & 0.4            & 5.326           & 13.629          & \textbf{1.481}   & 0.160               & \textbf{0.028}  \\ \hline
\multirow{5}{*}{Walker2d-v2}    & 0              & 95.792          & 95.792          & \textbf{75.649}  & \textbf{0.000}      & 0.009           \\ \cline{2-7} 
                                & 0.1            & 67.335          & 71.721          & \textbf{38.815}  & \textbf{0.010}      & 0.013           \\ \cline{2-7} 
                                & 0.2            & \textbf{8.616}  & 13.220          & 13.372           & 0.040               & \textbf{0.017}  \\ \cline{2-7} 
                                & 0.3            & 10.462          & \textbf{22.723} & 24.300           & 0.090               & \textbf{0.019}  \\ \cline{2-7} 
                                & 0.4            & 26.990          & 58.277          & \textbf{38.250}  & 0.161               & \textbf{0.017}  \\ \hline
\end{tabular}}
\caption{Variance and mean squared error under added Gaussian noise with variance $\sigma^2$.}

\end{table}

\end{document}